%% file: iclr2027_membenchmark.tex
\documentclass{article} 
\usepackage{iclr2027_conference,times}

\input{math_commands.tex}

\usepackage{booktabs}
\usepackage{array}
\usepackage{graphicx}
\usepackage{float}
\usepackage{amssymb}
\PassOptionsToPackage{hyphens}{url}
\usepackage{hyperref}
\usepackage{url}
\input{revision_macros}


\title{SceneMemBench: A Benchmark for Scene Memory in Interactive Video World Models}
\newcommand{\affmark}[1]{\textsuperscript{\mdseries #1}}
\author{Yuheng Si\affmark{1}\quad Linye Lyu\affmark{2}\quad Jiawei Zhou\affmark{2}\quad Lingyi Du\affmark{1}\quad Nanfei Ye\affmark{3}\\
  Aobo Liu\affmark{4}\quad Yu Liu\affmark{3}\quad Yu Li\affmark{1}\thanks{Corresponding author: Yu Li at
  \href{mailto:yu.li.sallylee@gmail.com}{yu.li.sallylee@gmail.com}.}\quad Cheng Zhuo\affmark{1}}
\iclrfinalcopy
\hypersetup{hidelinks,pdftitle={SceneMemBench: A Benchmark for Scene Memory in Interactive Video World Models},
  pdfauthor={Yuheng Si, Linye Lyu, Jiawei Zhou, Lingyi Du, Nanfei Ye, Aobo Liu, Yu Liu, Yu Li, Cheng Zhuo}}
\makeatletter
\newcommand{\resourcelink}[3]{\href{#1}{\raisebox{-0.17em}{\includegraphics[height=1em]{figures/link-icons/#2.pdf}}\,\textcolor[HTML]{1746A2}{#3}}}
\def\@maketitle{{\parindent=0pt\parskip=0pt\centering
  {\fontsize{17}{20}\selectfont\mdseries\scshape\@title\par}
  \vskip 6pt{\fontsize{12}{14}\selectfont\mdseries\@author\par}
  \vskip 2pt{\fontsize{10.95}{12.5}\selectfont\textsuperscript{1}Zhejiang University\quad
    \textsuperscript{2}Harbin Institute of Technology, Shenzhen\\
    \textsuperscript{3}Li Auto Inc.\quad \textsuperscript{4}Ocean University of China\par}
  \vskip 5pt{\centering\fontsize{11}{13}\selectfont\resourcelink{https://yuheng-si.github.io/SceneMemBench/}{rocket}{Project Page}\quad
    \resourcelink{https://github.com/Yuheng-Si/SceneMemBench}{github}{GitHub}\quad
    \resourcelink{https://huggingface.co/datasets/scenemembench/SceneMemBench}{dataset}{Dataset}\par}}\vskip 0.4em}
\makeatother

\begin{document}

\raggedbottom

\maketitle

\input{sections/00_abstract}

\begin{figure}[H]
  \centering
  \includegraphics[
    width=0.89\linewidth,
    trim=212pt 128pt 223pt 124pt,
    clip
  ]{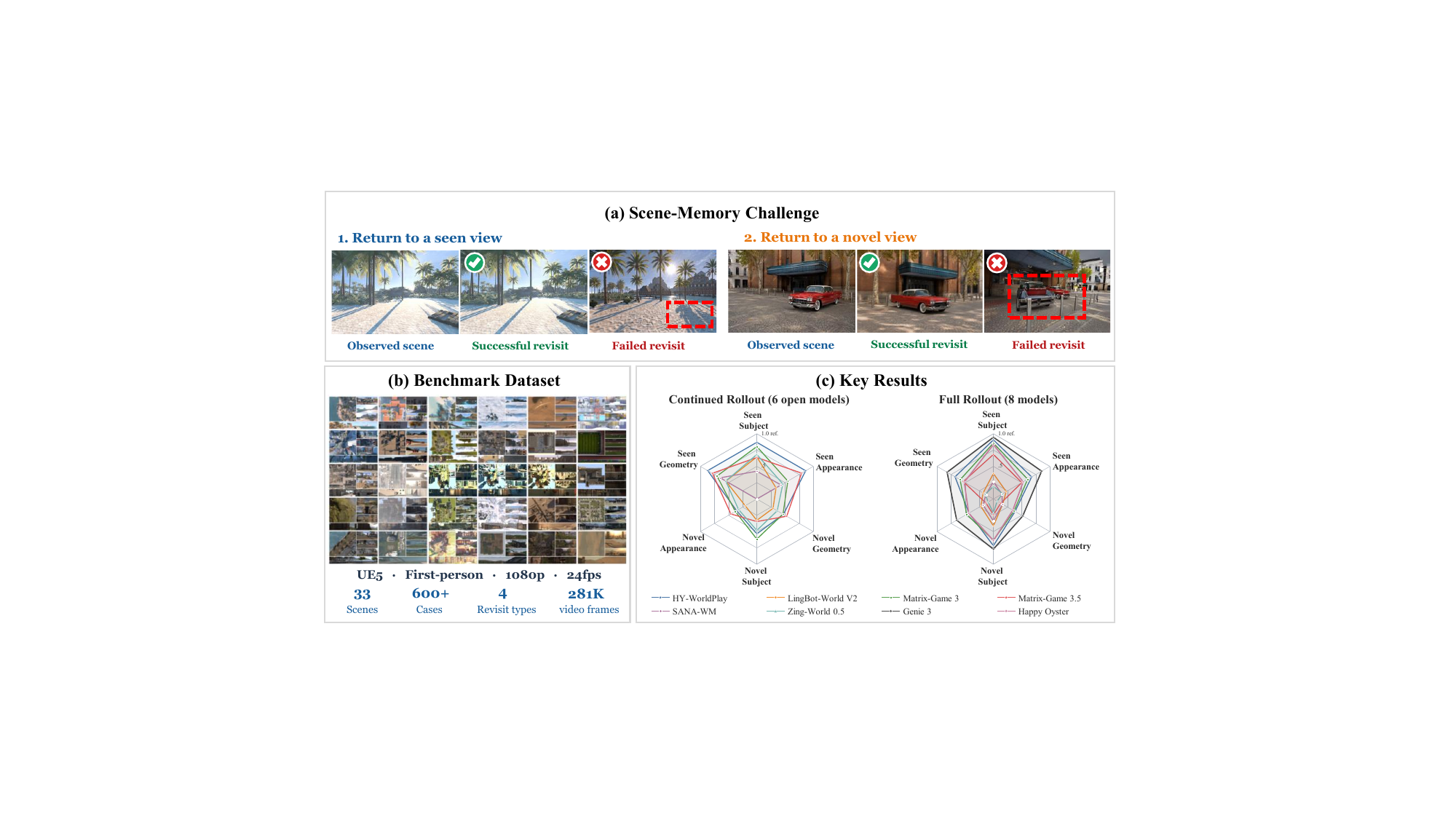}
\caption{
\textbf{SceneMemBench at a glance.}
(a) Representative successful and failed revisits for seen- and novel-view memory tasks.
(b) The benchmark contains 600 controlled trajectories across 33 scenes and 60 additional Full Rollout cases.
(c) Dimension-wise evaluation results under Continued Rollout and Full Rollout protocols.
}
\label{fig:teaser}
\end{figure}

\input{sections/01_introduction}
\input{sections/02_related_work}

\input{sections/03_scenemembench}
\input{sections/04_experiments}
\input{sections/05_conclusion}

\input{sections/06_statements}


\bibliography{references}
\bibliographystyle{iclr2027_conference}

\clearpage
\appendix
\input{sections/appendix}

\end{document}

%% file: math_commands.tex
\usepackage{amsmath,amsfonts,bm}

\def\eqref#1{equation~\ref{#1}}

\def\1{\bm{1}}

\DeclareMathAlphabet{\mathsfit}{\encodingdefault}{\sfdefault}{m}{sl}
\SetMathAlphabet{\mathsfit}{bold}{\encodingdefault}{\sfdefault}{bx}{n}



%% file: revision_macros.tex
\usepackage{xcolor}
\definecolor{revisionblue}{RGB}{0,70,160}
\newif\ifrevision
\revisionfalse
\newcommand{\revisioncolor}{\ifrevision\color{revisionblue}\fi}
\newcommand{\rev}[1]{{\revisioncolor #1}}
\newenvironment{revision}{\begingroup\revisioncolor}{\par\endgroup}

\newsavebox{\revisiontablebox}
\newcommand{\fitrevisiontable}[1]{%
  \sbox{\revisiontablebox}{#1}%
  \ifdim\wd\revisiontablebox>\linewidth
    \resizebox{\linewidth}{!}{\usebox{\revisiontablebox}}%
  \else\usebox{\revisiontablebox}\fi}

%% file: sections/00_abstract.tex
\begin{abstract}

Interactive video world models generate navigable worlds frame by frame, but a persistent world requires previously observed scenes to remain recoverable as users move away and return.
Existing memory evaluations, however, often focus on seen-view revisits and full rollouts, leaving \textbf{cross-view scene recovery} underexplored while entangling memory with trajectory-following errors and accumulated generation degradation.
We introduce \textsc{SceneMemBench}, a benchmark for scene memory in interactive video world models.
It evaluates two complementary dimensions: \textbf{seen-view recall versus novel-view recovery}, and \textbf{controlled diagnosis versus end-to-end interaction}.
We introduce Continued Rollout from a shared ground-truth observation history, alongside Full Rollout over complete model-generated interactions.
SceneMemBench contains 600 controlled trajectories across 33 Unreal Engine scenes and 60 Full Rollout cases, covering eight representative open- and closed-source models.
We evaluate scene recovery along subject consistency, appearance fidelity, and geometric consistency.
Results show that \textbf{\rev{novel-view recovery is generally more challenging than seen-view recall}}, while model profiles under controlled histories can differ markedly from those observed in full interaction.
Together, these findings reveal complementary failure modes obscured by seen-view revisits or end-to-end rollouts alone.

\end{abstract}

%% file: sections/01_introduction.tex
\section{Introduction}
\label{sec:introduction}

Recent video generation models have rapidly advanced in visual fidelity and temporal coherence \citep{kong2025hunyuanvideosystematicframeworklarge,wan2025wanopenadvancedlargescale,seedance2026seedance20advancingvideo,minimax2026h3}.
Building on this progress, interactive video world models generate navigable environments conditioned on user actions or camera controls \citep{genie3,wang2026matrixgame30realtimestreaming,sun2026worldplay,robbyantteam2026advancingopensourceworldmodels, mao2025yumeinteractiveworldgeneration}.
As users move through these generated worlds, previously observed content may leave the field of view and later reappear, possibly from a different viewpoint.
A persistent world therefore requires models not only to generate coherent video, but also to retain scene information and recover it consistently across interaction.

Recent benchmarks evaluate many aspects of interactive video world models, 
from geometric consistency~\citep{Duan_2025_ICCV} to multi-turn interaction over long-horizon objectives~\citep{ying2026wbenchcomprehensivemultiturnbenchmark,ding2026playworldbenchmarkingworldmodels}.
Among them, memory-oriented benchmarks examine whether previously observed content persists across camera motion, disappearance and reappearance, and out-of-sight state evolution~\citep{ye2026mindbenchmarkingmemoryconsistency,zhang2026mbenchcomprehensivebenchmarkmemory,ma2026sightmindevaluatingstate,chen2026memobenchbenchmarkingworldmodeling,xu2026worldroambenchopenworldbenchmarklonghorizon,gu2026r2mbenchevaluatingrevisitmemory}.
While these benchmarks provide valuable evidence of memory failures, two questions remain insufficiently addressed:
\textit{whether remembered scene information remains usable from previously unseen viewpoints, and how scene memory can be diagnosed without letting model-specific errors accumulated before the revisit dominate the measurement.}

These two gaps reflect two recurring design choices in existing memory benchmarks.
First, many protocols return the camera to a previously observed pose, which primarily tests seen-view recall rather than whether remembered scene information remains usable across viewpoints.
For example, after observing an object from position A, the camera may move to B while facing away and then turn back toward the object from a novel viewpoint.
The model should preserve the object's identity and surrounding spatial layout while producing changes in scale, perspective, and occlusion consistent with the new camera pose.
Second, memory is often assessed only after a complete model-generated rollout.
Differences in trajectory-following ability and generation stability can cause models to follow different paths  and ultimately reach different positions and viewing directions.
Trajectory drift and visual degradation complicate scene-memory diagnosis because pre-revisit histories differ across models.

To address these limitations, we introduce \textsc{SceneMemBench}, 
a benchmark for scene memory in interactive video world models.
To extend memory evaluation beyond seen-view recall, \textsc{SceneMemBench} defines two complementary memory tasks: \emph{Return to a Seen View}, which queries a camera pose present in the observation history, and \emph{Return to a Novel View}, which revisits the same scene from a previously unseen pose.
To reduce the influence of errors accumulated before the memory query, we introduce \emph{Continued Rollout}, which provides models with the same ground-truth observation history for each case and begins generation of a short revisit segment from a controlled query boundary.
We complement this diagnostic setting with \emph{Full Rollout}, in which the model generates the complete trajectory and must preserve scene memory throughout the interaction.
Together, these tasks and protocols evaluate scene recovery across viewpoints under both controlled histories and end-to-end model-generated rollouts.

Our contributions are summarized as follows:
\begin{itemize}

\item We introduce \textsc{SceneMemBench}, extending memory evaluation for interactive video world models beyond seen-view recall to novel-view recovery.
Its two complementary tasks, \emph{Return to a Seen View} and \emph{Return to a Novel View}, assess scene recovery from previously observed and unseen viewpoints.


\item  We introduce Continued Rollout as a controlled diagnostic protocol that provides all models with a shared ground-truth history and begins generation from a fixed query boundary, reducing errors accumulated before the memory query. We complement it with Full Rollout, which evaluates scene recovery throughout complete model-generated interactions.

\item We construct 600 controlled trajectories across 33 Unreal Engine scenes and 60 Full Rollout cases, and systematically evaluate eight representative open- and closed-source video world models.
Our results show that \rev{novel-view recovery is generally more challenging than seen-view recall}, while model profiles under controlled histories can differ markedly from those observed in full interaction.

\end{itemize}

%% file: sections/02_related_work.tex
\section{Related Work}
\label{sec:related_work}

\paragraph{Video Generation and Interactive Video World Models.}
Video generation has advanced from diffusion-based clip synthesis
to increasingly capable video foundation models
\citep{NEURIPS2022_39235c56,singer2022makeavideo}.
Recent open models, including CogVideoX, LTX-Video, HunyuanVideo,
and Wan, further improve generation quality, efficiency, and
versatility
\citep{yang2024cogvideox,hacohen2024ltxvideorealtimevideolatent,
kong2025hunyuanvideosystematicframeworklarge,wan2025wanopenadvancedlargescale}.
Building on these advances, interactive video world models extend
generation conditioned on an initial prompt or frame to sustain
interaction guided by successive user inputs, such as camera
trajectories or actions
\citep{wang2026matrixgame30realtimestreaming,qian2026matrixgame35enhancingrealtime,
zhu2026sanawmefficientminutescaleworld,robbyantteam2026advancingopensourceworldmodels,gao2026infiniteworldsversatileinteractions,
chen2026zing05playableworldsrealtime,genie3,happyoyster}.
Such interaction requires models to follow user instructions
while maintaining a consistent environment across multiple rounds.
In particular, users may leave a scene and later revisit it
from the same or a different viewpoint, requiring models to
retain its appearance and geometry beyond the current view.
Scene memory is therefore essential to persistent interactive
environments, yet generating long, visually coherent videos
alone does not establish this capability.
This motivates explicit memory evaluation to determine whether
interactive video world models can faithfully reconstruct previously
observed scenes when revisited.

\paragraph{Video World Model Memory Benchmarks.}
Existing benchmarks assess memory through revisits and the
persistence or evolution of content outside the field of
view~\citep{ye2026mindbenchmarkingmemoryconsistency,zhang2026mbenchcomprehensivebenchmarkmemory,ma2026sightmindevaluatingstate,chen2026memobenchbenchmarkingworldmodeling}.
However, these evaluations raise two distinct questions: whether remembered content remains usable across viewpoints, and whether scene memory can be diagnosed without being dominated by rollout errors.
Recovering a previously seen view alone does not establish that
a model can reconstruct the same scene from an unseen viewpoint.
Moreover, discrepancies at a revisit can reflect trajectory-following
errors and accumulated generation degradation, complicating the
assessment of memory.
Prior work mitigates these confounds through observed-history
conditioning, geometric alignment, temporal normalization, and
adaptive action execution~\citep{ye2026mindbenchmarkingmemoryconsistency,xu2026worldroambenchopenworldbenchmarklonghorizon,
gu2026r2mbenchevaluatingrevisitmemory,ding2026playworldbenchmarkingworldmodels}.
SceneMemBench examines both questions by explicitly distinguishing
\rev{seen-view recall from novel-view recovery} and evaluating
each through short continuations from ground-truth histories
and complete model-generated rollouts.
The former reduces errors accumulated before the memory query,
while the latter assesses scene recovery throughout the full
interaction.

In Table~\ref{tab:related_benchmarks}, Cases counts memory-evaluation examples,
and Evaluation Input lists input modalities supported by each benchmark's
implementation. GT Reference denotes scoring against ground-truth videos;
Controlled Memory Eval. denotes evaluation with a prescribed observation history that reduces pre-query rollout confounds.
Revisit Types summarizes camera motions used to revisit observed scenes,
while Rollout Length gives generated video lengths in seconds or frames.

\input{tables/table1_related_benchmarks}

%% file: tables/table1_related_benchmarks.tex
\begin{table}[!t]
  \caption{\textbf{Comparison of memory-oriented video world-model benchmarks.}
  $\checkmark$/$\triangle$/-- denote explicit/partial or related/unreported support.
  Txt./Img./Vid./Cam./Act./Obj. denote text/image/video/camera/action/objective.}
  \label{tab:related_benchmarks}
  \centering
  \scriptsize
  \setlength{\tabcolsep}{1.6pt}
  \renewcommand{\arraystretch}{1.12}
  \resizebox{\textwidth}{!}{%
  \begin{tabular}{@{}cccccccccc@{}}
    \toprule
    \raisebox{-3.1ex}[0pt][0pt]{Benchmark} &
    \raisebox{-3.1ex}[0pt][0pt]{\shortstack{Cases}} &
    \raisebox{-4.7ex}[0pt][0pt]{\shortstack{Evaluation\\Input}} &
    \raisebox{-4.7ex}[0pt][0pt]{\shortstack{GT\\Reference}} &
    \raisebox{-4.7ex}[0pt][0pt]{\shortstack{Controlled\\Memory Eval.}} &
    \multicolumn{4}{c}{Revisit Types} &
    \raisebox{-4.7ex}[0pt][0pt]{\shortstack{Rollout\\Length}} \\
    \cmidrule(lr){6-9}
    & & & &
    & \shortstack{Reversed\\Rotation}
    & \shortstack{Full\\Rotation}
    & \shortstack{Forward and\\Look Back}
    & \shortstack{New-Viewpoint\\Revisit} & \\
    \midrule
    MIND~\citep{ye2026mindbenchmarkingmemoryconsistency}
      & 150 & Txt. + Img./Vid. + Act. & $\checkmark$ & --
      & $\checkmark$ & -- & -- & --
      & 370--4,906 frames \\
    \addlinespace[1pt]
    MBench~\citep{zhang2026mbenchcomprehensivebenchmarkmemory}
      & 300 & Txt. + Img. + Cam./Act. & -- & $\triangle$
      & $\checkmark$ & -- & -- & -- & 25 s \\
    \addlinespace[1pt]
    MemoBench~\citep{chen2026memobenchbenchmarkingworldmodeling}
      & 360 & Txt. + Img. + Cam./Act. & $\checkmark$ & --
      & $\checkmark$ & -- & $\checkmark$ & --
      & 80--285 frames \\
    \addlinespace[1pt]
    WorldRoamBench~\citep{xu2026worldroambenchopenworldbenchmarklonghorizon}
      & 211 & Img. + Act. & -- & $\triangle$
      & $\checkmark$ & -- & -- & -- & 10--60 s \\
    \addlinespace[1pt]
    R2M-Bench~\citep{gu2026r2mbenchevaluatingrevisitmemory}
      & 300 & Txt. + Img. + Act./Cam. & -- & $\triangle$
      & $\checkmark$ & -- & -- & -- & $\approx481$ frames \\
    \addlinespace[1pt]
    PlayWorld~\citep{ding2026playworldbenchmarkingworldmodels}
      & 91 & Txt. + Img. + Obj. + Act. & -- & $\triangle$
      & $\checkmark$ & $\checkmark$ & $\checkmark$ & $\triangle$ & 10--60 s \\
    \addlinespace[1pt]
    \textbf{SceneMemBench (Ours)}
      & \textbf{660} & \textbf{Txt. + Img. + Vid. + Cam./Act.}
      & $\boldsymbol{\checkmark}$ & $\boldsymbol{\checkmark}$
      & $\boldsymbol{\checkmark}$ & $\boldsymbol{\checkmark}$
      & $\boldsymbol{\checkmark}$ & $\boldsymbol{\checkmark}$
      & \textbf{2--60 s} \\
    \bottomrule
  \end{tabular}%
  }
\end{table}

%% file: sections/03_scenemembench.tex
\section{SceneMemBench}
\label{sec:scenemembench}

\begin{figure}[!t]
  \centering
  \includegraphics[
    width=\textwidth,
    trim=40pt 90pt 100pt 49pt,
    clip
  ]{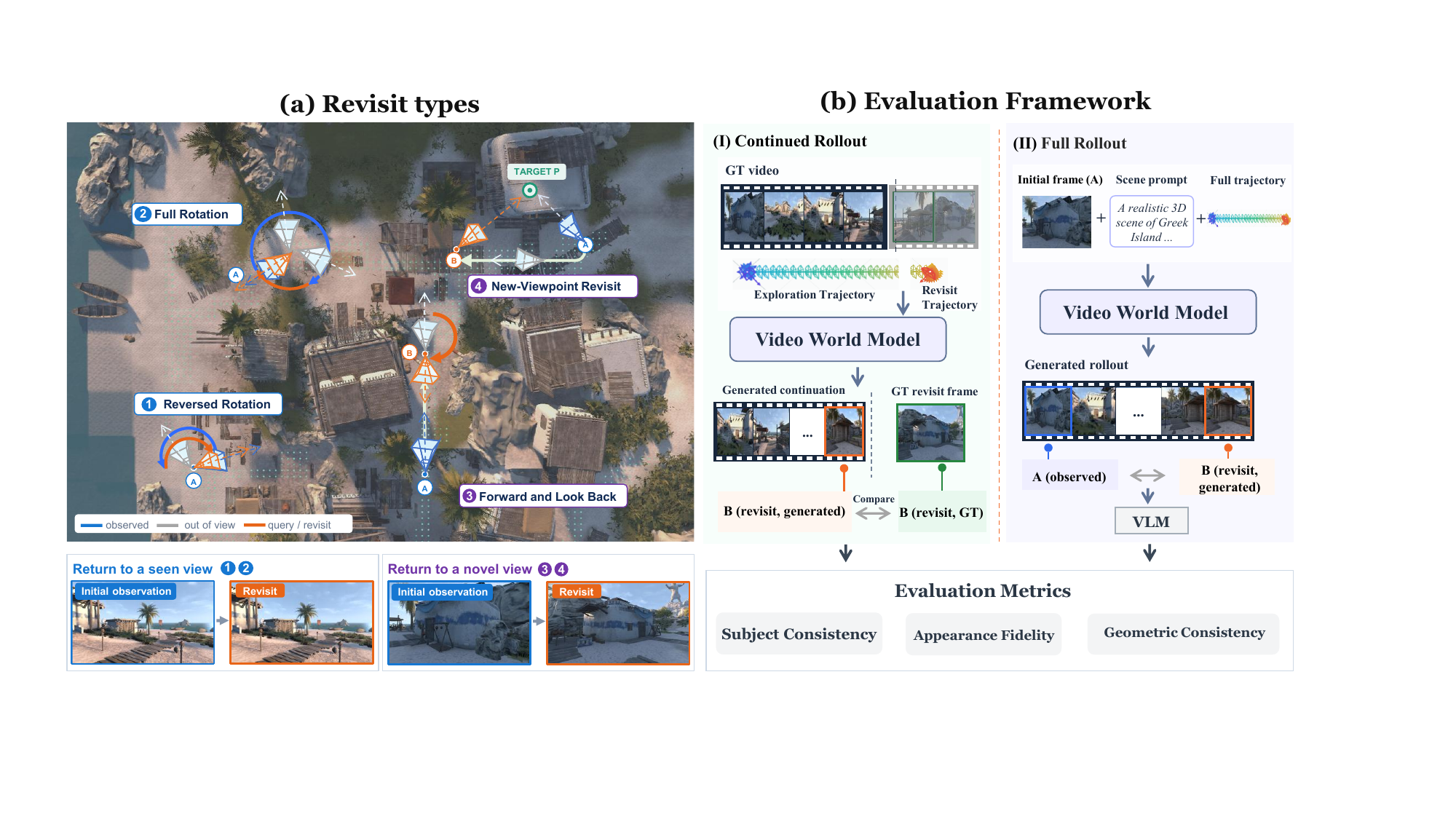}
  \caption{\textbf{Overview of SceneMemBench.} (a) Four revisit types
  instantiate two memory tasks: Reversed Rotation and Full Rotation return to
  seen views, whereas Forward and Look Back and New-Viewpoint Revisit query
  previously observed scene content from novel views. (b) The benchmark
  assesses these tasks under two complementary evaluation protocols. (I)
  Continued Rollout initializes the model with the prescribed GT history,
  generates the short query, and compares the output with held-out GT. (II)
  Full Rollout starts from the initial view, scene prompt, and reference
  actions, executes the complete interaction, and compares the revisit
  frame with the initial view. Each protocol reports Subject Consistency,
  Appearance Fidelity, and Geometric Consistency separately.}
  \label{fig:benchmark_overview}
\end{figure}

SceneMemBench evaluates whether interactive video world models can recover scene
content after it leaves the current view. As summarized in
Fig.~\ref{fig:benchmark_overview}, the benchmark combines ground-truth data
with two memory tasks, four revisit types, and two complementary
evaluation protocols. We first explain the data construction, then define the
memory tasks and revisit types, and finally present the evaluation framework.

\subsection{Dataset Construction}
\label{sec:dataset}

Even when asked to revisit the same scene, models generating a complete
interaction may follow different paths, run for different lengths of time,
and stop at different camera views. Their histories before the revisit can
therefore differ, as can the visual degradation accumulated along the way.
To compare scene memory after a common observation history, Continued Rollout
initializes each model from the same recorded GT history and starts the short
generated revisit at a fixed point in the trajectory. WorldMem and Memory
Forcing have used GT histories
to evaluate subsequent generation in Minecraft~\citep{NEURIPS2025_470629a4,huang2025memoryforcingspatiotemporalmemory},
but comparable trajectory data for diverse, realistic scenes remain limited.
We therefore record GT video trajectories in static rendered scenes with
replayable camera motion, preserving the pre-query observations, camera poses,
query boundary, and reference views needed for scoring.

We record trajectories in 33 Unreal Engine 5.6 scenes spanning urban and
residential areas, historical ruins and stylized architecture, industrial and
transportation facilities, indoor spaces, and natural terrain. Within each
scene, we select navigable regions with recognizable structures that remain
stable across viewpoints and construct deterministic rotational and translational
revisits. The resulting 600 controlled trajectories record rendered video,
frame-wise camera poses, a scene prompt, a reference action sequence, and an
explicit history--query boundary. For Full Rollout, the benchmark additionally
provides 60 image-anchored cases, each comprising an initial image, a scene
prompt, and a reference action sequence. Appendix~\ref{app:dataset_protocol_metric_details}
provides detailed capture settings and dataset statistics; the memory queries
supported by these data are defined next.

\subsection{Scene-Memory Tasks and Revisit Types}
\label{sec:tasks_protocols}

We define two scene-memory tasks by relating the scored query view to the
complete observation history, as illustrated in
Fig.~\ref{fig:benchmark_overview}(a). \emph{Return to a Seen View} asks whether
a model can recover a view that has already appeared in the history, whereas
\emph{Return to a Novel View} asks it to reconstruct previously observed scene
content from a camera pose absent from that history. The distinction therefore
separates view recall from reconstruction across camera poses while keeping the
underlying scene fixed.

Two rotational revisit types instantiate Return to a Seen View. In
\emph{Reversed Rotation}, the camera turns away from its initial view and then
retraces the rotational path in the opposite direction. In \emph{Full
Rotation}, the camera completes a continuous $360^\circ$ in-place rotation:
  for Continued Rollout, the first $270^\circ$ forms the Context, and the remaining $90^\circ$ forms the
query that closes the loop. Although their intervening motions differ, both
revisit types culminate in a view already represented in the history.

Two further revisit types instantiate Return to a Novel View. In \emph{Forward and Look Back}, the
camera moves from position $A$ to position $B$ and turns by $180^\circ$ at
$B$; the Continued Rollout Query is the final $90^\circ$ of this turn and does not translate back
to $A$. In \emph{New-Viewpoint Revisit}, the camera observes a target anchor
$P$ from $A$, moves to $B$ while $P$ leaves the field of view, and then turns
until $P$ reappears from the new position and orientation. In both types, novelty
is defined by the constructed trajectory rather than a universal threshold on
translation, rotation, or view overlap. Together, the four
revisit types provide the shared scene-memory queries evaluated under the two
protocols described next.

\subsection{Evaluation Framework}
\label{sec:evaluation_framework}

We assess scene recovery under two complementary evaluation protocols.
\emph{Continued Rollout} evaluates a short revisit generated from a prescribed
GT history, reducing errors accumulated before the query. \emph{Full Rollout}
evaluates recovery after a complete model-generated interaction, retaining
the effects of action execution and long-horizon generation stability.
Both assess \emph{Subject Consistency}, \emph{Appearance Fidelity}, and
\emph{Geometric Consistency}, but use different evidence and scoring procedures.
Their scores are reported separately, without combining them or directly
comparing their numerical values.

\subsubsection{Continued Rollout}
\label{sec:gt_anchored}

\paragraph{Evaluation Procedure.}
For each case, every model receives the same complete GT Context through its
native history or memory state. It then generates the short Query from the
prescribed boundary, using camera controls adapted to its native interface.
We score the final decoded frame against a reference from the GT Query,
which is withheld from generation. Initialization therefore supplies the complete
observation history rather than a single conditioning image.
Appendix~\ref{app:model_adapters} describes the native initialization and controls.

Reference selection accounts for residual camera-control errors that can
otherwise obscure scene recovery (Fig.~\ref{fig:hy_reference_comparison}).
For \emph{Return to a Seen View}, we keep the generated final frame fixed and
use image matching to select a GT reference with a better-matching viewpoint
only within a local neighborhood of the GT endpoint. This allows an incomplete
return turn while excluding references near the start of the continuation,
so a static output cannot gain an inflated score by matching its starting view.
For \emph{Return to a Novel View}, substantial content differences can make
viewpoint matching unreliable; we use the final GT Query frame without search.
All six metrics use the same generated/reference pair, and all models follow
the same rule within each task. Appendix~\ref{app:reference_selection} gives
the search bounds, matching algorithm, and fallback rule.
\begin{revision}
For comparisons across Seen and Novel tasks, Appendix~\ref{app:fixed_endpoint} provides a common fixed-endpoint reference rule; Appendix~\ref{app:rev_translation} adds translated Seen controls and viewpoint-angle analysis.
\end{revision}

\paragraph{Metrics.}
We measure three aspects of recovery from this frame pair.
\emph{Subject Consistency} assesses agreement in scene content using DINOv2
CLS cosine similarity~\citep{oquab2023dinov2}.
\emph{Appearance Fidelity} uses LPIPS, SSIM, and PSNR to measure perceptual
and image-level differences~\citep{Zhang_2018_CVPR,wang2004ssim}.
For \emph{Geometric Consistency}, Depth Anything 3 estimates depth
independently for each image~\citep{lin2025depth3recoveringvisual}; SI-AbsRel measures
depth agreement after scale fitting, while normal cosine measures agreement
between the derived surface orientations. These geometry measures are
monocular-depth proxies, not metric 3D reconstruction accuracy.
Appendix~\ref{app:gt_metric_details} details metric formulas and preprocessing.

\subsubsection{Full Rollout}
\label{sec:free_rollout}

\paragraph{Evaluation Procedure.}
Each model generates a complete interaction from an initial image, a scene
prompt, and a reference action sequence through its native control interface.
We compare the first decoded frame $A$ with the revisit frame $B$, fixed before
scoring. Since the revisit may change the viewpoint, recovery is assessed under
the view relation specified by the revisit type. Novel-view cases revisit
overlapping regions of the same scene, so shared structures and their spatial
relations provide evidence of recovery across viewpoints. Gemini 3.8 Flash
receives this required view relation and the two frames for dimension scoring;
GT novel-view images are not provided.

A separate VQA check screens for recognizable camera motion before scoring.
Only passing videos are scored, but the check does not require the prescribed
direction, exact angle, or trajectory completion. It therefore screens for
motion presence rather than verifying a successful revisit. Intermediate
frames are used only for this check, not for the three dimension scores.
Appendix~\ref{app:complete_vqa_prompts} gives the sampling settings, judge inputs,
and manual review of the evaluated videos.

\begin{figure}[!t]
  \centering
  \includegraphics[
    width=\textwidth,
    trim=65bp 28bp 92bp 41bp,
    clip
  ]{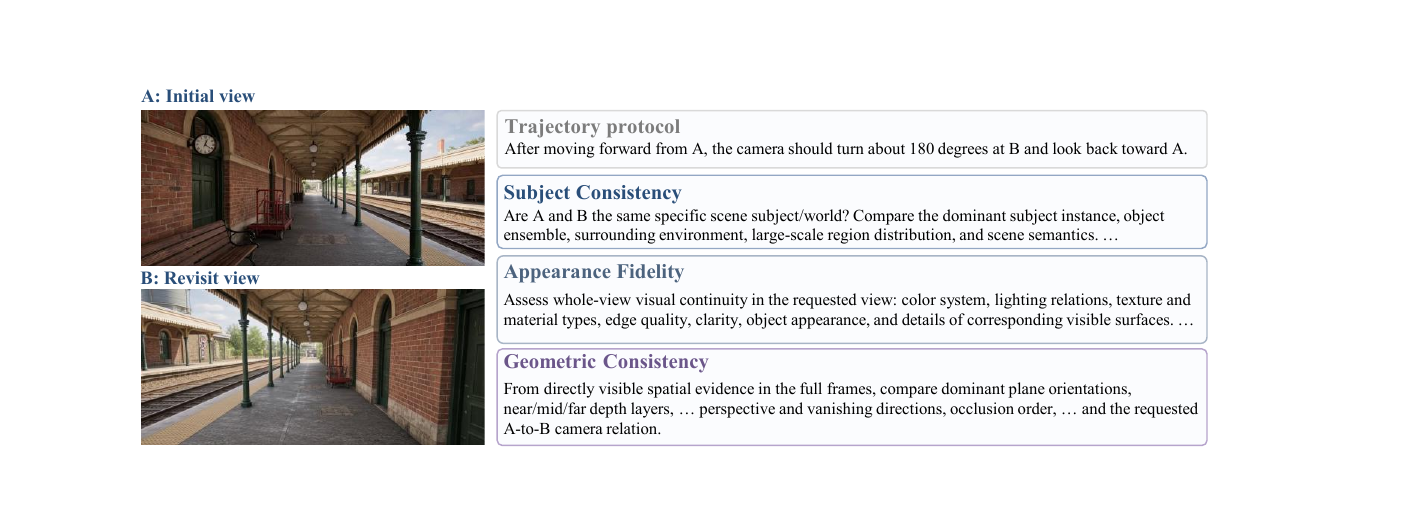}
  \caption{\textbf{Two-frame VQA for Full Rollout.} The initial view $A$ and
  revisit view $B$ are scored from $0$ to $1$ under the required view relation
  using three independent dimension questions.}
  \label{fig:full_rollout_vqa}
\end{figure}

\paragraph{Metrics.}
Three independent VQA questions assess recovery from the frame pair.
\emph{Subject Consistency} asks whether the frames depict the same specific scene.
\emph{Appearance Fidelity} assesses colors, materials, and visual details,
while \emph{Geometric Consistency} assesses layout, depth
ordering, perspective, and occlusion under the required view relation.
For novel-view queries, changes consistent with the requested viewpoint are
allowed. A wrong viewing direction does not automatically negate Subject
Consistency, but recognizing the same scene alone does not establish
appearance or geometry recovery from the required view.
For each dimension, the judge directly assigns a score from $0$ to $1$,
allowing partial correspondence to receive partial credit. We average these
scores over cases in each revisit type and task, reporting the three
dimensions separately. Figure~\ref{fig:full_rollout_vqa} illustrates the input
pair, required view relation, and dimension questions;
Appendix~\ref{app:full_vqa_details} gives the scoring guidance and prompt details.

%% file: sections/04_experiments.tex
\section{Experiments}
\label{sec:experiments}

Our experiments assess the memory capabilities of video world models through
their ability to recover previously observed scenes when revisiting
seen or novel viewpoints under both Continued Rollout and Full Rollout protocols.
We first describe the experimental setup
(Section~\ref{sec:experimental_setup}).
We then present the main results (Section~\ref{sec:main_results}),
comparing scene recovery under a shared ground-truth history in
Continued Rollout with recovery over complete interaction sequences in
Full Rollout, and examining failure cases across Subject Consistency,
Appearance Fidelity, and Geometric Consistency.

\subsection{Experimental Setup}
\label{sec:experimental_setup}

\paragraph{Testing Models.}
We evaluate six open-weight video world models---HY-WorldPlay, LingBot-World v2,
Matrix-Game 3, Matrix-Game 3.5, SANA-WM Streaming, and Zing-World
0.5---in both evaluation protocols, and additionally evaluate the closed
Genie 3 and Happy Oyster web services in Full Rollout. The six open-weight
models use the same checkpoints across protocols; model and inference
details are provided in Appendix~\ref{app:model_adapters}.

\paragraph{Evaluation Settings.}
Continued Rollout uses all 600 controlled cases per model with fixed
context/query boundaries. Full Rollout uses all 60
image-anchored cases, shared across all eight models with 15 per revisit type.
The two protocols yield 3,600 and 480 results, respectively.
Execution details appear in Appendices~\ref{app:model_adapters} and~\ref{app:full_vqa_details}.
To support interpretation of the benchmark scores, we examine history
dependence, uncertainty in model comparisons, and evaluator reliability.
We first check continuation quality, motion response, and scene recovery
on two Reversed Rotation cases (Appendix~\ref{app:gt_continuation_validation}).
History interventions on one case further support the models' dependence on
early observations (Appendix~\ref{app:history_dependence}).
\rev{Appendix~\ref{app:rev_history} extends this analysis to 48 stratified trajectories under Full, Recent-only, and No-history conditions.}
We report paired confidence intervals to quantify uncertainty in score
differences between models (Appendix~\ref{app:paired_ci};
Table~\ref{tab:full_paired_ci}). Human assessments support the VLM as an
auxiliary evaluator of relative preferences, although agreement is lower
on some novel-view tasks (Appendix~\ref{app:human_preference_agreement}).
\begin{revision}
Human preferences provide an additional validation of Continued Rollout metrics (Appendix~\ref{app:rev_human}).
\end{revision}

\begin{revision}
\paragraph{Additional validation.}
Appendix~\ref{app:revision_experiments} provides additional validation: 80 translated Seen controls and viewpoint-angle analysis (\ref{app:rev_translation}), a 48-case history ablation (\ref{app:rev_history}), human preference validation of Continued Rollout (\ref{app:rev_human}), evaluation on 25 real captures (\ref{app:rev_real}), and explicit 3D-memory analysis (\ref{app:rev_memory}). These supplement the original 600 controlled trajectories and 60 Full Rollout cases.
\end{revision}

\subsection{Main Results}
\label{sec:main_results}
\begingroup

Table~\ref{tab:combined_results} summarizes SceneMemBench results for
Return to a Seen View and Return to a Novel View under both Continued Rollout and Full Rollout evaluations.
The Continued Rollout block compares six open-weight models using
six metrics spanning Subject Consistency, Appearance Fidelity, and
Geometric Consistency.
The Full Rollout block extends the comparison to two closed-source models and
reports scores for the same three dimensions after complete
interaction sequences.
\rev{The main table compares models within each task; comparisons of Seen and Novel recovery difficulty use the common fixed-endpoint results in Appendix~\ref{app:fixed_endpoint}.}
Figure~\ref{fig:gt_anchored_analysis} illustrates
novel-view failures, including replaced landmarks and spurious
occluders despite a familiar overall appearance.
These examples show that visual resemblance alone does not establish
the recovery of scene content and spatial structure.
\begin{revision}
With fixed endpoints, the six-model means on all 80 translated Seen controls outperform Novel on all six metrics. See Appendix~\ref{app:rev_translation} for the translated Seen results and viewpoint-angle analysis.
\end{revision}


\input{tables/table2_combined_results}

\begin{figure}[!t]
  \centering
  \includegraphics[width=\textwidth]{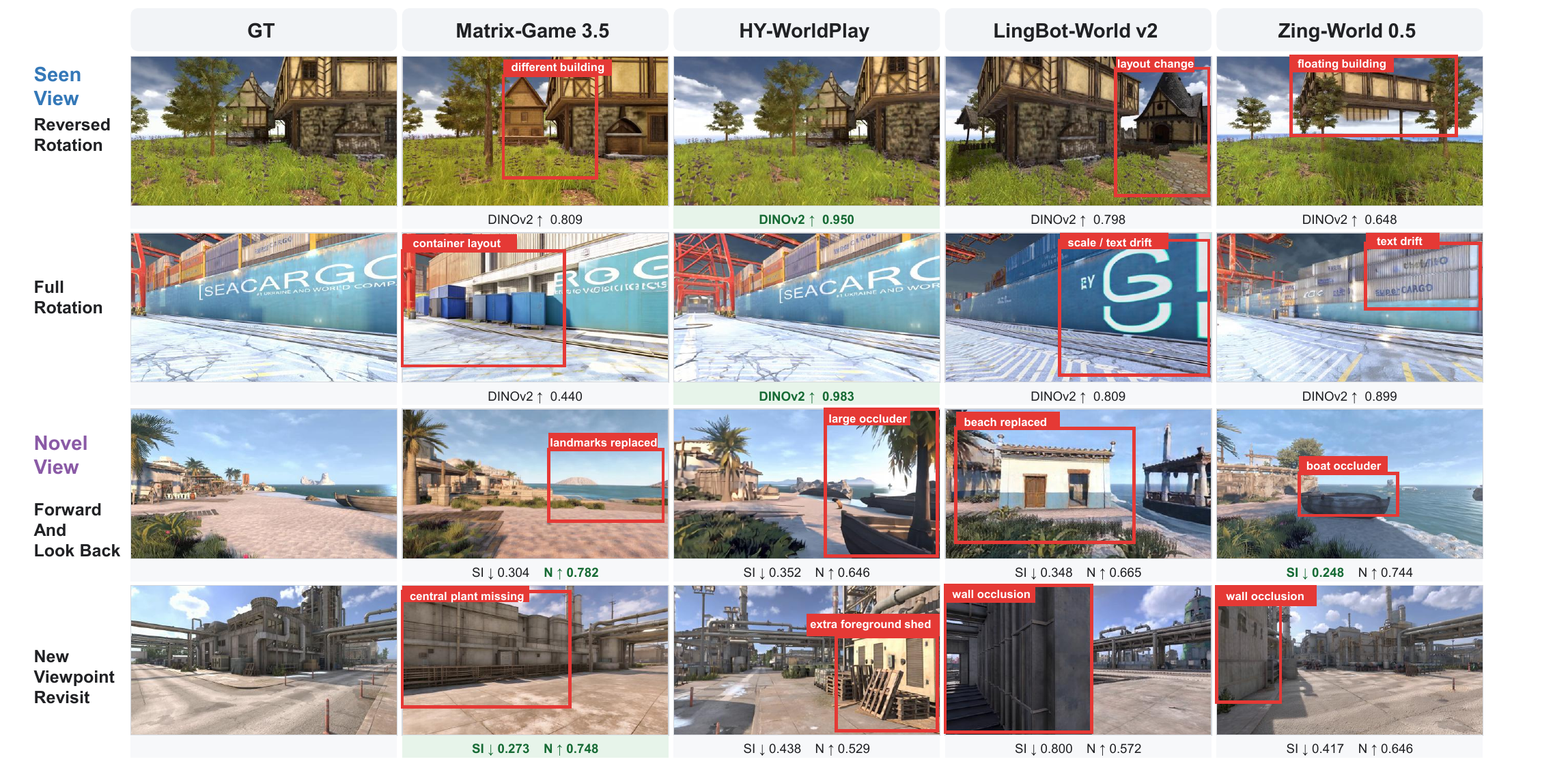}
  \caption{\textbf{Continued Rollout examples across all four revisit types.}
  To highlight Subject Consistency in Seen views and Geometric Consistency in
  Novel views, badges show DINOv2 and SI-AbsRel/normal cosine, respectively;
  green bold values and
  red boxes mark the best displayed results and visible errors.}
  \label{fig:gt_anchored_analysis}
\end{figure}

\begin{figure}[!t]
  \centering
  \includegraphics[width=\textwidth]{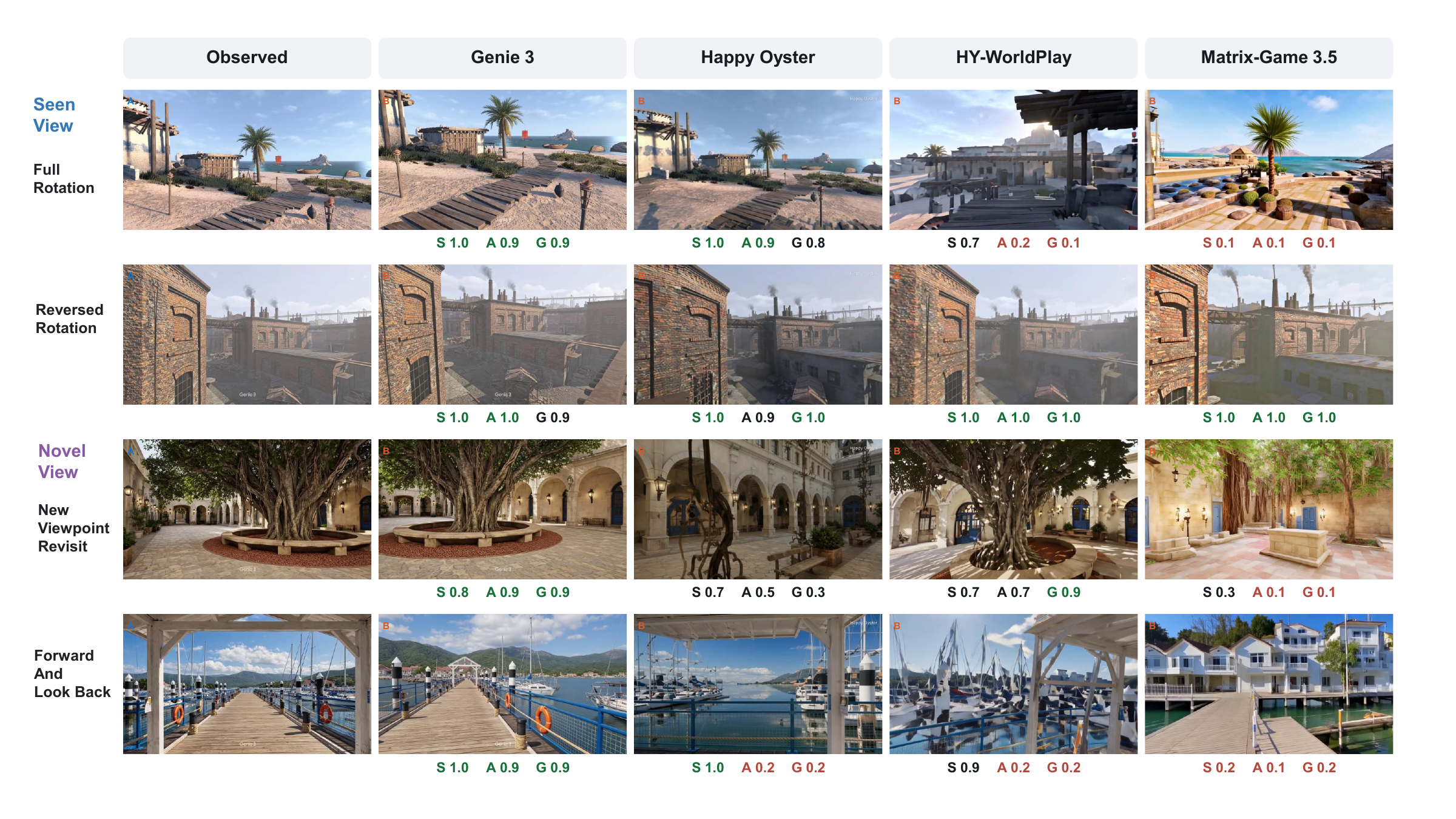}
  \caption{\textbf{Full Rollout examples across all four revisit types.}
  Rows compare the initial observation with four models' fixed-endpoint
  revisits. S/A/G are per-case Gemini 3.8 Flash scores for Subject Consistency,
  Appearance Fidelity, and Geometric Consistency.}
  \label{fig:free_rollout_analysis}
\end{figure}

\begin{figure}[!t]
  \centering
  \includegraphics[width=\textwidth,trim=0pt 235pt 0pt 35pt,clip]{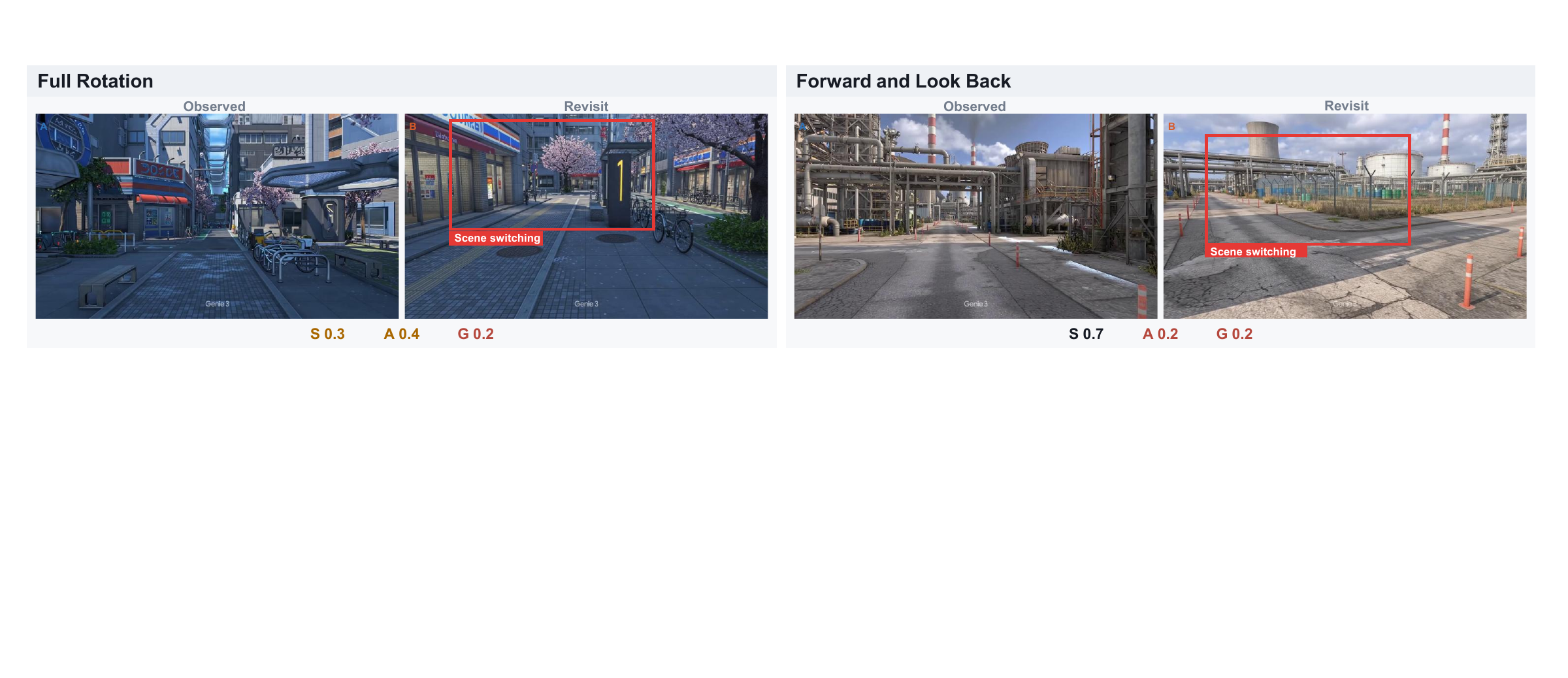}
  \caption{\textbf{Selected Genie 3 failures under Full Rollout.}
  The two examples pair the initial observation with its fixed-endpoint
  revisit; red boxes mark scene switching. S/A/G are per-case Gemini 3.8
  Flash scores for Subject Consistency, Appearance Fidelity, and Geometric
  Consistency.}
  \label{fig:genie3_failure_cases}
\end{figure}

Continued Rollout compares models with different memory designs.
HY-WorldPlay and Matrix-Game 3 retrieve frames by camera geometry and view
overlap; HY-WorldPlay also uses temporal reframing~\citep{sun2026worldplay,wang2026matrixgame30realtimestreaming}.
Matrix-Game 3.5 retrieves and aligns geometry-aware spatial patches~\citep{qian2026matrixgame35enhancingrealtime}.
\rev{Appendix~\ref{app:rev_memory} visualizes its retrieved historical views and target-view mosaic memory.}
LingBot-World v2 manages KV caches dynamically~\citep{gao2026infiniteworldsversatileinteractions};
Zing-World 0.5 uses a bounded cache with a prefix sink and sliding window~\citep{chen2026zing05playableworldsrealtime};
SANA-WM combines Gated DeltaNet states and softmax attention~\citep{zhu2026sanawmefficientminutescaleworld}.
Under the same GT history,
HY-WorldPlay leads five Seen means; Matrix-Game 3 leads Novel Subject Consistency
and LPIPS by mean score, and Matrix-Game 3.5 leads the other four Novel means.
Paired confidence intervals support several of these metric-specific advantages;
full comparisons for both protocols appear in Appendix~\ref{app:paired_ci}
and Table~\ref{tab:full_paired_ci}.

During Full Rollout, trajectory-following errors and accumulated generation
degradation may obscure the strengths of different memory designs in recovering
scene content, appearance, and spatial structure. Matrix-Game 3.5 provides a
clear example of the difference between the two settings: it achieves the best
mean values on both Novel geometry proxies under Continued Rollout, yet obtains
low Geometric Consistency scores under Full Rollout and exhibits structural
errors (Figure~\ref{fig:free_rollout_analysis}). Its performance illustrates why
recovery from a shared GT history should be examined alongside complete interaction.



Although Genie 3 achieves the highest mean Full Rollout scores across
all the evaluation dimensions
(Table~\ref{tab:combined_results}), we still find some failure cases as shown in
Figure~\ref{fig:genie3_failure_cases} which show significant differences between initial and final frames in both scene content
and layout.
These failures illustrate that recognizable scene elements can
coexist with substantial discrepancies in appearance and spatial
structure, underscoring the need to evaluate Subject Consistency
jointly with Appearance Fidelity and Geometric Consistency.

\endgroup

%% file: tables/table2_combined_results.tex
\begin{table}[!t]
  \caption{\textbf{SceneMemBench results under Continued Rollout and Full Rollout.}
  Entries are per-model task means. \textbf{Bold} and \underline{underline}
  mark the best and second-best means within each task and protocol, not
  statistical significance. Continued Rollout paired intervals:
  Appendix~\ref{app:paired_ci}.}
  \label{tab:combined_results}
  \label{tab:gt_anchored_results}
  \label{tab:free_rollout_results}
  \centering
  \scriptsize
  \setlength{\tabcolsep}{4pt}
  \renewcommand{\arraystretch}{1.00}
  \begin{tabular}{@{}lrrrrrr@{}}
    \toprule
    \multicolumn{7}{l}{\small\bfseries Continued Rollout} \\
    \addlinespace[2pt]
    & \multicolumn{1}{c}{Subject Consistency}
    & \multicolumn{3}{c}{Appearance Fidelity}
    & \multicolumn{2}{c}{Geometric Consistency} \\
    \cmidrule(lr){2-2}\cmidrule(lr){3-5}\cmidrule(lr){6-7}
    Model & DINOv2$\uparrow$ & LPIPS$\downarrow$ & SSIM$\uparrow$
      & PSNR$\uparrow$ & SI-AbsRel$\downarrow$ & Normal Cos.$\uparrow$ \\
    \midrule
    \multicolumn{7}{l}{\textit{Return to a Seen View}} \\
    \midrule
    HY-WorldPlay
      & \textbf{0.9290} & \textbf{0.2775} & \underline{0.4339}
      & \textbf{16.8042} & \textbf{0.2057} & \textbf{0.8227} \\
    LingBot-World V2
      & 0.7749 & 0.5128 & 0.3286 & 13.2291 & 0.3998 & 0.7021 \\
    Matrix-Game 3
      & \underline{0.8840} & 0.4182 & 0.3631 & 14.8309 & 0.2946
      & 0.7554 \\
    Matrix-Game 3.5
      & 0.7690 & \underline{0.3501} & \textbf{0.4367}
      & \underline{16.3734} & \underline{0.2678} & \underline{0.7979} \\
    SANA-WM Streaming
      & 0.6245 & 0.4665 & 0.3200 & 14.4377 & 0.3638 & 0.7522 \\
    Zing-World 0.5
      & 0.8007 & 0.4688 & 0.3549 & 14.1265 & 0.4544 & 0.7298 \\
    \midrule
    \multicolumn{7}{l}{\textit{Return to a Novel View}} \\
    \midrule
    HY-WorldPlay
      & \underline{0.6987} & 0.5401 & \underline{0.3410}
      & 13.5762 & \underline{0.4622} & 0.6809 \\
    LingBot-World V2
      & 0.5569 & 0.5834 & 0.3026 & 12.4137 & 0.6246 & 0.6369 \\
    Matrix-Game 3
      & \textbf{0.7543} & \textbf{0.5124} & 0.3394
      & \underline{13.6729} & 0.5130 & \underline{0.7004} \\
    Matrix-Game 3.5
      & 0.5694 & \underline{0.5141} & \textbf{0.3674}
      & \textbf{14.3759} & \textbf{0.4368} & \textbf{0.7079} \\
    SANA-WM Streaming
      & 0.3318 & 0.6699 & 0.2387 & 11.5343 & 0.8050 & 0.5325 \\
    Zing-World 0.5
      & 0.6633 & 0.5630 & 0.3349 & 13.4154 & 0.6102 & 0.6795 \\
    \addlinespace[1pt]
    \midrule
    {\small\bfseries Full Rollout}
    & \multicolumn{3}{c}{Return to a Seen View}
    & \multicolumn{3}{c}{Return to a Novel View} \\
    \cmidrule(lr){2-4}\cmidrule(lr){5-7}
    Model & Subject$\uparrow$ & Appearance$\uparrow$ & Geometry$\uparrow$
      & Subject$\uparrow$ & Appearance$\uparrow$ & Geometry$\uparrow$ \\
    \midrule
    \multicolumn{7}{l}{\textit{Closed-source models}} \\
    \midrule
    Genie 3
      & \textbf{0.947} & \textbf{0.853} & \textbf{0.820}
      & \textbf{0.770} & \textbf{0.650} & \textbf{0.520} \\
    Happy Oyster
      & 0.833 & 0.527 & 0.517
      & 0.637 & 0.437 & 0.343 \\
    \midrule
    \multicolumn{7}{l}{\textit{Open-source models}} \\
    \midrule
    HY-WorldPlay
      & \underline{0.907} & \underline{0.677} & \underline{0.680}
      & \underline{0.717} & 0.433 & \underline{0.383} \\
    LingBot-World-v2
      & 0.390 & 0.227 & 0.183
      & 0.413 & 0.223 & 0.163 \\
    Matrix-Game 3
      & 0.857 & 0.607 & 0.590
      & 0.633 & \underline{0.473} & 0.367 \\
    Matrix-Game 3.5
      & 0.683 & 0.503 & 0.533
      & 0.243 & 0.130 & 0.130 \\
    SANA-WM
      & 0.267 & 0.150 & 0.133
      & 0.337 & 0.183 & 0.160 \\
    Zing-World 0.5
      & 0.193 & 0.163 & 0.143
      & 0.213 & 0.093 & 0.067 \\
    \bottomrule
  \end{tabular}
\end{table}

%% file: sections/05_conclusion.tex
\section{Conclusion}
\label{sec:conclusion}


We introduced \textsc{SceneMemBench}, a benchmark for evaluating scene
memory in interactive video world models through seen-view recall and
novel-view recovery.
Its two tasks, \emph{Return to a Seen View} and \emph{Return to a Novel View},
assess whether models can recover previously observed scene content
with both previously seen and unobserved camera poses.
We introduce Continued Rollout, which supports controlled diagnosis from a ground-truth
observation history, while Full Rollout evaluates revisit scenes after the complete
model-generated interactions.
Across eight representative models, our evaluation shows that \rev{novel-view recovery is generally more challenging than seen-view recall}, and that model
profiles under controlled histories can differ markedly from those
observed in full interaction. These findings underscore the need to evaluate both recovery tasks under
both protocols: each exposes complementary strengths and failure modes
that a single task or protocol can obscure.

%% file: sections/06_statements.tex

\section*{AI Use Statement}
Large language models assisted with writing, translation, and result analysis
and interpretation; coding agents assisted with code development.
The authors reviewed these outputs and tested the assisted code, and take
full responsibility for the paper. Models evaluated and used as judges
are described in Sections~\ref{sec:free_rollout} and~\ref{sec:experimental_setup}.

\section*{Ethics Statement}
\rev{Our UE data were collected using UnrealZoo, with the required permissions for asset use and data release. The real-world videos contain no identifiable faces, license plates, or other personal information, and all necessary permissions and consent for recording and release have been obtained. We will release the dataset videos, camera poses, and source code under the confirmed licenses. Genie 3 and Happy Oyster outputs were obtained through their official online services, in accordance with the applicable terms of service.}
VLM judgments can be biased; reported memory scores do not establish safe or reliable deployment in real environments.

\section*{Reproducibility Statement}
Sections~\ref{sec:scenemembench}--\ref{sec:experiments} describe tasks,
protocols, and experimental settings. \rev{Appendices~\ref{app:dataset_protocol_metric_details}--\ref{app:revision_experiments}}
provide dataset statistics, model adaptations, metric formulas, scoring
prompts, and validation results.
The project page is available at \url{https://yuheng-si.github.io/SceneMemBench/}.

%% file: sections/appendix.tex
\section{Dataset and Trajectory Design Details}
\label{app:dataset_protocol_metric_details}
\subsection{Dataset Coverage and Capture}
The capture and trajectory specifications below define the observation
histories and revisit queries used in the benchmark.
SceneMemBench contains 600 controlled trajectories across 33 scenes and
60 image-anchored cases, covering indoor, urban, industrial, and natural
environments. We use UnrealZoo~\citep{Zhong_2025_ICCV} as the capture
platform for the 600 GT video trajectories. Tables~\ref{tab:appendix_dataset_statistics}
and~\ref{tab:appendix_scene_coverage} summarize capture specifications,
annotations, and per-scene case counts. The image-anchored cases overlap the controlled corpus in scenes and initial views. Capture fixes lighting and rendering settings,
synchronizes RGB and camera poses, and checks trajectory motion and video integrity.

\input{tables/appendix_dataset_statistics}
\input{tables/appendix_scene_coverage}

\subsection{Trajectory Construction and Query Boundaries}
Table~\ref{tab:trajectory_design} summarizes motion, Context coverage, and
Query onset for all four revisit types. $A$ denotes the initial camera position,
$B$ the displaced position, and $P$ the revisited target. Boundaries are fixed
before inference. Durations are computed from captured frame counts and frame
rate, including holds within each interval. Translation distance is horizontal
path length, excluding travel back after the revisit.

\input{tables/appendix_trajectory_design}

New-Viewpoint Revisit uses horizontal $\angle APB$ design bins of
$45^\circ$, $90^\circ$, and $135^\circ$, containing 48, 87, and 65 cases,
respectively. These are target bins; realized angles vary with navigable paths
and target positions. Reversed Rotation includes 80 yaw and 20 pitch trajectories.

\section{Continued Rollout Implementation and Validation}
\label{app:model_adapters}
\subsection{Model Selection and History Initialization}
\label{app:history_initialization}
To compare recovery after a shared observation history, we supply the same
GT Context through each model's native memory mechanism.
GT Context is encoded into each model's native history representation, then
used for continuation through its cache, retrieval, or history-replay mechanism.
Table~\ref{tab:history_mechanisms} summarizes history initialization and use.
The adapters retain native encoding, denoising, and history-update mechanisms,
incorporating newly generated content into the history.
Context is resampled to the native resolution and frame rate; the final source
frame is preserved and repeated only when required by VAE or chunk cadence.
Camera motion uses the model's supported pose or action interface.

\input{tables/appendix_history_read_evidence}

\subsection{Inference Settings}
\label{app:shared_inference}
Across both protocols, the six open-weight models share released distilled
checkpoints, seed 42, and native output resolutions and frame rates.
Experiments use six NVIDIA A800 80~GB GPUs.

\subsection{Evaluation Reference Selection}
\label{app:reference_selection}
Residual camera-control errors can shift the generated revisit away from
the prescribed GT endpoint. Local reference selection aims to reduce this
view mismatch when evaluating recovery of a seen view.
For Return to a Seen View, candidate GT frames $r_i$ are restricted to a local
angular window around the GT endpoint: $\pm30^\circ$ for Full Rotation,
and $30^\circ$ on the approach side for Reversed Rotation, allowing an
incomplete return turn. The generated endpoint $g$ is compared with these
candidates. Mutual SIFT matching and
homography RANSAC identify inliers; $C_g$ and $C_{r_i}$ are their convex-hull
coverage in the respective images. We maximize
\begin{equation}
S_{\mathrm{sel}}(g,r_i)
=
N_{\mathrm{inlier}}(g,r_i)
\min\!\left(C_g,C_{r_i}\right).
\label{eq:gt_reference_selection}
\end{equation}
If the selected pair has fewer than 12 inliers or a minimum coverage below
$0.01$, we use the fixed endpoint reference. Return to a Novel View always
uses the final decoded GT Query frame, without content-adaptive search.
Matching does not verify the generated camera pose. All six metrics share the
frame pair, and all models share each task's rule. No subsequent warping, pose
alignment, temporal interpolation, or content-adaptive cropping is applied;
the selection score is not a reported memory metric.

\subsection{Validation of Continued Rollout}
\label{app:gt_continuation_validation}

We check whether initialization from GT history supports usable,
action-responsive continuations and recovery of earlier scene content.
For this check, we compare native Full Rollout returns (A) with GT-history continuations (B)
for six models on two Reversed Rotation cases, assessing quality, motion, and
scene recovery.

\begin{figure}[!p]
  \centering
  \begingroup
  \def\validationearly{Early view}
  \def\validationboundary{Before return}
  \def\validationreturn{Generated return}
  \input{figures/figure7/panels}
  \endgroup
  \caption{\textbf{Six-model comparison on one Reversed Rotation case.}
  Each model has a Full Rollout row followed by a Continued Rollout row.
  Columns show the initial frame, the frame before return control, and the
  final output; in Continued Rollout, the first two columns are the early
  and actual last GT-history frames. The platform reappears in all six
  continuations. Endpoints are fixed last frames, not similarity-selected;
  rollout lengths and histories differ. The table uses two cases and the
  first 2 seconds, whereas this figure shows one case at its final endpoints.}
  \label{fig:appendix_gt_continuation}
\end{figure}
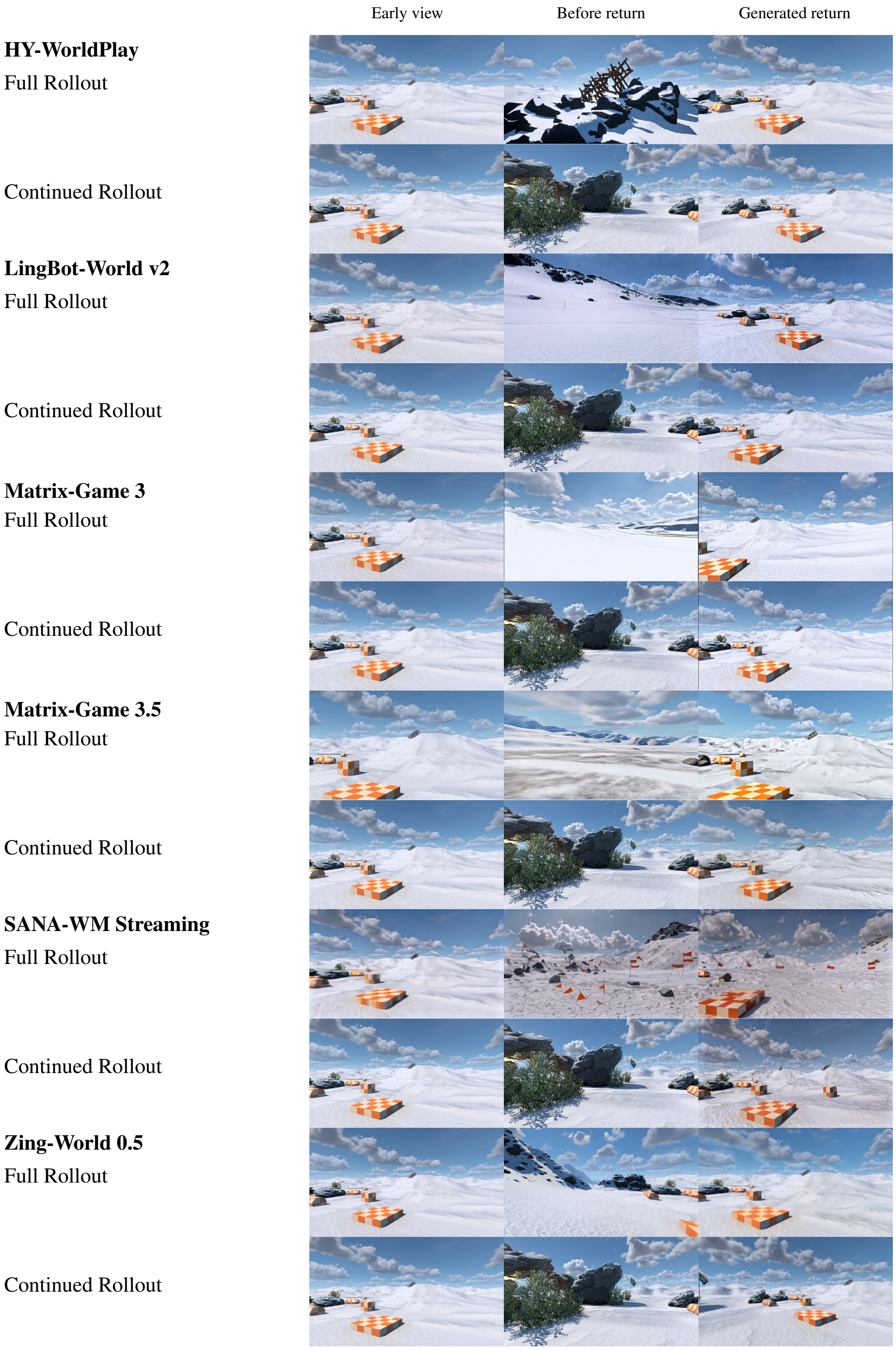

\subsubsection{Generation Quality and Motion Response}
We score the first 2 generated seconds, excluding GT history. Quality uses
MUSIQ-SPAQ means and percentile 10 (P10); smoothness is one minus normalized
AMT-S interpolation error, both at a maximum edge of 512 pixels and native
frame rates. Flow at 8 Hz measures the fraction of moving pairs and the fraction
of moving pairs with correct direction. Table~\ref{tab:continuation_fidelity}
 equally averages both cases: continuation quality is broadly comparable to
native generation, and all six models attain 100\% on both motion measures,
supporting usable, action-responsive short continuations.

\input{tables/appendix_continuation_fidelity}

\subsubsection{Recovery of Earlier Scene Structure}
Figure~\ref{fig:appendix_gt_continuation} compares all six models on one snow-scene case.
The previously observed checkerboard platform is outside the view when continuation
begins and reappears in all six returns. History-selection records collected during generation show that HY-WorldPlay and
Matrix-Game 3 use GT-history latents retrieved by field-of-view (FOV) overlap,
while Matrix-Game 3.5 performs patch queries and selects GT-sourced context.
The remaining models use the cache or history-replay mechanisms in
Table~\ref{tab:history_mechanisms}. In a Full Rotation example, HY-WorldPlay retrieves a historical
frame from the $0^\circ$ view when returning to the initial viewpoint after
a $360^\circ$ rotation.

\paragraph{Sensitivity to earlier history.}
\label{app:history_dependence}

\begin{figure}[!p]
  \centering
  \includegraphics[width=0.90\textwidth]{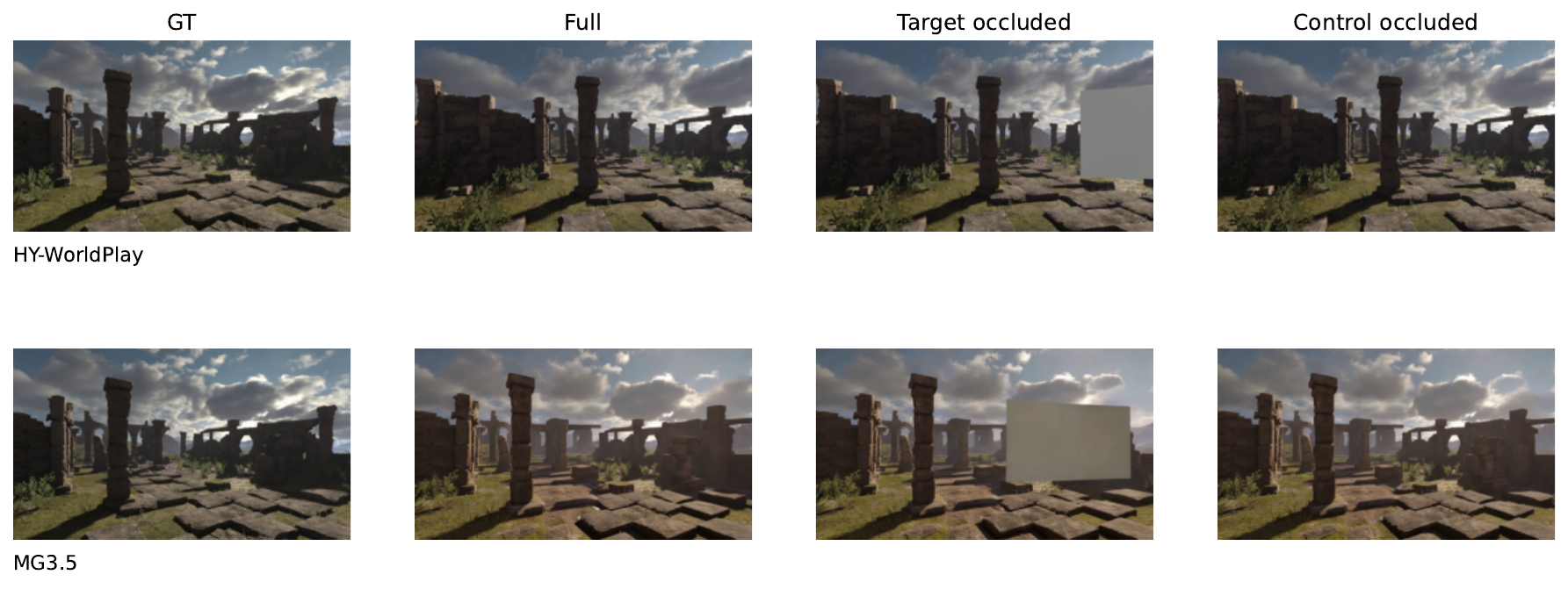}
  \caption{\textbf{History interventions with a shared initial frame.}
  All branches start from the same target-occluded image. Full provides
  original target evidence after that frame; Target occludes the early target
  region, while Control occludes an equal-area non-target region.
  The recovered opening in Full and Control, and gray region in Target,
  support a role for subsequent early observations.}
  \label{fig:history_common_anchor}
\end{figure}

To test whether recovered content depends on earlier observations, we
intervene on the history supplied to the models.
On one representative Reversed Rotation case, we compare
full history, genuine truncation of early history, and replacement of early
target observations, while holding recent RGB context and Query controls fixed
across conditions. Each condition rebuilds its states independently, with
effective Query noise paired within each model. Target-region scores compare
fixed generated endpoints with the unmodified GT endpoint. A two-model
follow-up uses a shared target-occluded initial frame and an equal-area
non-target occlusion control (Fig.~\ref{fig:history_common_anchor}).

\begin{figure}[!p]
  \centering
  \includegraphics[width=0.90\textwidth]{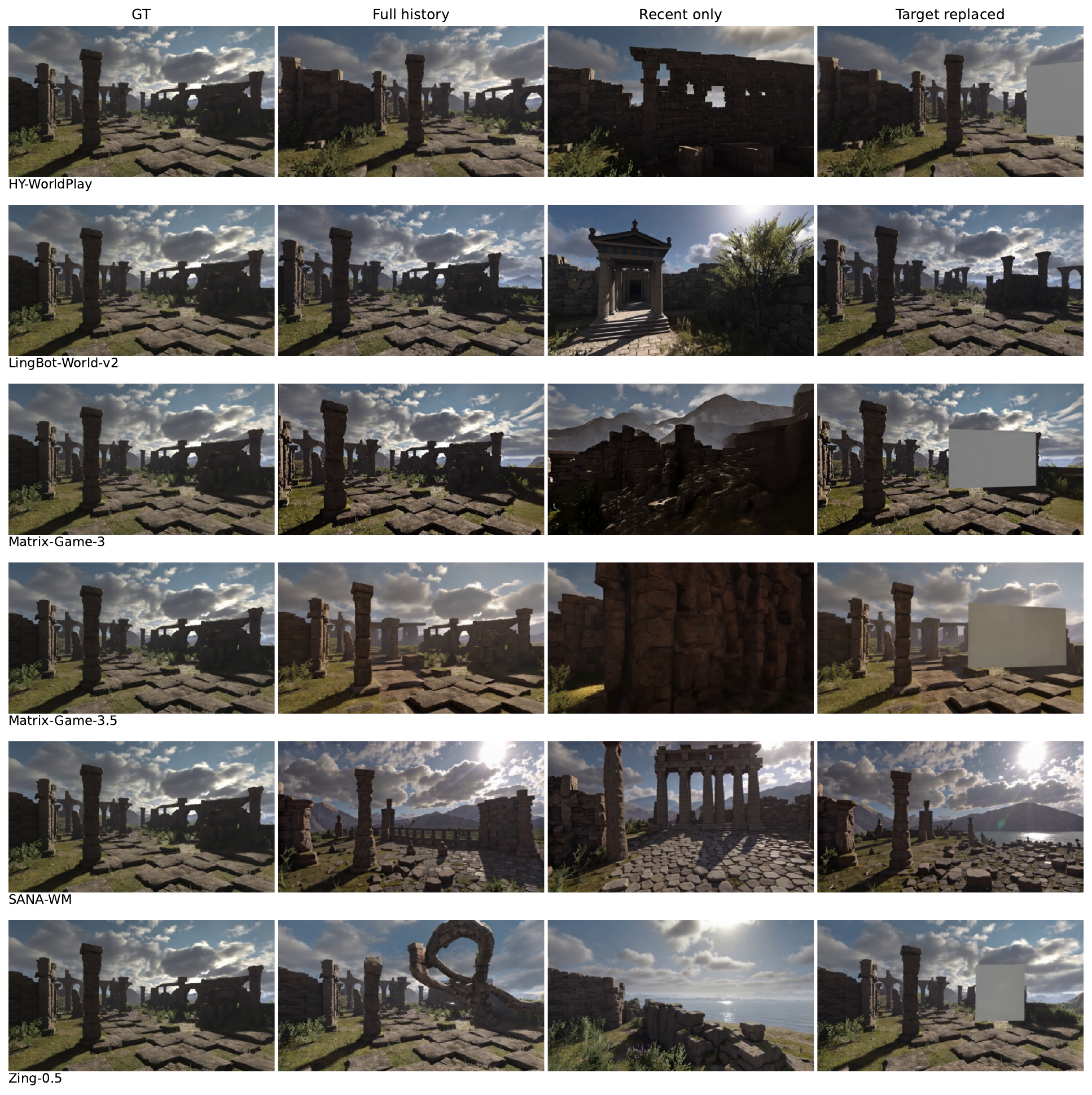}
  \caption{\textbf{Six-model history interventions on one Reversed Rotation case.}
  Columns show GT, Full history, Recent only, and Target replaced.
  Recent RGB inputs, Query controls, and effective Query noise are paired
  within each model. All conditions rebuild their states independently.
  Images are fixed generated endpoints; gray regions in the replacement
  condition can reflect the altered history rather than forgetting.}
  \label{fig:history_dependence}
\end{figure}

Truncating early history worsens target-region LPIPS for all six models and
lowers DINOv2 similarity for five (Table~\ref{tab:history_dependence}).
Replacing early target observations also changes recovered content and scores
(Fig.~\ref{fig:history_dependence}). LingBot-World v2 changes the distant wall
and its opening; despite a slight improvement in ROI LPIPS, this content
change and lower ROI DINOv2 support history sensitivity. The shared-initial-frame
experiment with HY-WorldPlay and Matrix-Game 3.5 further shows that early
evidence after the initial frame affects continuation. These results provide
direct intervention evidence for history dependence in Continued Rollout,
supporting its validity as a memory diagnostic on this case.

\input{tables/appendix_history_dependence}

\section{Evaluation Metric Formulas}
\label{app:gt_metric_details}
The formulas and preprocessing below specify how Continued Rollout scores
the three recovery dimensions from a shared image pair.
All metrics compare the generated endpoint $g$ with the GT reference $r$
selected in Appendix~\ref{app:reference_selection}. Image metrics use Lanczos
resizing and center cropping to $W\times H=448\times252$, with RGB in $[0,1]$.

\subsection{Subject Consistency}
Let $f$ denote the final-layer DINOv2 CLS feature, using ImageNet channel-mean
and standard-deviation input normalization~\citep{oquab2023dinov2}.
\begin{equation}
S_{\mathrm{subj}}(g,r)=\frac{f(g)^{\mathsf T}f(r)}{\lVert f(g)\rVert_2\lVert f(r)\rVert_2}.
\label{eq:scene_identity}
\end{equation}

\subsection{Appearance Fidelity}
LPIPS uses AlexNet features and learned channel weights~\citep{Zhang_2018_CVPR}:
\begin{equation}
D_{\mathrm{LPIPS}}(g,r)=\sum_l\frac{1}{H_lW_l}\sum_{h,w}
\left\|w_l\odot\bigl(\hat f_l(g)_{hw}-\hat f_l(r)_{hw}\bigr)\right\|_2^2.
\label{eq:lpips}
\end{equation}
Here $\hat f_l$ is the channel-normalized feature at layer $l$, $H_l,W_l$
are its spatial dimensions, $w_l$ contains channel weights, and $\odot$ is
elementwise multiplication. Network inputs are mapped from $[0,1]$ to $[-1,1]$.
SSIM averages local-window scores over positions and RGB channels~\citep{wang2004ssim}:
\begin{equation}
\operatorname{SSIM}(g,r)=\operatorname{mean}_{u,c}
\frac{(2\mu_g\mu_r+C_1)(2\sigma_{gr}+C_2)}
{(\mu_g^2+\mu_r^2+C_1)(\sigma_g^2+\sigma_r^2+C_2)}.
\label{eq:ssim}
\end{equation}
The window means, variances, and covariance are $\mu,\sigma^2,\sigma_{gr}$;
$u,c$ index window positions and channels. We use an $11\times11$ Gaussian
window with standard deviation $1.5$, $C_1=0.01^2$, and $C_2=0.03^2$, retaining
scikit-image's sample-covariance and border-cropping conventions. PSNR is
\begin{equation}
\operatorname{MSE}=\frac{1}{3HW}\sum_{u,c}(g_{uc}-r_{uc})^2,\qquad
\operatorname{PSNR}=-10\log_{10}\max(\operatorname{MSE},10^{-10}).
\label{eq:psnr}
\end{equation}

\subsection{Geometric Consistency}
Depth Anything 3 estimates $d^g,d^r$ independently, without joint-image
inference or pose/scale alignment~\citep{lin2025depth3recoveringvisual}. The reference
is resized to the generated image dimensions, with intrinsics $K$ scaled accordingly.
Define $\mathcal V_x=\{i:d_i^x\text{ is finite and }d_i^x>10^{-4}\}$ for
$x\in\{g,r\}$, and $\mathcal V=\mathcal V_g\cap\mathcal V_r$.
A global scale is fitted from the depth ratios on this intersection:
\begin{equation}
q_i=d_i^r/d_i^g,\qquad
s=\operatorname*{median}_{i\in\mathcal V:\,P_{10}\le q_i\le P_{90}}q_i.
\label{eq:depth_scale}
\end{equation}
$P_{10},P_{90}$ are percentiles 10 and 90 of the ratios; fewer than
10 ratios are used without trimming. Scale-invariant AbsRel assigns unit
error to reference-valid pixels with invalid generated depth:
\begin{equation}
E_{\mathrm{SI}}=\frac{\sum_{i\in\mathcal V}|d_i^r-sd_i^g|/d_i^r
+|\mathcal V_r\setminus\mathcal V|}{|\mathcal V_r|}.
\label{eq:si_absrel}
\end{equation}
The depth score is undefined when $\mathcal V$ is empty.
Camera-space normals use back-projection and central differences:
\begin{equation}
X_{uv}=d_{uv}K^{-1}[u,v,1]^{\mathsf T},\qquad
\tilde n_{uv}=(X_{u+1,v}-X_{u-1,v})\times(X_{u,v+1}-X_{u,v-1}),
\end{equation}
\begin{equation}
S_{\mathrm n}=\frac{1}{|\mathcal N|}\sum_{i\in\mathcal N}(n_i^r)^{\mathsf T}n_i^g.
\label{eq:normal_cosine}
\end{equation}
Normalize $\tilde n$ and orient it toward the camera to obtain the unit normal $n$.
$\mathcal N$ contains only interior pixels where, in both images, the center
and four difference neighbors have valid depth, the normal is finite, and
$\|\tilde n\|_2>10^{-8}$. An empty set yields an undefined score.
These metrics measure agreement between monocular depth predictions,
not ground-truth 3D reconstruction accuracy.

\section{Full Rollout VLM Evaluation}
\label{app:full_vqa_details}
To make Full Rollout scores interpretable, we describe the visual evidence
and scoring instructions supplied to the VLM, then assess its agreement
with human preferences. The evaluated videos are generated under the
native-interface conditions below.
Full Rollout adapts action counts and durations to each model's native interface;
realized paths and endpoints may differ.
The six open-weight models use the shared settings in
Appendix~\ref{app:shared_inference}. Genie 3 and Happy Oyster use native web
interfaces without controllable backend versions or seeds.
Table~\ref{tab:appendix_native_formats} lists output formats.

\input{tables/appendix_native_formats}

\subsection{VQA Prompts and Scoring Details}
\label{app:complete_vqa_prompts}
\paragraph{Judge and visual evidence.}
Gemini 3.8 Flash evaluates all 480 Full Rollout outputs with fixed scoring prompts.
Before scoring, a separate VQA check examines 12 uniformly time-sampled
intermediate frames for recognizable camera motion. All 480 videos passed.
This check does not require a prescribed direction, exact angle, or trajectory completion.
Separately, we visually reviewed each of the 480 videos and found no static
outputs or outputs limited to a small camera excursion followed by a return
to nearly the initial view. This is an observation about the evaluated videos,
not a minimum-motion guarantee of the VQA prompt.
For dimension scoring, the evaluator receives the initial frame $A$, the
revisit frame $B$ fixed before scoring, and their side-by-side composite;
no intermediate video frames or GT novel-view images are supplied.
Novel-view cases target overlapping regions of the same scene. In Forward
and Look Back, the initial and return views observe this region from opposing
directions after a $180^\circ$ turn. Shared objects and scene structures provide
the basis for judging spatial relations across these views; missing or
inconsistent structures are recovery errors.
Subject Consistency, Appearance Fidelity, and Geometric Consistency are
queried independently, without passing answers between dimensions.
Dimension scores are used directly, without human editing or programmatic
caps. We average 15 cases per revisit type and 30 per Task, reporting the
three dimensions separately.

\paragraph{Unscored camera-motion check prompt.}
\input{prompts/motion_check}

\paragraph{Endpoint scoring prompts.}
The twelve questions share the instructions and scale below. Each revisit
type has three dimension questions with a common view requirement.

\input{prompts/vqa_english}

\subsection{Agreement with Human Preferences}
\label{app:human_preference_agreement}
We assess whether the VLM's relative preferences agree with human judgments
of scene recovery, using paired comparisons of the evaluated outputs.

\paragraph{Paired assessment.}
Two annotators jointly assessed all 240 pairs, covering all 480 Full Rollout
outputs, and produced final human consensus labels.
Within each of the 60 trajectory--case groups, the eight model outputs were
partitioned into four disjoint pairs, so each output appeared once.
Pairing balanced model combinations and randomized left/right
positions without consulting Gemini scores. The annotators viewed each candidate's
initial and revisit frames and judged Subject Consistency, Appearance Fidelity,
and Geometric Consistency separately: left preferred (L), right preferred (R),
tie (T), or unable to judge (U). Human labels were locked before Gemini scores
were revealed. Figure~\ref{fig:human_preference_examples} illustrates three annotated pairs
with their original frames and labels; Gemini scores are added for comparison.

\paragraph{Agreement and scope.}
Gemini preferences are derived by comparing the two dimension scores, with
equal scores treated as ties. Agreement requires an exact match with the human
consensus L/R/T label; U labels are excluded for that dimension.
Overall, VLM preferences agree with human consensus on 193/240 (80.4\%) pairs
for Subject Consistency, 185/240 (77.1\%) for Appearance Fidelity, and
177/232 (76.3\%) for Geometric Consistency, with eight geometry U labels
excluded. These results support the VLM as an auxiliary evaluator of relative
preferences. However, agreement is lower on some novel-view tasks
(Table~\ref{tab:human_agreement_groups}), indicating discrepancies between VLM
and human judgments that limit its reliability as a cross-view consistency
evaluator in these conditions. Future work could incorporate intermediate
frames to provide richer visual evidence for cross-view correspondence.

\input{tables/appendix_human_agreement}

\begin{figure}[!t]
  \centering
  \input{figures/figure8/panels}
  \caption{\textbf{Examples of human--Gemini preference assessment.}
  (a, b) Agreement on all three dimensions.
  (c) Disagreement on all three dimensions.}
  \label{fig:human_preference_examples}
\end{figure}

\section{Additional Results and Confidence Intervals}
\label{app:additional_results}
We examine how the main results vary across revisit types, how sensitive
they are to GT reference selection, and how uncertain the observed model
differences are under resampling.
The following tables disaggregate Table~\ref{tab:combined_results} by
revisit type. Each entry is an arithmetic mean over the fixed cases for one
model and revisit type. The main-table Task score equally averages its two
revisit type means: 100 cases per Seen revisit type and 200 per Novel revisit type in
Continued Rollout, and 15 per revisit type in Full Rollout. 

\subsection{Continued Rollout by revisit type}
\input{tables/appendix_continued_seen}
\input{tables/appendix_continued_novel}

In Tables~\ref{tab:appendix_continued_seen} and~\ref{tab:appendix_continued_novel},
HY-WorldPlay has the
highest Subject Consistency in both Seen revisit types, whereas Matrix-Game 3
leads this metric in both Novel revisit types. Matrix-Game 3.5 has the strongest
Novel normal cosine in both revisit types, but its lowest SI-AbsRel is specific
to New-Viewpoint Revisit: HY-WorldPlay has lower SI-AbsRel in Forward and
Look Back.

\paragraph{Sensitivity to the GT reference frame.}
\label{app:fixed_endpoint}
We examine how GT reference selection affects model comparisons within
Return to a Seen View.
Table~\ref{tab:fixed_endpoint} evaluates the same 3,600 outputs using the final
GT Query frame for both tasks, with the same metrics and aggregation as
Table~\ref{tab:combined_results}. Novel scores are unchanged because this
task already uses fixed endpoint references.

\begin{figure}[!t]
  \centering
  \begin{tabular}{@{}c@{\hspace{0.01\linewidth}}c@{\hspace{0.01\linewidth}}c@{}}
  \small HY generated endpoint & \small Matched GT frame & \small Prescribed GT endpoint \\
  \includegraphics[width=0.326\linewidth]{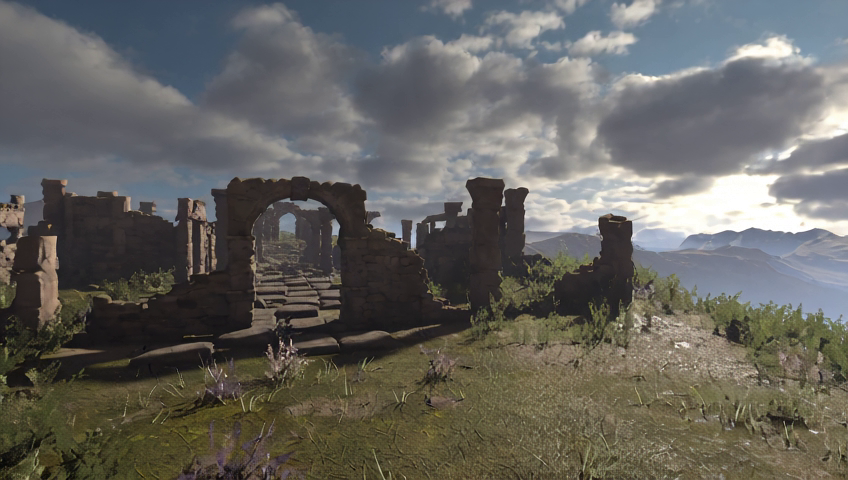} &
  \includegraphics[width=0.326\linewidth]{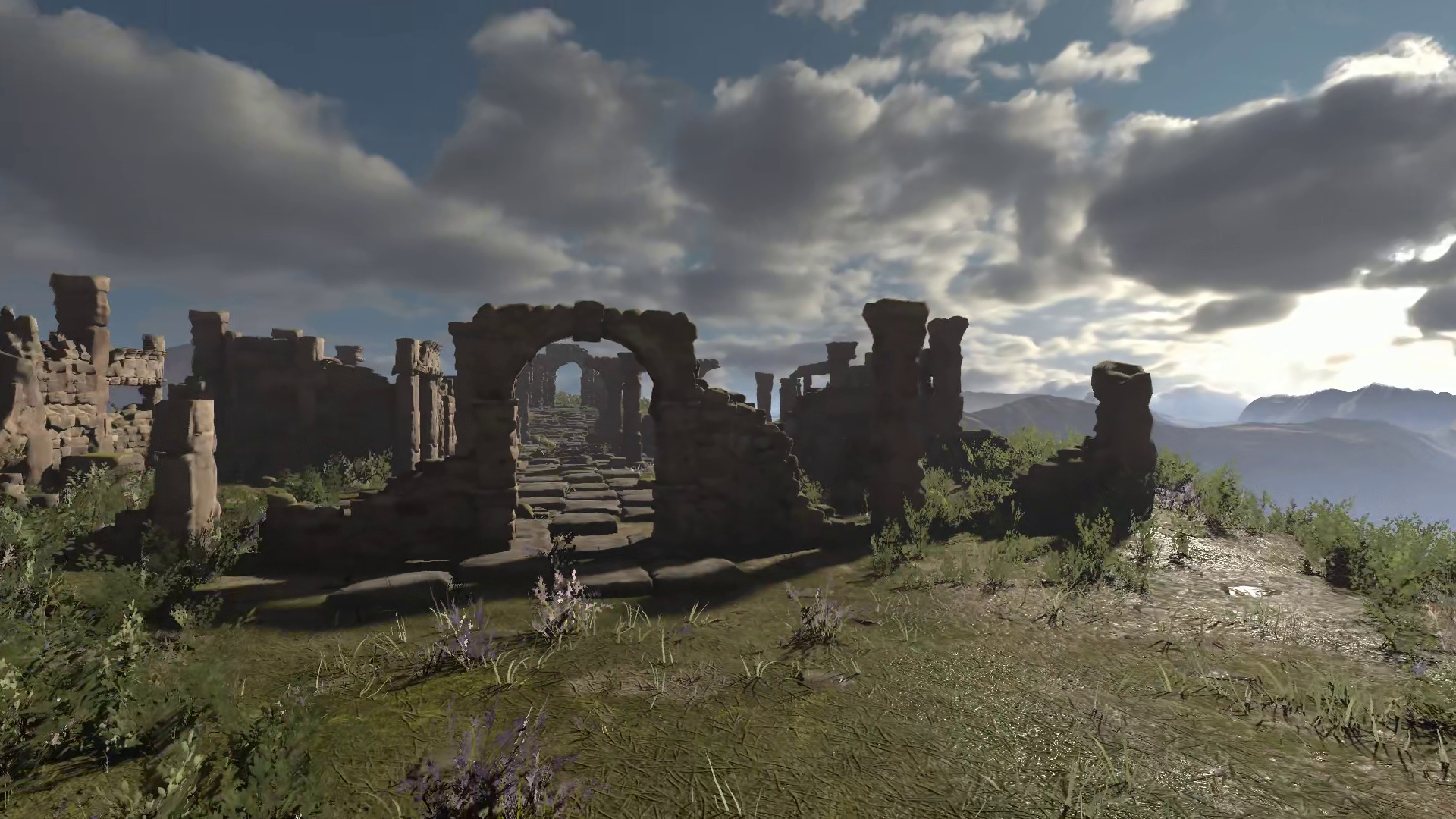} &
  \includegraphics[width=0.326\linewidth]{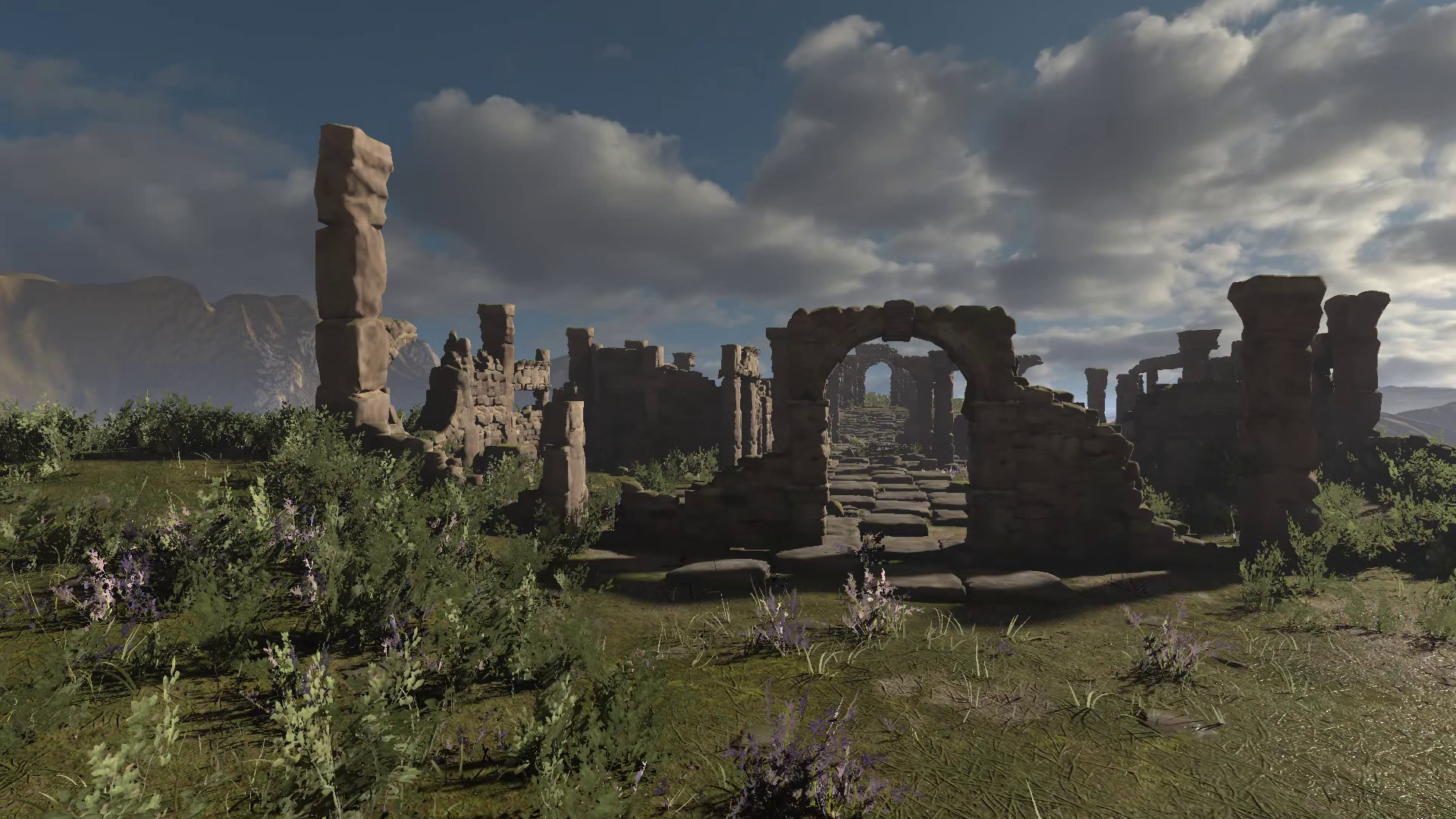}
  \end{tabular}
  \caption{\textbf{Residual rotation mismatch in a Reversed Rotation case.}
  HY-WorldPlay's generated endpoint preserves recognizable ruins, while the
  matched GT frame shows closer image-space correspondence than the
  prescribed endpoint. The matched GT frame precedes the prescribed endpoint along the return
  trajectory, with a GT camera rotation difference of $24^\circ$.
}
  \label{fig:hy_reference_comparison}
\end{figure}

Fixed references change the Seen rankings: Matrix-Game 3 leads DINOv2,
and Matrix-Game 3.5 leads the other five means. Fixed GT endpoint comparisons
reflect both scene-recovery errors and residual camera-control errors.
Figure~\ref{fig:hy_reference_comparison} shows that HY-WorldPlay preserves
the main ruins, including the archway, steps, and walls, but its endpoint
view more closely resembles an earlier GT frame along the return trajectory.
Comparing directly with the prescribed endpoint also counts structure
displacements and changes in visible extent caused by the rotational mismatch
as image error. When assessing scene recovery, this error should not be
attributed entirely to forgetting. Local reference matching therefore aims
to reduce the influence of residual view mismatch. The fixed-reference analysis tests sensitivity to reference selection; it
does not match history duration or other trajectory conditions between Seen
and Novel.

\input{tables/appendix_fixed_endpoint}

\subsection{Full Rollout by revisit type}
\input{tables/appendix_full_seen}
\input{tables/appendix_full_novel}

In Tables~\ref{tab:appendix_full_seen} and~\ref{tab:appendix_full_novel}, all six
open-weight models score lower on Full Rotation than on Reversed Rotation
in all three dimensions. Among Novel revisit types, Genie 3 retains high
Subject Consistency in Forward and Look Back, but its geometry score is
0.300, compared with 0.740 in New-Viewpoint Revisit. Matrix-Game 3.5 has
low geometry scores in both Novel revisit types (0.140 and 0.120). These
results reflect memory, action following, and generation stability together.

\paragraph{Paired confidence intervals.}
To quantify uncertainty in the observed Full Rollout mean differences, we
resample cases while preserving model pairings.
Table~\ref{tab:full_paired_ci} compares the two highest observed task means
per dimension. In each of 10,000 bootstrap replicates (seed 20260926), we
resample 15 cases with replacement within each revisit type, keeping the eight
models and three dimensions paired. Task means weight their two types equally.
The 2.5 and 97.5 percentiles of leader-minus-runner-up differences give pointwise
95\% intervals, unadjusted for selecting the leaders or multiple comparisons.
These intervals cover case sampling for the existing outputs and scores,
not scene clustering or variation across generation seeds and judge calls.
Genie 3's Appearance Fidelity and Geometric Consistency advantages exclude
zero in both tasks, whereas its Subject Consistency intervals include zero.

\input{tables/appendix_full_paired_ci}

\subsection{Paired Confidence Intervals for Continued Rollout}
\label{app:paired_ci}
We quantify how Continued Rollout model means and their paired differences
vary under scene resampling, keeping trajectories from each scene together.

\input{tables/appendix_model_ci}
\input{tables/appendix_paired_ci}

\paragraph{Method.}
From 600 trajectories (3,600 outputs), 10,000 scene-cluster bootstrap replicates
resample 29 Seen or 33 Novel scenes with replacement, retaining all their
trajectories. Six models and all metrics share weights; revisit-type means
are averaged equally. Percentiles 2.5 and 97.5 define 95\% intervals.
Tables~\ref{tab:appendix_model_ci} and~\ref{tab:appendix_paired_ci} report model
means and paired differences between the two highest-ranked observed means
per metric. Differences favor the leader when positive: leader minus runner-up
for higher-is-better metrics, reversed for lower-is-better metrics.

\paragraph{Main findings.}
For Seen, HY-WorldPlay's DINOv2, LPIPS, SI-AbsRel, and normal-cosine advantages
exclude zero. For Novel, Matrix-Game 3's DINOv2 and Matrix-Game 3.5's SSIM
and PSNR advantages exclude zero. Each comparison is against the respective
mean runner-up; the remaining paired intervals include zero.

\clearpage
\input{sections/rebuttal_additions}

%% file: tables/appendix_dataset_statistics.tex
\begin{table}[H]
  \caption{\textbf{Dataset statistics.} Trajectory statistics cover
  Context start through Query completion,
  excluding subsequent travel back to the starting position.}
  \label{tab:appendix_dataset_statistics}
  \centering
  \footnotesize
  \begin{tabular}{p{1.35in}p{3.80in}}
    \toprule
    Dataset detail & Value \\
    \midrule
    Total cases & 660: 600 controlled trajectories + 60 Full Rollout cases \\
    \midrule
    \multicolumn{2}{l}{\textit{Controlled trajectories}} \\
    Case length & 168--877 frames (mean 468.3); 7.0--36.5 seconds
      (mean 19.5) \\
    Total observations & 280,973 RGB frames with synchronized camera poses \\
    Capture format & $1920\times1080$, 24 FPS, $90^\circ$ horizontal FOV \\
    Annotations & Camera poses, actions, trajectory phases, and
      Context/Query boundaries \\
    \midrule
    \multicolumn{2}{l}{\textit{Full Rollout cases}} \\
    Case contents & 60 initial images, each paired with a scene prompt
      and a reference action sequence \\
    Evaluation cases & 60 shared cases; 15 per revisit type \\
    \bottomrule
  \end{tabular}
\end{table}

%% file: tables/appendix_scene_coverage.tex
\begin{table}[H]
  \caption{\textbf{Scene coverage and case counts: 33 scenes and 600 cases.}}
  \label{tab:appendix_scene_coverage}
  \centering
  \footnotesize
  \setlength{\tabcolsep}{3pt}
  \begin{tabular}{@{}>{\raggedright\arraybackslash}p{0.39\textwidth}r@{\hspace{12pt}}>{\raggedright\arraybackslash}p{0.39\textwidth}r@{}}
    \toprule
    Scene & Cases & Scene & Cases \\
    \midrule
    AbandonedDistrict & 30 & Map\_\allowbreak{}ChemicalPlant\_\allowbreak{}1 & 29 \\
    AncientRuins & 29 & Medieval\_\allowbreak{}Daytime & 15 \\
    Brass\_\allowbreak{}Gardens & 27 & MiddleEast & 2 \\
    Brass\_\allowbreak{}Palace & 29 & ModularBuilding & 6 \\
    Brass\_\allowbreak{}Streets & 32 & ModularNeighborhood & 19 \\
    Cabin\_\allowbreak{}Lake & 18 & ModularSciFiVillage & 5 \\
    ChineseWaterTown\_\allowbreak{}Ver1 & 26 & ModularVictorianCity & 11 \\
    ContainerYard\_\allowbreak{}Night & 29 & PostSoviet\_\allowbreak{}Village & 7 \\
    CourtYard & 25 & RussianWinterTownDemo01 & 12 \\
    Demonstration\_\allowbreak{}Castle & 23 & SnowMap & 10 \\
    DesertMap & 18 & Stadium & 4 \\
    Desert\_\allowbreak{}ruins & 29 & SuburbNeighborhood\_\allowbreak{}Day\_\allowbreak{}dooropen & 6 \\
    Greek\_\allowbreak{}Island & 35 & SunsetMap & 1 \\
    IndustrialArea & 28 & Tokyo & 15 \\
    InteriorDemo\_\allowbreak{}NEW & 21 & UltimateFarming & 4 \\
    JapanTrainStation\_\allowbreak{}Optimised & 22 & Venice & 6 \\
    KoreanPalace & 27 &  &  \\
    \bottomrule
  \end{tabular}
\end{table}

%% file: tables/appendix_trajectory_design.tex
\begin{table}[H]
\caption{\textbf{Collection design of the four revisit types. Time gives Context / Query duration ranges in seconds; distance is horizontal translation path length in meters.}}
\label{tab:trajectory_design}
\centering\footnotesize
\setlength{\tabcolsep}{3pt}
\begin{tabular}{@{}>{\raggedright\arraybackslash}p{0.14\textwidth}>{\raggedright\arraybackslash}p{0.30\textwidth}>{\raggedright\arraybackslash}p{0.22\textwidth}>{\raggedright\arraybackslash}p{0.17\textwidth}>{\raggedright\arraybackslash}p{0.07\textwidth}@{}}
\toprule
Type (cases) & Motion and Context & Query onset & Time (s) & Dist. (m) \\
\midrule
Reversed Rotation (100) & Turn $90^\circ$ or $180^\circ$ in place, then reverse; Context includes the outward turn and return motion before the final segment. & Final approximately $90^\circ$ of the reverse turn. & \shortstack[l]{3.50--8.46 /\\3.50--3.54} & 0 \\
\addlinespace[5pt]
Full Rotation (100) & Rotate $360^\circ$ in place; Context covers the first $270^\circ$. & At $270^\circ$, before the final $90^\circ$. & \shortstack[l]{8.46 /\\3.54} & 0 \\
\addlinespace[5pt]
Forward and Look Back (200) & Observe a full turn at $A$, move straight to $B$; Context includes the first half of the turnaround at $B$. & Final $90^\circ$ of the $180^\circ$ turnaround at $B$. & \shortstack[l]{17.21--19.29 /\\2.54} & 18--26 \\
\addlinespace[5pt]
New-Viewpoint Revisit (200) & Observe a full turn at $A$, move to $B$, then turn toward $P$; Context precedes the final turn segment. $\angle APB$ bins: $45^\circ$, $90^\circ$, $135^\circ$. & Final turn toward $P$ at $B$, with at most $90^\circ$ remaining. & \shortstack[l]{18.04--33.00 /\\2.58--3.58} & 18--28 \\
\bottomrule
\end{tabular}
\end{table}

%% file: tables/appendix_history_read_evidence.tex
\begin{table}[H]
  \caption{\textbf{Model-native GT-history initialization and continuation.}}
  \label{tab:history_mechanisms}
  \centering
  \footnotesize
  \setlength{\tabcolsep}{4pt}
  \begin{tabular}{@{}p{0.20\textwidth}p{0.34\textwidth}p{0.40\textwidth}@{}}
    \toprule
    Model & GT Context initialization & History use during continuation \\
    \midrule
    HY-WorldPlay & Encode history latents, conditions, and poses. & Select anchors, recent history, and FOV-retrieved blocks; rebuild the vision cache. \\
    \addlinespace[4pt]
    Matrix-Game 3 & Build a history-latent bank and camera trajectory. & Retrieve latents by FOV overlap and condition on the recent overlap segment. \\
    \addlinespace[4pt]
    Matrix-Game 3.5 & Teacher-force GT chunks into rolling KV, patch memory, and dynamic context. & Combine rolling KV, spatial patch queries, and dynamic context. \\
    \addlinespace[4pt]
    SANA-WM Streaming & Store GT latents, poses, and refiner history. & Replay the history prefix to rebuild state and generate a new suffix each call. \\
    \addlinespace[4pt]
    Zing-World 0.5 & VAE-encode GT; fill KV with clean latent chunks and actions. & Continue with the same KV cache and write back new latents. \\
    \addlinespace[4pt]
    LingBot-World v2 & Streaming-VAE encode GT; fill KV in native four-latent chunks. & Continue using sink and local-window caches. \\
    \bottomrule
  \end{tabular}
\end{table}

%% file: figures/figure7/panels.tex
\begingroup
\setlength{\tabcolsep}{0pt}
\renewcommand{\arraystretch}{0}
\newcommand{\validationstrip}[3]{%
  \raisebox{-0.5\height}{\includegraphics[width=0.62\textwidth,
  trim=0bp #2bp 0bp #3bp,clip]{figures/figure7/GCMC026_all_models_#1.png}}}
\begin{tabular}{@{}p{0.31\textwidth}@{\hspace{0.015\textwidth}}c@{}}
 & \makebox[0.206\textwidth]{\scriptsize\validationearly}%
   \makebox[0.206\textwidth]{\scriptsize\validationboundary}%
   \makebox[0.206\textwidth]{\scriptsize\validationreturn}\\[7pt]
\shortstack[l]{\small\textbf{HY-WorldPlay}\\[3pt]\footnotesize Full Rollout} & \validationstrip{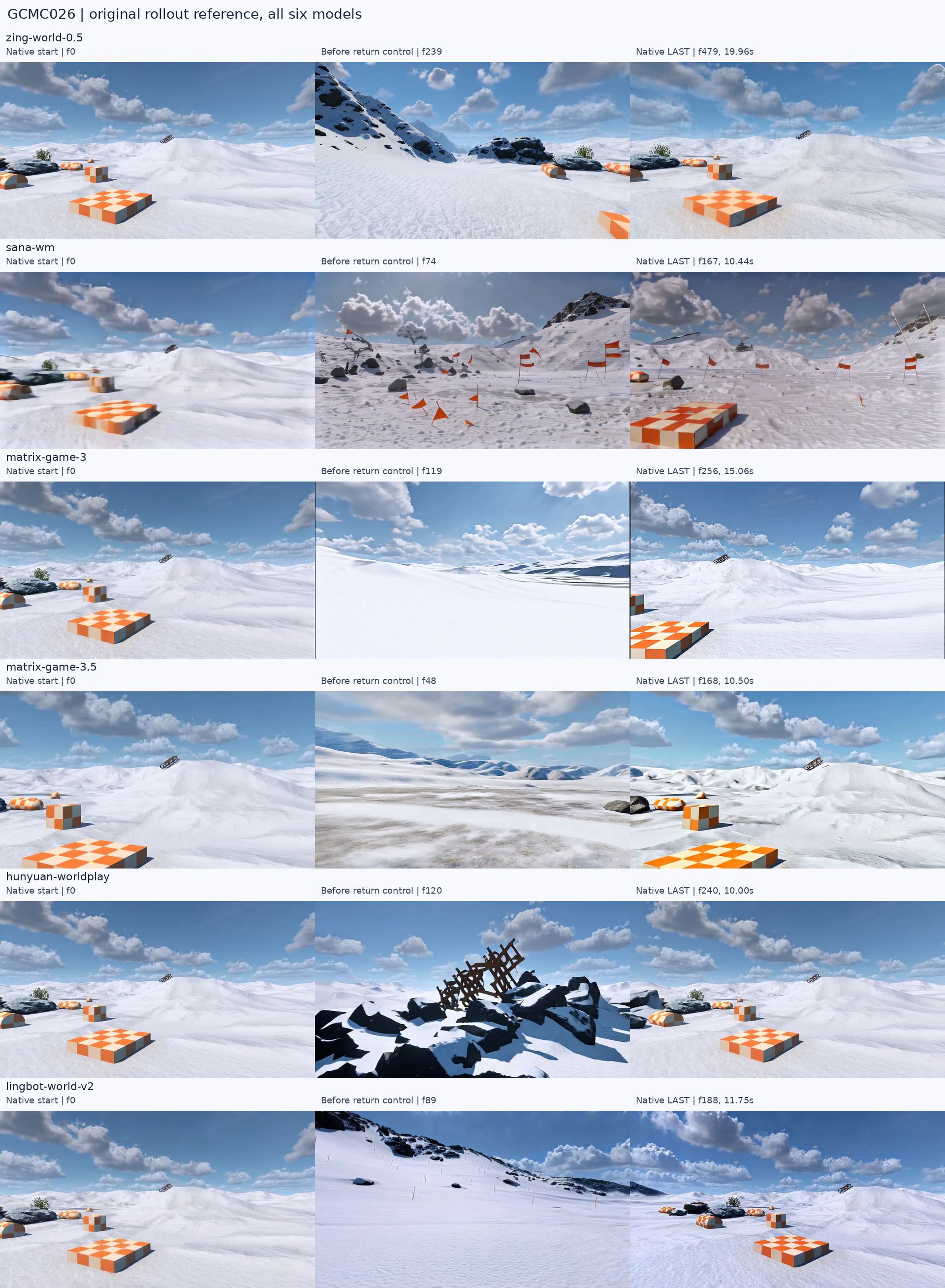}{426}{1830}\\[1pt]
\shortstack[l]{\footnotesize Continued Rollout} & \validationstrip{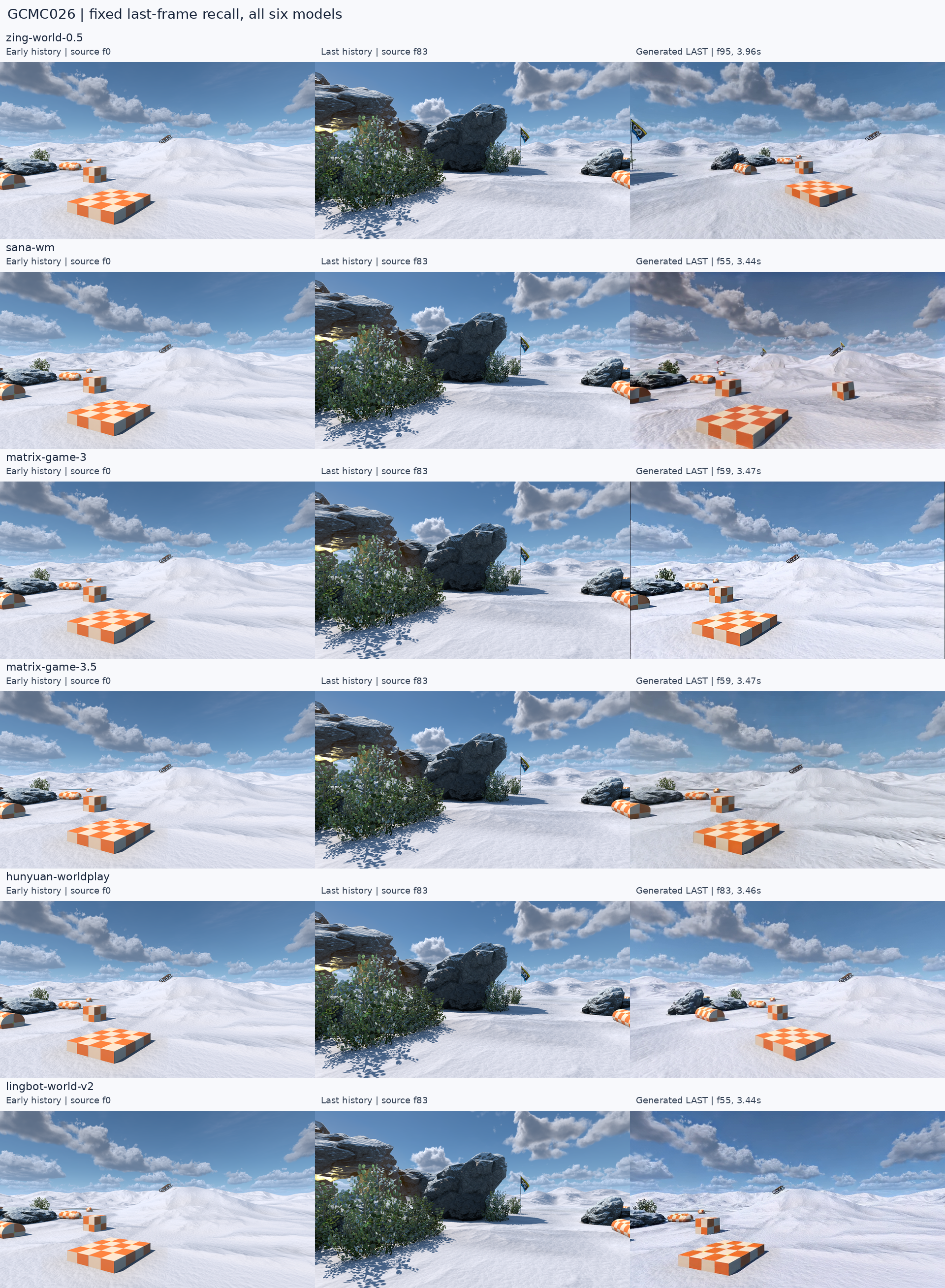}{426}{1830}\\[4pt]
\shortstack[l]{\small\textbf{LingBot-World v2}\\[3pt]\footnotesize Full Rollout} & \validationstrip{native}{0}{2256}\\[1pt]
\shortstack[l]{\footnotesize Continued Rollout} & \validationstrip{recall}{0}{2256}\\[4pt]
\shortstack[l]{\small\textbf{Matrix-Game 3}\\[3pt]\footnotesize Full Rollout} & \validationstrip{native}{1278}{978}\\[1pt]
\shortstack[l]{\footnotesize Continued Rollout} & \validationstrip{recall}{1278}{978}\\[4pt]
\shortstack[l]{\small\textbf{Matrix-Game 3.5}\\[3pt]\footnotesize Full Rollout} & \validationstrip{native}{852}{1404}\\[1pt]
\shortstack[l]{\footnotesize Continued Rollout} & \validationstrip{recall}{852}{1404}\\[4pt]
\shortstack[l]{\small\textbf{SANA-WM Streaming}\\[3pt]\footnotesize Full Rollout} & \validationstrip{native}{1704}{552}\\[1pt]
\shortstack[l]{\footnotesize Continued Rollout} & \validationstrip{recall}{1704}{552}\\[4pt]
\shortstack[l]{\small\textbf{Zing-World 0.5}\\[3pt]\footnotesize Full Rollout} & \validationstrip{native}{2130}{126}\\[1pt]
\shortstack[l]{\footnotesize Continued Rollout} & \validationstrip{recall}{2130}{126}\\[4pt]
\end{tabular}
\endgroup

%% file: tables/appendix_continuation_fidelity.tex
\begin{table}[!t]
  \caption{\textbf{Generation quality and motion response on two Reversed Rotation cases.}
  A: native Full Rollout return; B: GT-history continuation.
  Scores cover the first 2 seconds and equally average the two cases.
  Quality uses raw MUSIQ-SPAQ scores, not percentages.}
  \label{tab:continuation_fidelity}
  \centering
  \small
  \setlength{\tabcolsep}{3pt}
  \begin{tabular}{@{}lccccc@{}}
    \toprule
    Model & \shortstack{Quality mean\\A / B $\uparrow$} & \shortstack{Quality P10\\A / B $\uparrow$} & \shortstack{Smoothness\\A / B $\uparrow$} & \shortstack{B effective\\motion $\uparrow$} & \shortstack{B direction\\accuracy $\uparrow$} \\
    \midrule
    HY-WorldPlay & 78.94 / 79.00 & 77.81 / 78.15 & 0.9898 / 0.9855 & 100\% & 100\% \\
    LingBot-World v2 & 78.29 / 77.66 & 76.32 / 75.95 & 0.9613 / 0.9437 & 100\% & 100\% \\
    Matrix-Game 3 & 73.01 / 78.33 & 70.68 / 76.62 & 0.9783 / 0.9622 & 100\% & 100\% \\
    Matrix-Game 3.5 & 78.02 / 77.02 & 77.08 / 74.28 & 0.9711 / 0.9596 & 100\% & 100\% \\
    SANA-WM Streaming & 69.22 / 71.29 & 62.47 / 59.56 & 0.9798 / 0.9705 & 100\% & 100\% \\
    Zing-World 0.5 & 78.46 / 77.99 & 78.02 / 77.17 & 0.9845 / 0.9755 & 100\% & 100\% \\
    \bottomrule
  \end{tabular}
\end{table}

%% file: tables/appendix_history_dependence.tex
\begin{table}[!t]
\centering
\small
\setlength{\tabcolsep}{4pt}
\caption{\textbf{Target-region scores under history interventions.} One Reversed Rotation case, scored against the fixed, unmodified GT endpoint. Top: six-model interventions. Bottom: a shared occluded initial frame; Full retains the original history after that frame.}
\label{tab:history_dependence}
\begin{tabular}{lrrrrrr}
\toprule
& \multicolumn{3}{c}{ROI DINOv2 $\uparrow$} & \multicolumn{3}{c}{ROI LPIPS $\downarrow$} \\
\cmidrule(lr){2-4}\cmidrule(lr){5-7}
Model & Full & Recent & Replaced & Full & Recent & Replaced \\
\midrule
HY-WorldPlay & 0.7171 & 0.4149 & 0.5775 & 0.6110 & 0.6341 & 0.6733 \\
LingBot-World v2 & 0.8579 & 0.4325 & 0.6196 & 0.6129 & 0.6811 & 0.5932 \\
Matrix-Game 3 & 0.7128 & 0.3192 & 0.3312 & 0.6305 & 0.6470 & 0.8084 \\
Matrix-Game 3.5 & 0.8361 & 0.3809 & -0.0342 & 0.3310 & 0.6924 & 0.8213 \\
SANA-WM Streaming & 0.3806 & 0.4191 & 0.1366 & 0.5930 & 0.6484 & 0.6798 \\
Zing-World 0.5 & 0.3545 & 0.1409 & 0.4089 & 0.6113 & 0.7714 & 0.7498 \\
\midrule
Model & Full & Target & Control & Full & Target & Control \\
\midrule
HY-WorldPlay & 0.6998 & 0.5775 & 0.7229 & 0.6205 & 0.6733 & 0.6182 \\
Matrix-Game 3.5 & 0.7941 & -0.0342 & 0.8028 & 0.4597 & 0.8213 & 0.4609 \\
\bottomrule
\end{tabular}
\end{table}

%% file: tables/appendix_native_formats.tex
\begin{table}[!t]
  \caption{\textbf{Native Full Rollout output formats.} The six open-weight
  models also participate in Continued Rollout using the same checkpoints.
  These frame rates describe Full Rollout outputs only.}
  \label{tab:appendix_native_formats}
  \centering
  \footnotesize
  \begin{tabular}{@{}llrr@{}}
    \toprule
    Model & Access & Resolution & FPS \\
    \midrule
    HY-WorldPlay & Open weights & $848\times480$ & 24 \\
    LingBot-World v2 & Open weights & $832\times464$ & 16 \\
    Matrix-Game 3 & Open weights & $1280\times704$ & 17 \\
    Matrix-Game 3.5 & Open weights & $1280\times704$ & 17 \\
    SANA-WM Streaming & Open weights & $1280\times704$ & 16 \\
    Zing-World 0.5 & Open weights & $1280\times704$ & 24 \\
    Genie 3 & Web service & $1280\times704$ & 20 \\
    Happy Oyster & Web service & $1296\times720$ & 24 \\
    \bottomrule
  \end{tabular}
\end{table}

%% file: prompts/motion_check.tex
\begin{quote}
\small
This is a separate unscored camera-motion check, NOT a memory or image-quality assessment.
Inspect all 12 chronological intermediate frames. Decide whether recognizable camera
movement through/viewing a scene is present (rotation, translation, or mixed motion).
Do not require an exact angle, prescribed direction, complete loop, return to the initial
view, or successful completion of the requested trajectory. Partial motion and overshoot
are acceptable. Do not reject because subjects, appearance or geometry drift, or because
the final view is wrong: those are separate endpoint assessments. Static images, only
object animation, exposure flicker, or cuts with no discernible camera movement do not
establish camera motion. Use uncertain when evidence cannot distinguish these.
The trajectory label is context, not evidence. Sparse frames cannot certify full-video
continuity. Cite visible changes and frame numbers. Do not assign S/A/G or numeric scores.
\end{quote}

%% file: prompts/vqa_english.tex
\begin{quote}
\small
\par\medskip\noindent\textbf{Common instructions}\par\nopagebreak\smallskip
Use only visible evidence in the complete frames A and B; the side-by-side image repeats them. Do not assume successful motion or infer off-screen content. Allow legitimate viewpoint, parallax, scale, and occlusion changes. Assess the coverage of specific correspondence and genuine conflicts. Give partial credit for supported regions, but not a high scene-level score for one landmark or shared style. Score each dimension independently; report no conflict when none is visible.

\par\medskip\noindent\textbf{Shared dimension definitions}\par\nopagebreak\smallskip
Subject Consistency assesses the same specific world through its main subjects and environment. Appearance Fidelity assesses colors, lighting, materials, textures, clarity, and detail on corresponding surfaces in the requested view. Geometric Consistency assesses planes, depth ordering, boundaries, perspective, occlusion, scale, and spatial relations independently of appearance.

\par\medskip\noindent\begin{minipage}{\linewidth}
\par\medskip\noindent\textbf{Scoring scale}\par\nopagebreak\smallskip
Assign a score from 0.0 to 1.0 in steps of 0.1.

\begin{description}
\item[1.0] Nearly all major scene information is strongly supported without contradiction.
\item[0.9] Broad strong agreement, with only minor differences.
\item[0.8] Most important information is supported, with limited gaps.
\item[0.7] The core scene and several major regions agree, with noticeable differences or incomplete coverage.
\item[0.6] Scene-level agreement predominates, with substantial unsupported or conflicting content.
\item[0.5] Scene-level agreement and disagreement or missing evidence are roughly balanced.
\item[0.4] One substantial region matches, but most of the view differs or lacks support.
\item[0.3] Specific partial correspondence is insufficient to support the whole view.
\item[0.2] Only limited local structural or visual correspondence.
\item[0.1] Only category, color, or style is similar.
\item[0.0] No meaningful specific correspondence, or the main scene evidence conflicts.
\end{description}
\end{minipage}\par
\par\medskip\noindent\textbf{1. Reversed Rotation}\par\nopagebreak\smallskip
After leaving A, the camera should return to A's viewing direction: a same-view revisit.

\par\medskip\noindent\textbf{Q1. Subject Consistency}\par\nopagebreak\smallskip
After the reverse turn, does B recover the same specific subjects and surrounding scene as A? Allow small residual pose differences; one recognizable landmark is insufficient.

\par\medskip\noindent\textbf{Q2. Appearance Fidelity}\par\nopagebreak\smallskip
Do surfaces visible again after the reverse turn preserve their materials, colors, lighting, and detail? Distinguish small alignment changes from blur, texture melting, or repainting.

\par\medskip\noindent\textbf{Q3. Geometric Consistency}\par\nopagebreak\smallskip
Do planes, depth ordering, boundaries, and relative positions agree with A? Allow residual pose and edge-cropping differences without treating them as structural replacement.

\par\medskip\noindent\textbf{2. Full Rotation}\par\nopagebreak\smallskip
After a 360-degree turn, B should return to A's initial viewing direction: a same-view revisit.

\par\medskip\noindent\textbf{Q4. Subject Consistency}\par\nopagebreak\smallskip
After the full turn, do the subjects and environment identify the same specific world? A wrong final viewing direction does not alone imply a different world.

\par\medskip\noindent\textbf{Q5. Appearance Fidelity}\par\nopagebreak\smallskip
Has the initial direction been recovered with matching visible appearance? A clear side view or approximately 180-degree view of the same location does not restore the requested view.

\par\medskip\noindent\textbf{Q6. Geometric Consistency}\par\nopagebreak\smallskip
Do subject orientation, left/right layout, dominant planes, and vanishing directions support closure to the initial view? Recognizing the location alone does not demonstrate geometric closure.

\par\medskip\noindent\textbf{3. Forward and Look Back}\par\nopagebreak\smallskip
After moving forward from A, the camera should turn about 180 degrees at B and look back toward A.

\par\medskip\noindent\textbf{Q7. Subject Consistency}\par\nopagebreak\smallskip
Does the view from B show the same specific path and connected environment seen from A? Shared corridor, road, or architectural style alone is insufficient.

\par\medskip\noindent\textbf{Q8. Appearance Fidelity}\par\nopagebreak\smallskip
Do the same path or connected structures retain their appearance when viewed backward? Allow new foreground after forward travel; a clear forward or side view does not satisfy the look-back.

\par\medskip\noindent\textbf{Q9. Geometric Consistency}\par\nopagebreak\smallskip
Do path boundaries, connected openings, depth directions, and left/right relations support forward motion followed by an approximately 180-degree turn? Allow new foreground and elements lost through distance or occlusion.

\par\medskip\noindent\textbf{4. New-Viewpoint Revisit}\par\nopagebreak\smallskip
B should be a meaningful new view of the same specific 3D world as A; pixel alignment is not required.

\par\medskip\noindent\textbf{Q10. Subject Consistency}\par\nopagebreak\smallskip
Does B show the same specific subject and directly visible environmental anchors under the new viewpoint? An isolated recognizable object without verifiable surroundings supports limited scene identity, typically 0.2--0.4; this is a rubric anchor, not a hard cap.

\par\medskip\noindent\textbf{Q11. Appearance Fidelity}\par\nopagebreak\smallskip
Do corresponding surfaces preserve appearance with sufficient surrounding scene correspondence? An isolated matching object with unverifiable surroundings typically supports 0.2--0.4; contradictory surroundings merit less. These are rubric anchors, not hard caps.

\par\medskip\noindent\textbf{Q12. Geometric Consistency}\par\nopagebreak\smallskip
Can subject-ground-background adjacency, support surfaces, occlusion, and perspective follow one plausible camera translation and turn? Require direct non-reflected anchors and continuity between shared regions and newly revealed content.

\end{quote}

%% file: tables/appendix_human_agreement.tex
\begin{table}[!t]
\caption{\textbf{Human--Gemini preference agreement by task and revisit type.} Cells give matches/valid pairs and percentages. U is excluded per dimension; ties agree only when both labels are ties. Task rows pool their two revisit types and are not independent samples.}
\label{tab:human_agreement_groups}
\centering
\footnotesize
\setlength{\tabcolsep}{4pt}
\begin{tabular}{@{}lrrr@{}}
\toprule
Task / revisit type & S & A & G \\
\midrule
Seen & 97/120 (80.8\%) & 101/120 (84.2\%) & 100/120 (83.3\%) \\
Novel & 96/120 (80.0\%) & 84/120 (70.0\%) & 77/112 (68.8\%) \\
Reversed Rotation & 46/60 (76.7\%) & 49/60 (81.7\%) & 48/60 (80.0\%) \\
Full Rotation & 51/60 (85.0\%) & 52/60 (86.7\%) & 52/60 (86.7\%) \\
Forward and Look Back & 45/60 (75.0\%) & 35/60 (58.3\%) & 33/53 (62.3\%) \\
New-Viewpoint Revisit & 51/60 (85.0\%) & 49/60 (81.7\%) & 44/59 (74.6\%) \\
All & 193/240 (80.4\%) & 185/240 (77.1\%) & 177/232 (76.3\%) \\
\bottomrule
\end{tabular}
\end{table}

%% file: figures/figure8/panels.tex
\begingroup
\setlength{\tabcolsep}{2pt}
\newcommand{\prefimage}[1]{\includegraphics[width=0.235\textwidth]{figures/figure8/#1.png}}
\small
\begin{tabular}{@{}cccc@{}}
\multicolumn{4}{l}{\textbf{(a) Reversed Rotation: agreement on all three dimensions}}\\[4pt]
\multicolumn{2}{c}{Left candidate} & \multicolumn{2}{c}{Right candidate}\\
Initial observation & Revisit & Initial observation & Revisit\\[2pt]
\prefimage{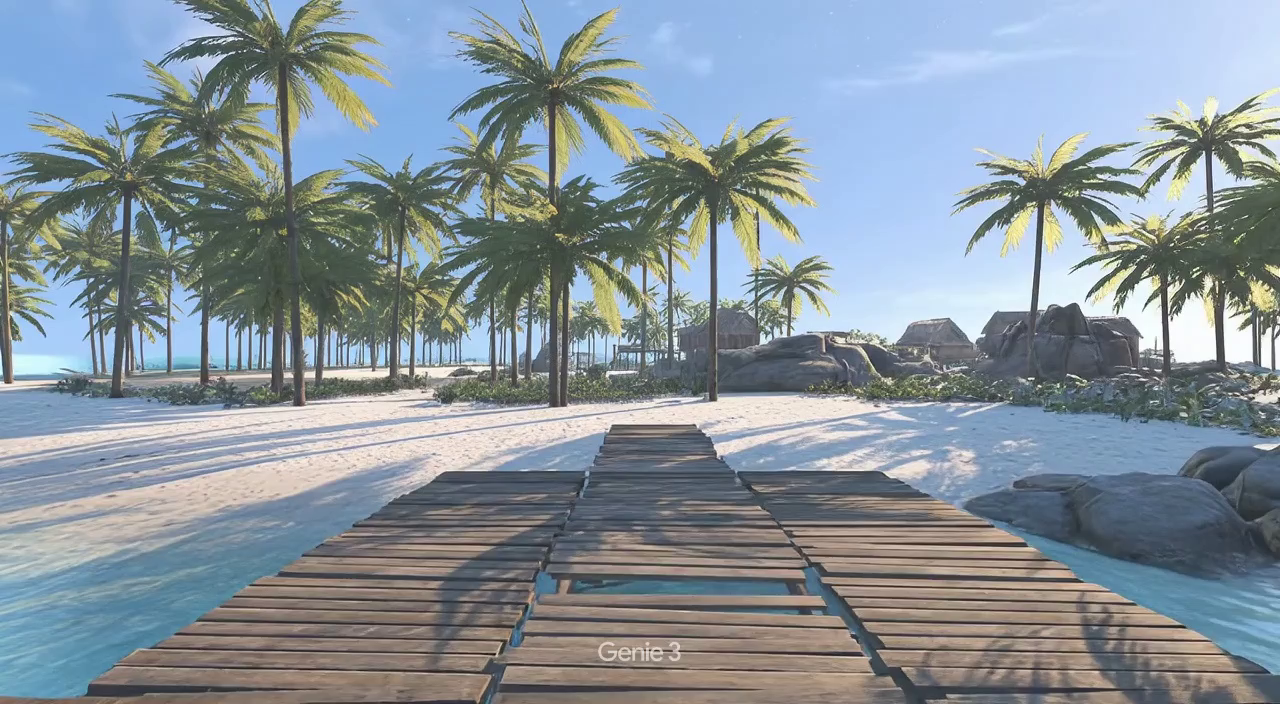} & \prefimage{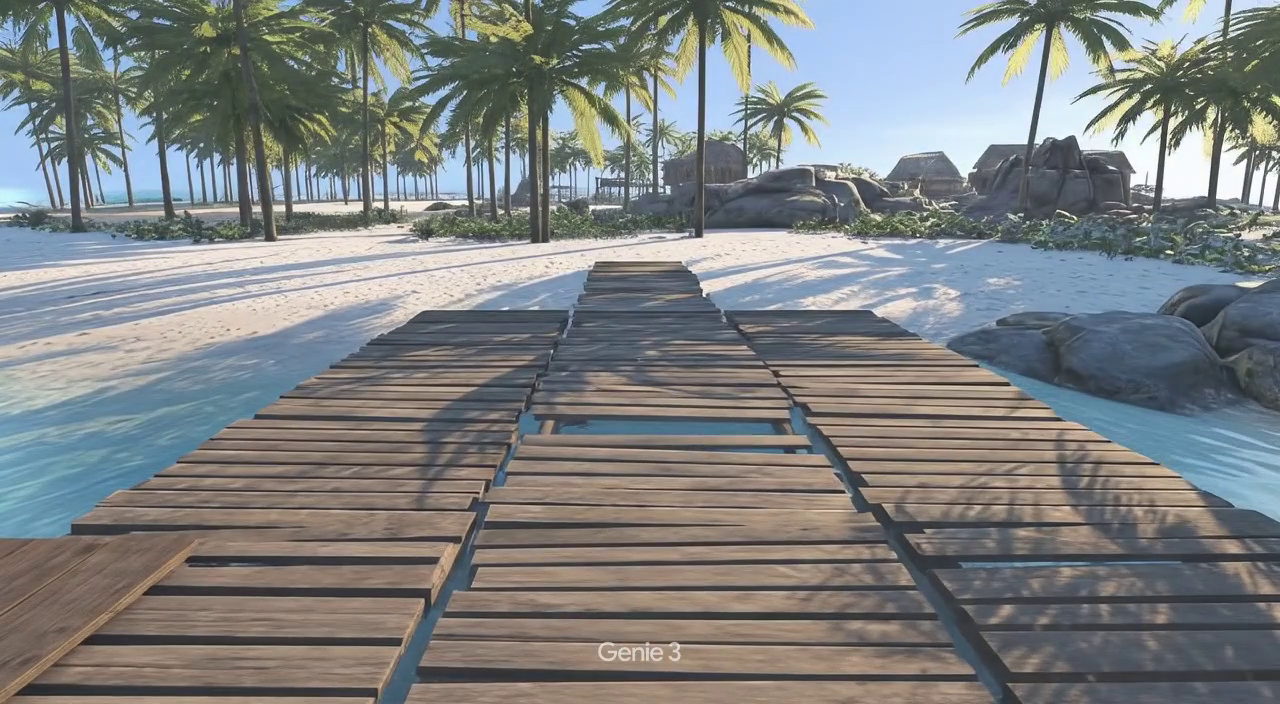} &
\prefimage{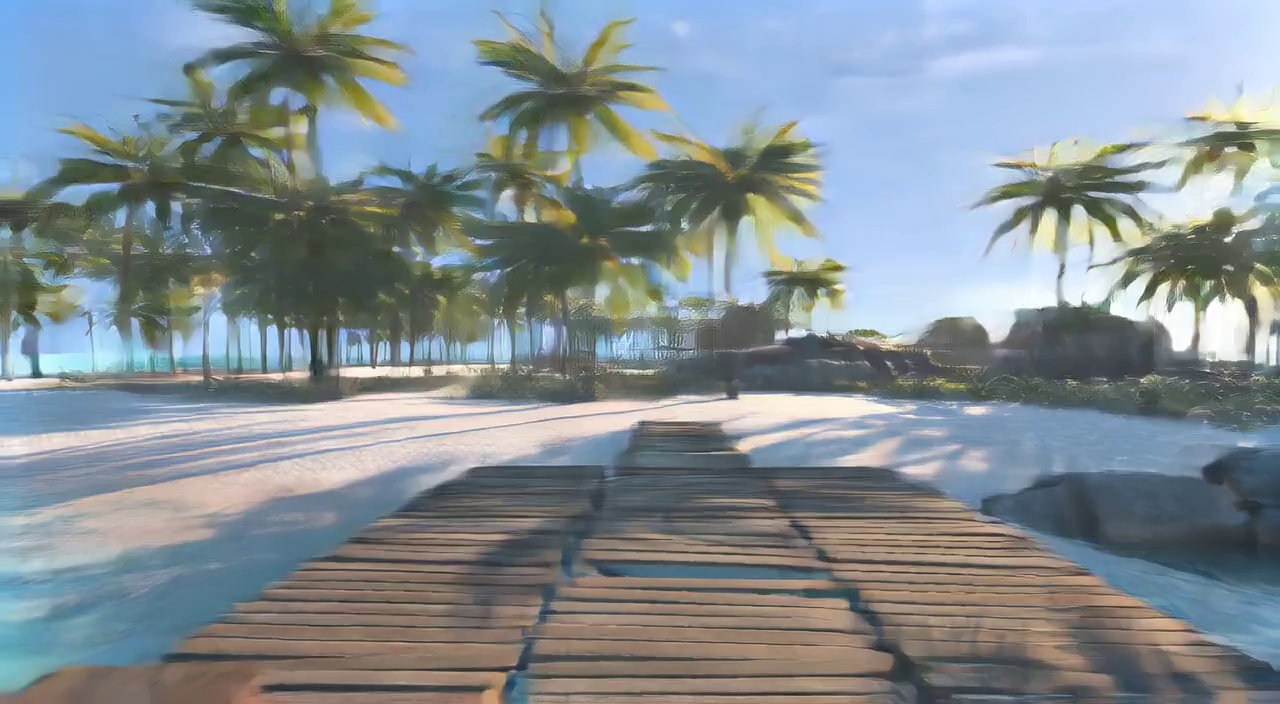} & \prefimage{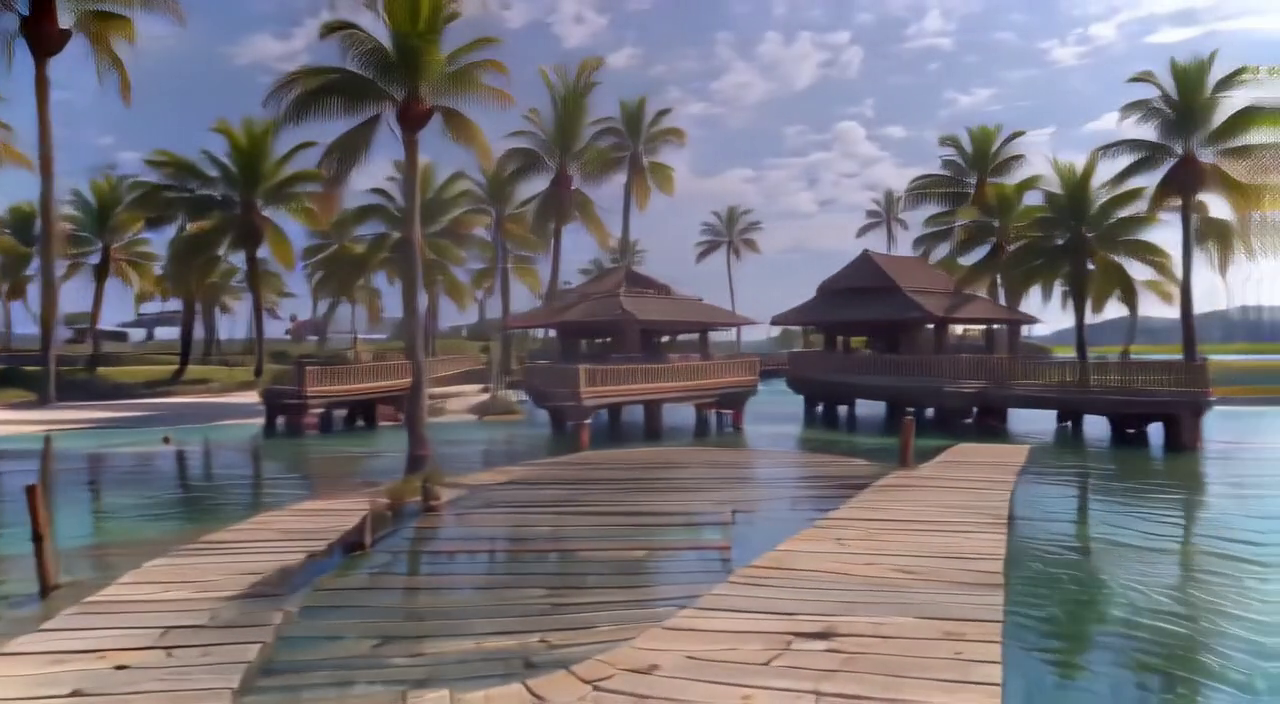}\\[5pt]
\multicolumn{4}{c}{\begin{tabular}{lccc}
 & Subject & Appearance & Geometry\\
Human preference & L & L & L\\
Gemini scores (left / right) & 1.0 / 0.1 & 1.0 / 0.2 & 0.9 / 0.1\\
Gemini preference & L & L & L
\end{tabular}}\\[12pt]
\multicolumn{4}{l}{\textbf{(b) Full Rotation: agreement on all three dimensions}}\\[4pt]
\multicolumn{2}{c}{Left candidate} & \multicolumn{2}{c}{Right candidate}\\
Initial observation & Revisit & Initial observation & Revisit\\[2pt]
\prefimage{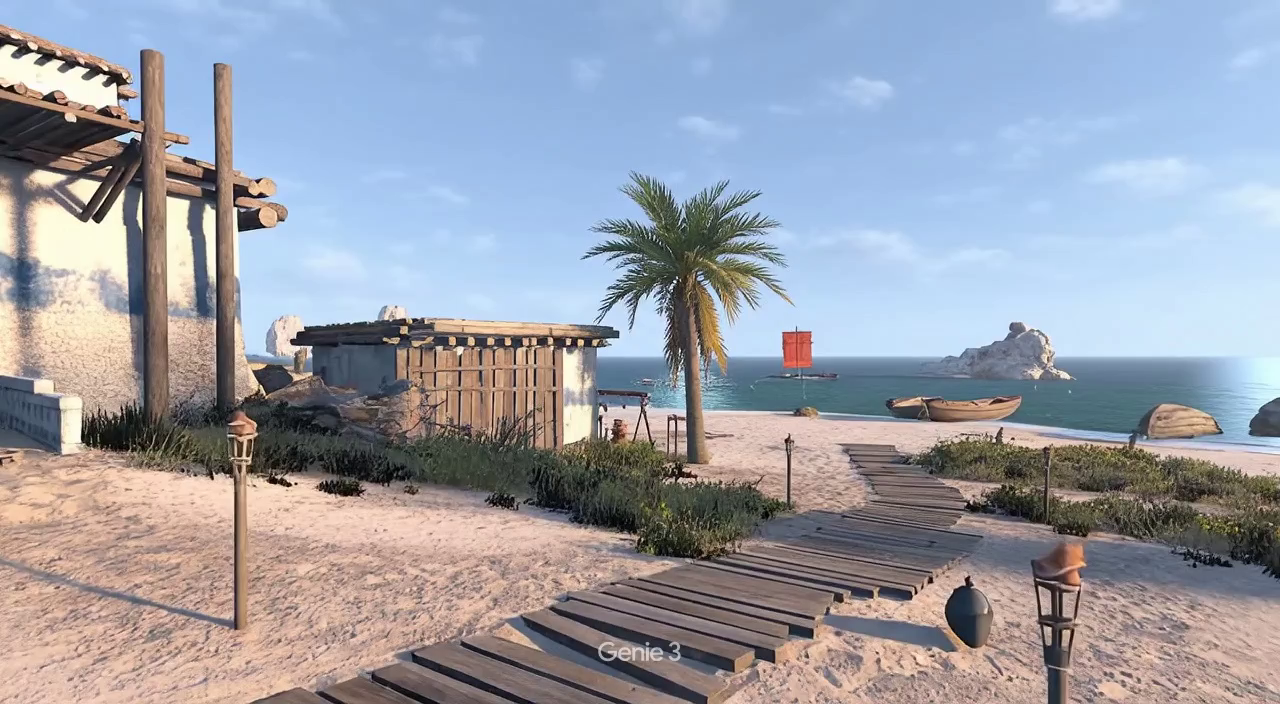} & \prefimage{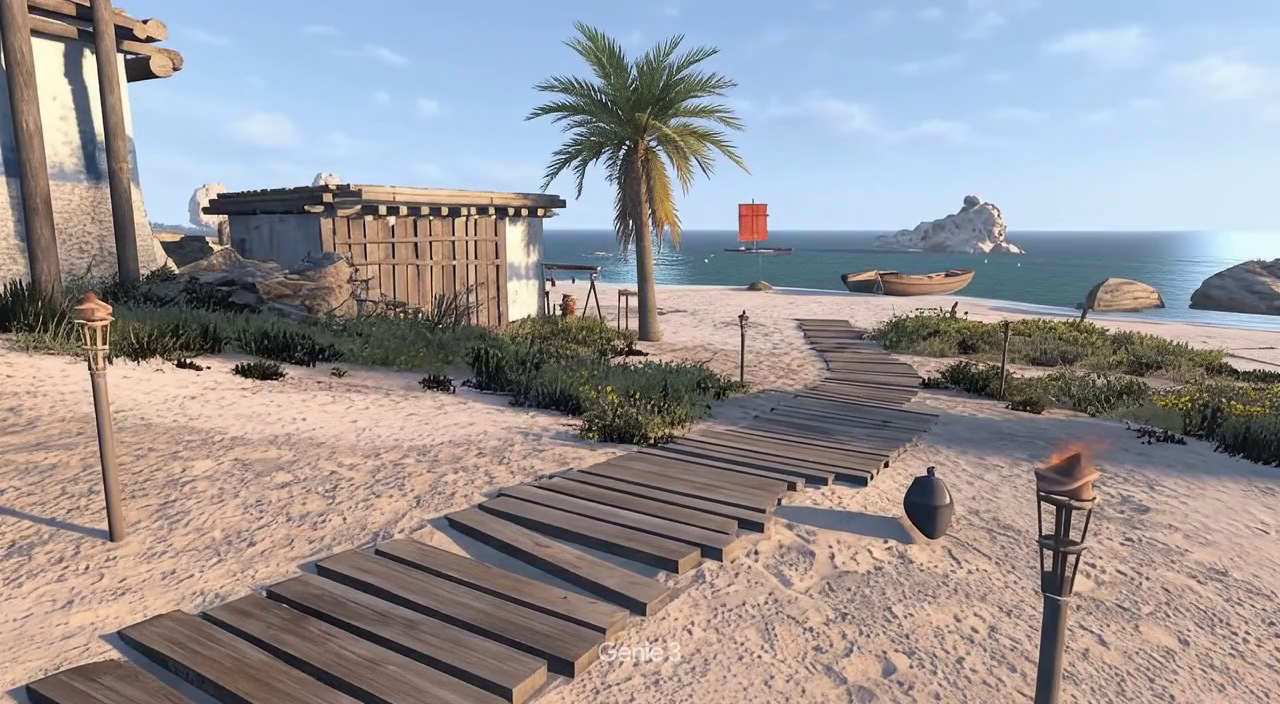} &
\prefimage{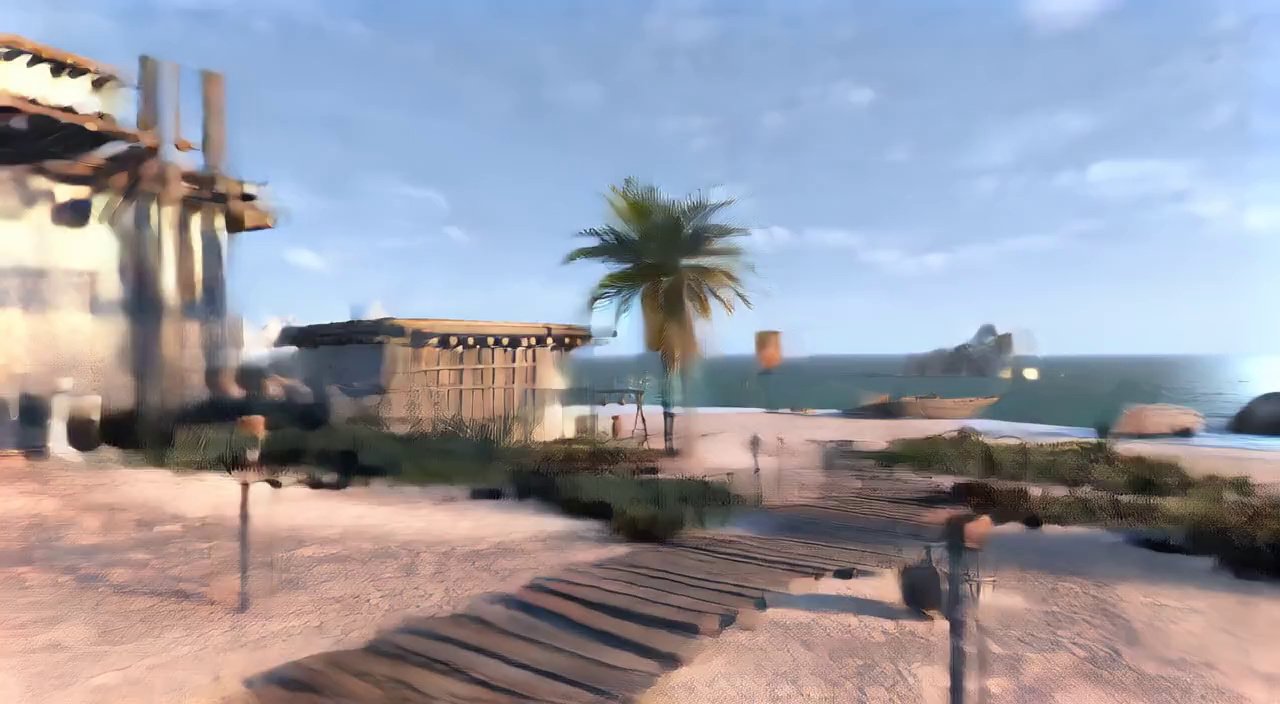} & \prefimage{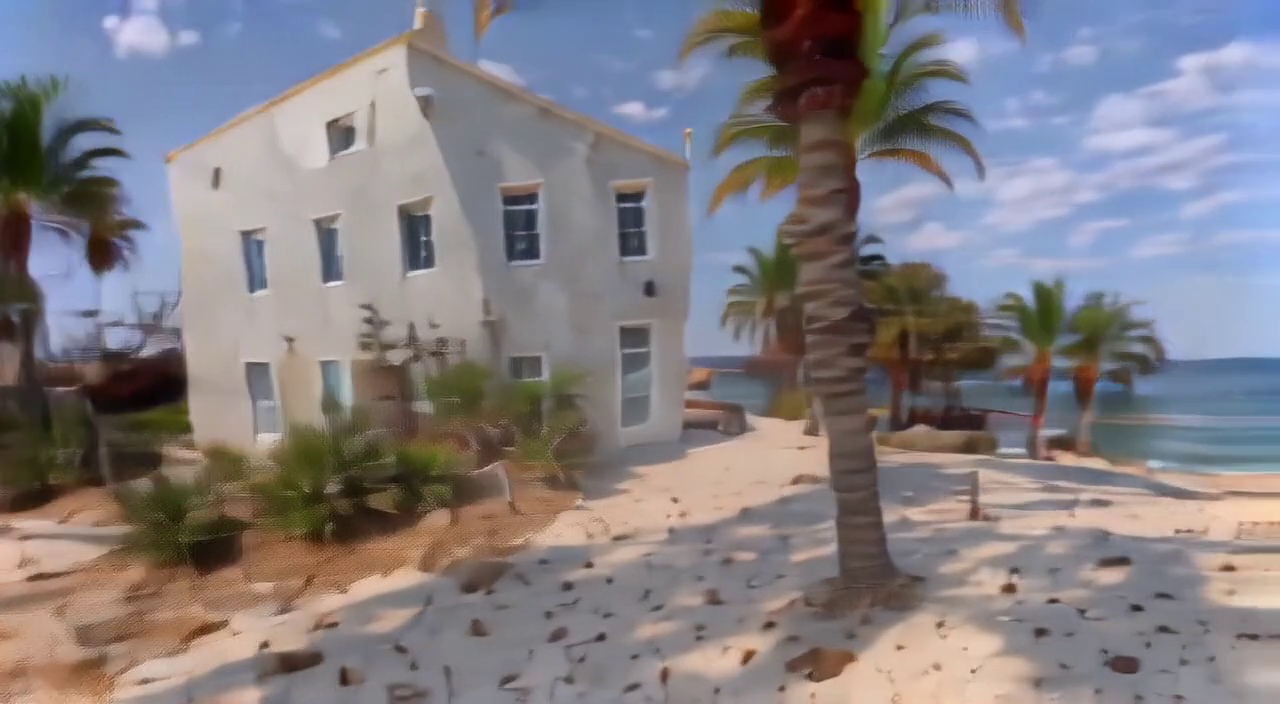}\\[5pt]
\multicolumn{4}{c}{\begin{tabular}{lccc}
 & Subject & Appearance & Geometry\\
Human preference & L & L & L\\
Gemini scores (left / right) & 1.0 / 0.1 & 0.9 / 0.1 & 0.9 / 0.1\\
Gemini preference & L & L & L
\end{tabular}}\\[12pt]
\multicolumn{4}{l}{\textbf{(c) Forward and Look Back: disagreement on all three dimensions}}\\[4pt]
\multicolumn{2}{c}{Left candidate} & \multicolumn{2}{c}{Right candidate}\\
Initial observation & Revisit & Initial observation & Revisit\\[2pt]
\prefimage{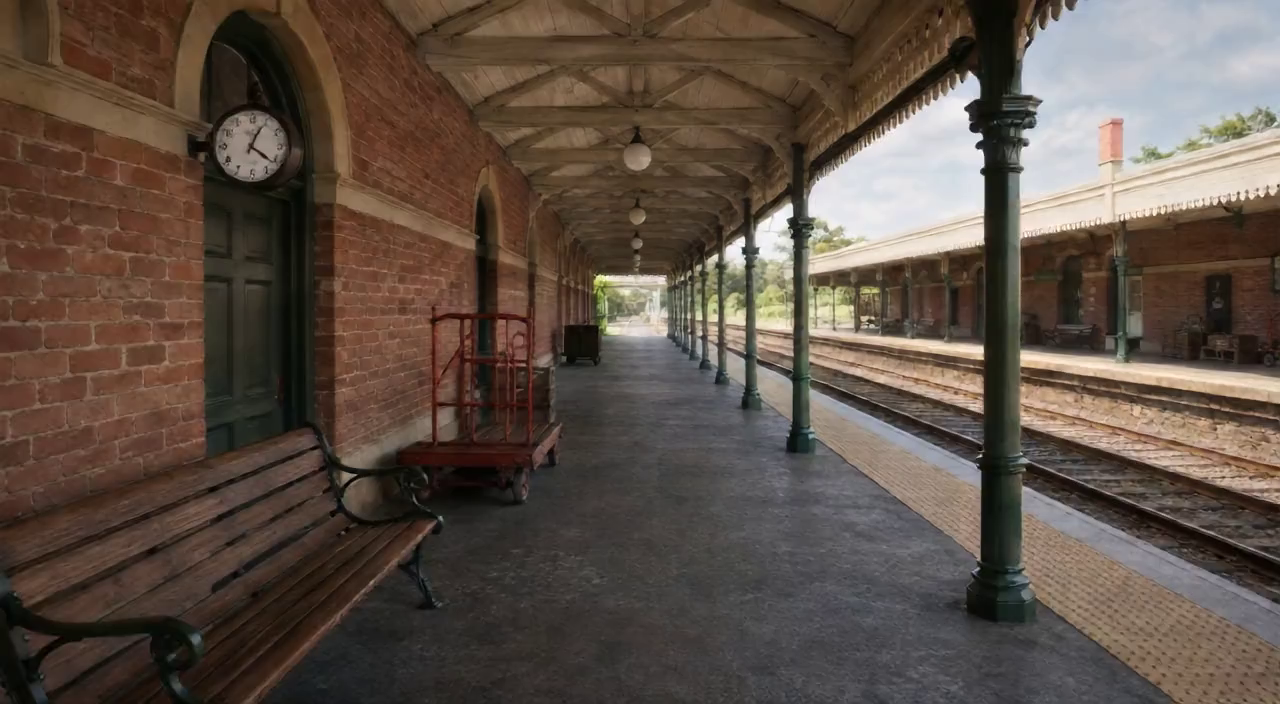} & \prefimage{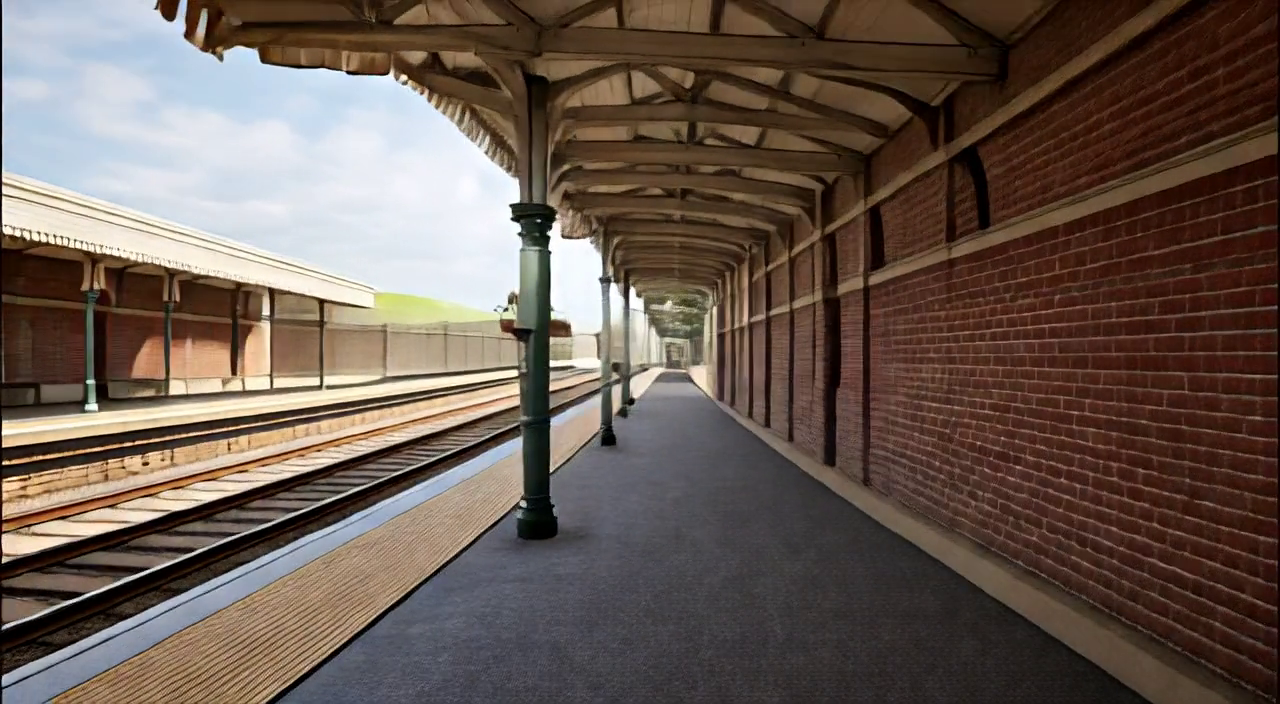} &
\prefimage{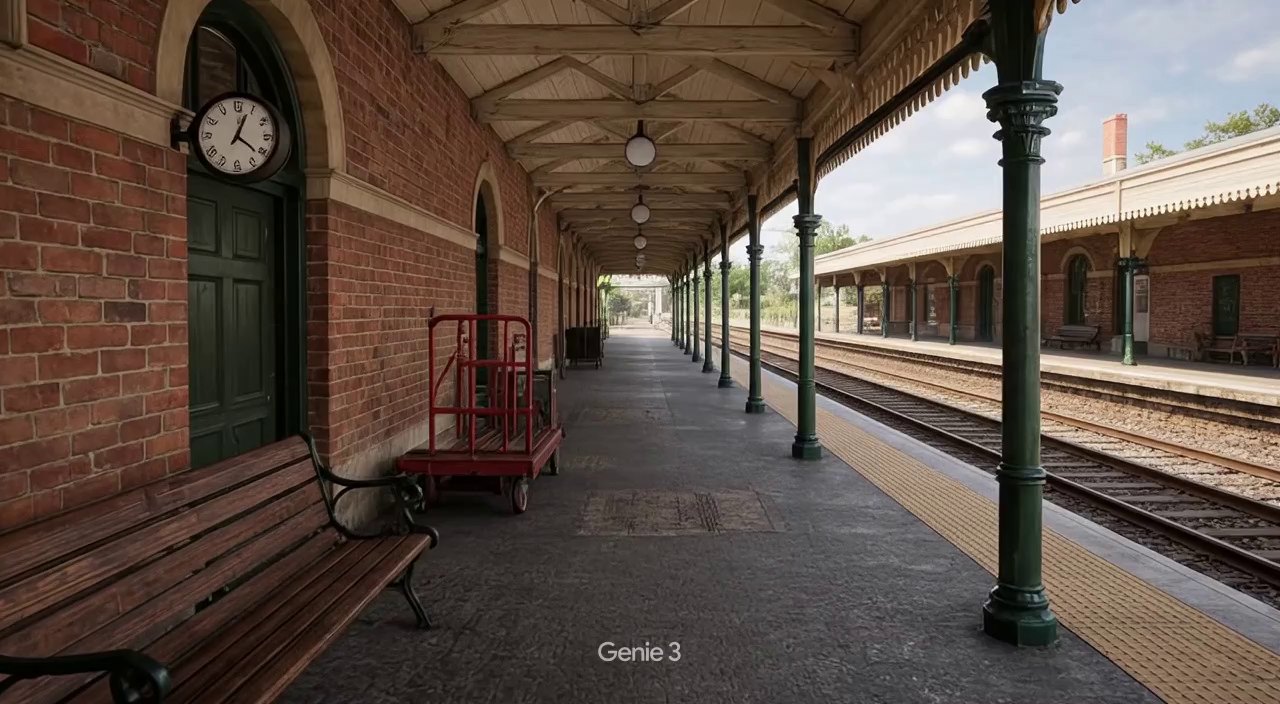} & \prefimage{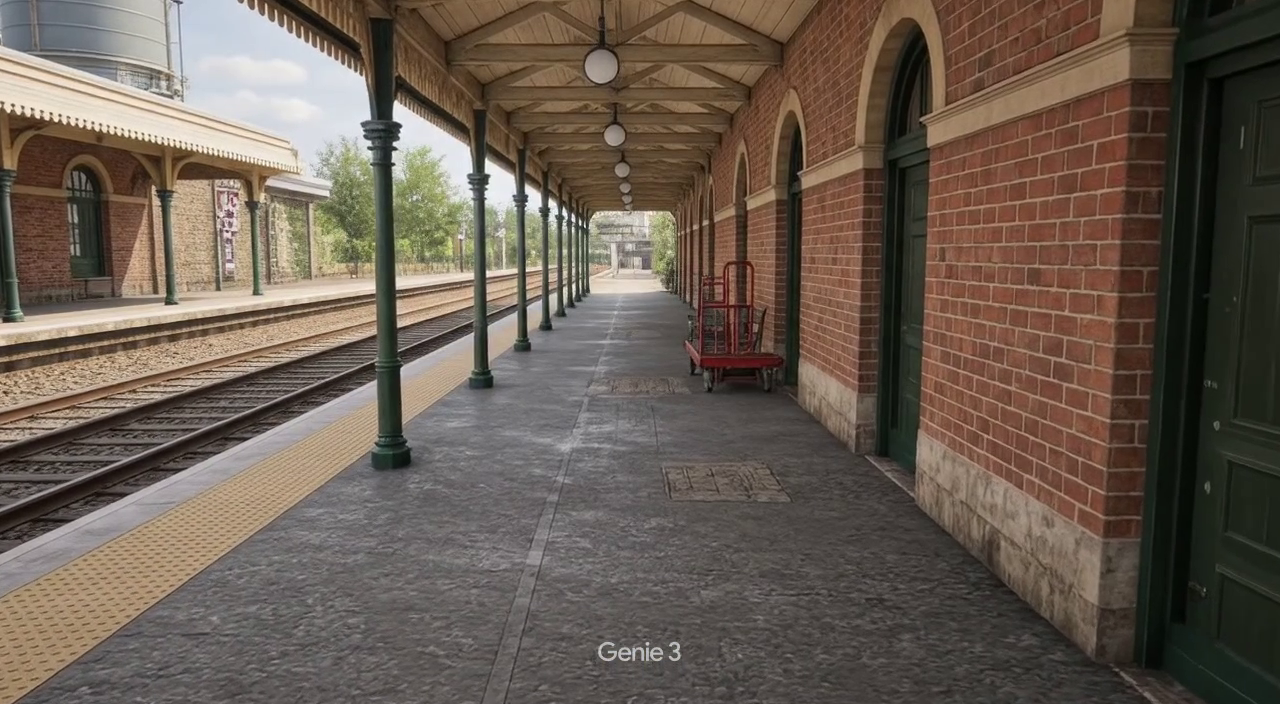}\\[5pt]
\multicolumn{4}{c}{\begin{tabular}{lccc}
 & Subject & Appearance & Geometry\\
Human preference & R & R & R\\
Gemini scores (left / right) & 1.0 / 1.0 & 0.8 / 0.3 & 0.9 / 0.9\\
Gemini preference & T & L & T
\end{tabular}}
\end{tabular}
\endgroup

%% file: tables/appendix_continued_seen.tex
\begin{table}[!htbp]
  \caption{\textbf{Continued Rollout by revisit type: Return to a Seen View.} Arithmetic means over all scheduled cases; metric definitions follow Appendix~\ref{app:gt_metric_details}. PSNR is in dB.}
  \label{tab:appendix_continued_seen}
  \centering
  \footnotesize
  \setlength{\tabcolsep}{3pt}
  \begin{tabular}{@{}lrrrrrr@{}}
    \toprule
    Model & DINOv2$\uparrow$ & LPIPS$\downarrow$ & SSIM$\uparrow$ & PSNR$\uparrow$ & SI-AbsRel$\downarrow$ & Normal Cos.$\uparrow$ \\
    \midrule
    \multicolumn{7}{l}{\textit{Reversed Rotation} (100 cases per model)} \\
    HY-WorldPlay & 0.9386 & 0.2268 & 0.4712 & 17.6024 & 0.1730 & 0.8519 \\
    LingBot-World v2 & 0.8134 & 0.5292 & 0.3185 & 13.2270 & 0.3876 & 0.6950 \\
    Matrix-Game 3 & 0.9014 & 0.3502 & 0.3887 & 15.6712 & 0.2338 & 0.7815 \\
    Matrix-Game 3.5 & 0.7869 & 0.3531 & 0.4325 & 16.2107 & 0.3071 & 0.7668 \\
    SANA-WM Streaming & 0.7262 & 0.4413 & 0.3293 & 14.6233 & 0.3636 & 0.7596 \\
    Zing-World 0.5 & 0.8769 & 0.3844 & 0.3735 & 15.2095 & 0.3204 & 0.7794 \\
    \midrule
    \multicolumn{7}{l}{\textit{Full Rotation} (100 cases per model)} \\
    HY-WorldPlay & 0.9194 & 0.3281 & 0.3967 & 16.0060 & 0.2383 & 0.7935 \\
    LingBot-World v2 & 0.7364 & 0.4963 & 0.3386 & 13.2312 & 0.4119 & 0.7092 \\
    Matrix-Game 3 & 0.8666 & 0.4862 & 0.3376 & 13.9905 & 0.3554 & 0.7293 \\
    Matrix-Game 3.5 & 0.7511 & 0.3472 & 0.4409 & 16.5361 & 0.2286 & 0.8290 \\
    SANA-WM Streaming & 0.5227 & 0.4916 & 0.3108 & 14.2522 & 0.3641 & 0.7448 \\
    Zing-World 0.5 & 0.7246 & 0.5532 & 0.3362 & 13.0436 & 0.5885 & 0.6802 \\
    \bottomrule
  \end{tabular}
\end{table}

%% file: tables/appendix_continued_novel.tex
\begin{table}[!t]
  \caption{\textbf{Continued Rollout by revisit type: Return to a Novel View.} Arithmetic means over all scheduled cases; metric definitions follow Appendix~\ref{app:gt_metric_details}. PSNR is in dB.}
  \label{tab:appendix_continued_novel}
  \centering
  \footnotesize
  \setlength{\tabcolsep}{3pt}
  \begin{tabular}{@{}lrrrrrr@{}}
    \toprule
    Model & DINOv2$\uparrow$ & LPIPS$\downarrow$ & SSIM$\uparrow$ & PSNR$\uparrow$ & SI-AbsRel$\downarrow$ & Normal Cos.$\uparrow$ \\
    \midrule
    \multicolumn{7}{l}{\textit{Forward and Look Back} (200 cases per model)} \\
    HY-WorldPlay & 0.6754 & 0.5323 & 0.3469 & 13.6646 & 0.4503 & 0.6874 \\
    LingBot-World v2 & 0.5374 & 0.5792 & 0.2938 & 12.5741 & 0.5802 & 0.6434 \\
    Matrix-Game 3 & 0.7392 & 0.5104 & 0.3412 & 13.5699 & 0.5281 & 0.6855 \\
    Matrix-Game 3.5 & 0.5199 & 0.5228 & 0.3634 & 14.1885 & 0.4663 & 0.6882 \\
    SANA-WM Streaming & 0.3414 & 0.6707 & 0.2351 & 11.5193 & 0.8454 & 0.5235 \\
    Zing-World 0.5 & 0.6527 & 0.5753 & 0.3200 & 13.0158 & 0.6436 & 0.6659 \\
    \midrule
    \multicolumn{7}{l}{\textit{New-Viewpoint Revisit} (200 cases per model)} \\
    HY-WorldPlay & 0.7220 & 0.5478 & 0.3352 & 13.4877 & 0.4740 & 0.6745 \\
    LingBot-World v2 & 0.5765 & 0.5876 & 0.3115 & 12.2532 & 0.6691 & 0.6305 \\
    Matrix-Game 3 & 0.7695 & 0.5143 & 0.3376 & 13.7758 & 0.4979 & 0.7154 \\
    Matrix-Game 3.5 & 0.6190 & 0.5054 & 0.3713 & 14.5632 & 0.4073 & 0.7276 \\
    SANA-WM Streaming & 0.3221 & 0.6690 & 0.2422 & 11.5493 & 0.7646 & 0.5416 \\
    Zing-World 0.5 & 0.6739 & 0.5508 & 0.3499 & 13.8150 & 0.5768 & 0.6931 \\
    \bottomrule
  \end{tabular}
\end{table}

%% file: tables/appendix_fixed_endpoint.tex
\begin{table}[!t]
\caption{\textbf{Continued Rollout with fixed GT endpoint references.} The same 3,600 generated outputs are compared with the final GT Query frame in both tasks. Seen/Novel contain 200/400 cases per model, with equal revisit-type weights. Novel results are unchanged from Table~\ref{tab:combined_results}. Bold marks the best means within each task.}
\label{tab:fixed_endpoint}
\centering
\footnotesize
\setlength{\tabcolsep}{3pt}
\begin{tabular}{@{}lrrrrrr@{}}
\toprule
Model & DINOv2$\uparrow$ & LPIPS$\downarrow$ & SSIM$\uparrow$ & PSNR$\uparrow$ & SI-AbsRel$\downarrow$ & Normal cos.$\uparrow$ \\
\midrule
\multicolumn{7}{l}{\textit{Return to a Seen View}} \\
\midrule
HY-WorldPlay & 0.8468 & 0.5383 & 0.3360 & 13.2550 & 0.4665 & 0.6771 \\
LingBot-World v2 & 0.7757 & 0.5085 & 0.3306 & 13.2874 & 0.3911 & 0.7078 \\
Matrix-Game 3 & \textbf{0.8808} & 0.4158 & 0.3684 & 14.9269 & 0.2864 & 0.7624 \\
Matrix-Game 3.5 & 0.7748 & \textbf{0.3028} & \textbf{0.4821} & \textbf{17.3612} & \textbf{0.2419} & \textbf{0.8275} \\
SANA-WM Streaming & 0.6274 & 0.4460 & 0.3270 & 14.6909 & 0.3392 & 0.7643 \\
Zing-World 0.5 & 0.8010 & 0.4981 & 0.3452 & 13.8171 & 0.4601 & 0.7158 \\
\midrule
\multicolumn{7}{l}{\textit{Return to a Novel View}} \\
\midrule
HY-WorldPlay & 0.6987 & 0.5401 & 0.3410 & 13.5762 & 0.4622 & 0.6809 \\
LingBot-World v2 & 0.5569 & 0.5834 & 0.3026 & 12.4137 & 0.6246 & 0.6369 \\
Matrix-Game 3 & \textbf{0.7543} & \textbf{0.5124} & 0.3394 & 13.6729 & 0.5130 & 0.7004 \\
Matrix-Game 3.5 & 0.5694 & 0.5141 & \textbf{0.3674} & \textbf{14.3759} & \textbf{0.4368} & \textbf{0.7079} \\
SANA-WM Streaming & 0.3318 & 0.6699 & 0.2387 & 11.5343 & 0.8050 & 0.5325 \\
Zing-World 0.5 & 0.6633 & 0.5630 & 0.3349 & 13.4154 & 0.6102 & 0.6795 \\
\bottomrule
\end{tabular}
\end{table}

%% file: tables/appendix_full_seen.tex
\begin{table}[!t]
  \caption{\textbf{Full Rollout by revisit type: Return to a Seen View.} Each entry averages 15 shared cases, scored by Gemini 3.8 Flash. S/A/G denote Subject Consistency, Appearance Fidelity, and Geometric Consistency; all scores lie in $[0,1]$ and higher is better.}
  \label{tab:appendix_full_seen}
  \centering
  \footnotesize
  \setlength{\tabcolsep}{3pt}
  \begin{tabular}{@{}lrrrrrr@{}}
    \toprule
    & \multicolumn{3}{c}{Reversed Rotation} & \multicolumn{3}{c}{Full Rotation} \\
    \cmidrule(lr){2-4}\cmidrule(lr){5-7}
    Model & S$\uparrow$ & A$\uparrow$ & G$\uparrow$ & S$\uparrow$ & A$\uparrow$ & G$\uparrow$ \\
    \midrule
    Genie 3 & 0.980 & 0.900 & 0.887 & 0.913 & 0.807 & 0.753 \\
    Happy Oyster & 0.793 & 0.573 & 0.567 & 0.873 & 0.480 & 0.467 \\
    HY-WorldPlay & 1.000 & 0.973 & 0.973 & 0.813 & 0.380 & 0.387 \\
    LingBot-World v2 & 0.400 & 0.287 & 0.247 & 0.380 & 0.167 & 0.120 \\
    Matrix-Game 3 & 0.907 & 0.700 & 0.693 & 0.807 & 0.513 & 0.487 \\
    Matrix-Game 3.5 & 0.973 & 0.820 & 0.893 & 0.393 & 0.187 & 0.173 \\
    SANA-WM Streaming & 0.413 & 0.233 & 0.227 & 0.120 & 0.067 & 0.040 \\
    Zing-World 0.5 & 0.333 & 0.287 & 0.273 & 0.053 & 0.040 & 0.013 \\
    \bottomrule
  \end{tabular}
\end{table}

%% file: tables/appendix_full_novel.tex
\begin{table}[!t]
  \caption{\textbf{Full Rollout by revisit type: Return to a Novel View.} Each entry averages 15 shared cases, scored by Gemini 3.8 Flash. S/A/G denote Subject Consistency, Appearance Fidelity, and Geometric Consistency; all scores lie in $[0,1]$ and higher is better.}
  \label{tab:appendix_full_novel}
  \centering
  \footnotesize
  \setlength{\tabcolsep}{3pt}
  \begin{tabular}{@{}lrrrrrr@{}}
    \toprule
    & \multicolumn{3}{c}{Forward and Look Back} & \multicolumn{3}{c}{New-Viewpoint Revisit} \\
    \cmidrule(lr){2-4}\cmidrule(lr){5-7}
    Model & S$\uparrow$ & A$\uparrow$ & G$\uparrow$ & S$\uparrow$ & A$\uparrow$ & G$\uparrow$ \\
    \midrule
    Genie 3 & 0.800 & 0.440 & 0.300 & 0.740 & 0.860 & 0.740 \\
    Happy Oyster & 0.660 & 0.287 & 0.193 & 0.613 & 0.587 & 0.493 \\
    HY-WorldPlay & 0.813 & 0.333 & 0.287 & 0.620 & 0.533 & 0.480 \\
    LingBot-World v2 & 0.427 & 0.200 & 0.140 & 0.400 & 0.247 & 0.187 \\
    Matrix-Game 3 & 0.600 & 0.307 & 0.213 & 0.667 & 0.640 & 0.520 \\
    Matrix-Game 3.5 & 0.207 & 0.140 & 0.140 & 0.280 & 0.120 & 0.120 \\
    SANA-WM Streaming & 0.267 & 0.147 & 0.127 & 0.407 & 0.220 & 0.193 \\
    Zing-World 0.5 & 0.180 & 0.100 & 0.087 & 0.247 & 0.087 & 0.047 \\
    \bottomrule
  \end{tabular}
\end{table}

%% file: tables/appendix_full_paired_ci.tex
\begin{table}[!t]
\caption{\textbf{Full Rollout paired differences between the two leading means.} Positive values favor the leader; pointwise 95\% case-bootstrap intervals. HY: HY-WorldPlay; MG3: Matrix-Game 3. S/A/G denote the three evaluation dimensions.}
\label{tab:full_paired_ci}
\centering
\footnotesize
\setlength{\tabcolsep}{4pt}
\begin{tabular}{@{}lllrr@{}}
\toprule
Task & Dimension & Leader / runner-up & Advantage & 95\% interval \\
\midrule
Seen & S & Genie 3 / HY & $0.0400$ & $[-0.0300, 0.0933]$ \\
Seen & A & Genie 3 / HY & $0.1767$ & $[0.0833, 0.2600]$ \\
Seen & G & Genie 3 / HY & $0.1400$ & $[0.0333, 0.2367]$ \\
Novel & S & Genie 3 / HY & $0.0533$ & $[-0.0400, 0.1367]$ \\
Novel & A & Genie 3 / MG3 & $0.1767$ & $[0.0667, 0.2900]$ \\
Novel & G & Genie 3 / HY & $0.1367$ & $[0.0067, 0.2667]$ \\
\bottomrule
\end{tabular}
\end{table}

%% file: tables/appendix_model_ci.tex
\begin{table}[!t]
  \caption{\textbf{Continued Rollout means and 95\% confidence intervals.} Each cell gives the task mean above its pointwise scene-cluster bootstrap interval; all 600 trajectories, including pitch, are used.}
  \label{tab:appendix_model_ci}
  \centering
  \scriptsize
  \setlength{\tabcolsep}{2pt}
  \renewcommand{\arraystretch}{1.0}
  \begin{tabular}{@{}lrrrrrr@{}}
    \toprule
    \multicolumn{7}{l}{\textbf{Return to a Seen View}} \\
    Model & DINOv2$\uparrow$ & LPIPS$\downarrow$ & SSIM$\uparrow$ & PSNR$\uparrow$ & SI-AbsRel$\downarrow$ & Normal cos.$\uparrow$ \\
    \midrule
    HY-WorldPlay & \shortstack{$0.929$\\$[0.917, 0.940]$} & \shortstack{$0.277$\\$[0.253, 0.300]$} & \shortstack{$0.434$\\$[0.385, 0.488]$} & \shortstack{$16.80$\\$[15.59, 18.10]$} & \shortstack{$0.206$\\$[0.168, 0.247]$} & \shortstack{$0.823$\\$[0.802, 0.846]$} \\
    LingBot-World v2 & \shortstack{$0.775$\\$[0.726, 0.820]$} & \shortstack{$0.513$\\$[0.485, 0.539]$} & \shortstack{$0.329$\\$[0.290, 0.372]$} & \shortstack{$13.23$\\$[12.26, 14.30]$} & \shortstack{$0.400$\\$[0.340, 0.461]$} & \shortstack{$0.702$\\$[0.674, 0.731]$} \\
    Matrix-Game 3 & \shortstack{$0.884$\\$[0.870, 0.897]$} & \shortstack{$0.418$\\$[0.396, 0.439]$} & \shortstack{$0.363$\\$[0.323, 0.410]$} & \shortstack{$14.83$\\$[13.70, 16.12]$} & \shortstack{$0.295$\\$[0.255, 0.335]$} & \shortstack{$0.755$\\$[0.731, 0.782]$} \\
    Matrix-Game 3.5 & \shortstack{$0.769$\\$[0.746, 0.790]$} & \shortstack{$0.350$\\$[0.335, 0.366]$} & \shortstack{$0.437$\\$[0.397, 0.481]$} & \shortstack{$16.37$\\$[15.45, 17.37]$} & \shortstack{$0.268$\\$[0.219, 0.336]$} & \shortstack{$0.798$\\$[0.774, 0.821]$} \\
    SANA-WM Streaming & \shortstack{$0.624$\\$[0.590, 0.662]$} & \shortstack{$0.466$\\$[0.445, 0.487]$} & \shortstack{$0.320$\\$[0.278, 0.367]$} & \shortstack{$14.44$\\$[13.47, 15.46]$} & \shortstack{$0.364$\\$[0.267, 0.520]$} & \shortstack{$0.752$\\$[0.725, 0.782]$} \\
    Zing-World 0.5 & \shortstack{$0.801$\\$[0.778, 0.824]$} & \shortstack{$0.469$\\$[0.445, 0.491]$} & \shortstack{$0.355$\\$[0.311, 0.405]$} & \shortstack{$14.13$\\$[13.10, 15.22]$} & \shortstack{$0.454$\\$[0.335, 0.598]$} & \shortstack{$0.730$\\$[0.705, 0.757]$} \\
    \midrule
    \multicolumn{7}{l}{\textbf{Return to a Novel View}} \\
    Model & DINOv2$\uparrow$ & LPIPS$\downarrow$ & SSIM$\uparrow$ & PSNR$\uparrow$ & SI-AbsRel$\downarrow$ & Normal cos.$\uparrow$ \\
    \midrule
    HY-WorldPlay & \shortstack{$0.699$\\$[0.672, 0.725]$} & \shortstack{$0.540$\\$[0.521, 0.557]$} & \shortstack{$0.341$\\$[0.295, 0.390]$} & \shortstack{$13.58$\\$[12.47, 14.77]$} & \shortstack{$0.462$\\$[0.406, 0.525]$} & \shortstack{$0.681$\\$[0.653, 0.711]$} \\
    LingBot-World v2 & \shortstack{$0.557$\\$[0.500, 0.613]$} & \shortstack{$0.583$\\$[0.558, 0.605]$} & \shortstack{$0.303$\\$[0.266, 0.343]$} & \shortstack{$12.41$\\$[11.40, 13.57]$} & \shortstack{$0.625$\\$[0.544, 0.705]$} & \shortstack{$0.637$\\$[0.606, 0.673]$} \\
    Matrix-Game 3 & \shortstack{$0.754$\\$[0.726, 0.784]$} & \shortstack{$0.512$\\$[0.487, 0.536]$} & \shortstack{$0.339$\\$[0.294, 0.389]$} & \shortstack{$13.67$\\$[12.57, 14.85]$} & \shortstack{$0.513$\\$[0.421, 0.620]$} & \shortstack{$0.700$\\$[0.671, 0.731]$} \\
    Matrix-Game 3.5 & \shortstack{$0.569$\\$[0.527, 0.614]$} & \shortstack{$0.514$\\$[0.493, 0.535]$} & \shortstack{$0.367$\\$[0.320, 0.415]$} & \shortstack{$14.38$\\$[13.38, 15.45]$} & \shortstack{$0.437$\\$[0.372, 0.507]$} & \shortstack{$0.708$\\$[0.678, 0.739]$} \\
    SANA-WM Streaming & \shortstack{$0.332$\\$[0.292, 0.375]$} & \shortstack{$0.670$\\$[0.657, 0.684]$} & \shortstack{$0.239$\\$[0.205, 0.274]$} & \shortstack{$11.53$\\$[10.77, 12.35]$} & \shortstack{$0.805$\\$[0.716, 0.894]$} & \shortstack{$0.533$\\$[0.505, 0.562]$} \\
    Zing-World 0.5 & \shortstack{$0.663$\\$[0.625, 0.701]$} & \shortstack{$0.563$\\$[0.543, 0.581]$} & \shortstack{$0.335$\\$[0.290, 0.382]$} & \shortstack{$13.42$\\$[12.50, 14.41]$} & \shortstack{$0.610$\\$[0.524, 0.702]$} & \shortstack{$0.679$\\$[0.650, 0.711]$} \\
    \bottomrule
  \end{tabular}
\end{table}

%% file: tables/appendix_paired_ci.tex
\begin{table}[!t]
  \caption{\textbf{Paired differences between the two leading means.} Positive values favor the leader; intervals are pointwise 95\%. HY: HY-WorldPlay; MG3/MG3.5: Matrix-Game 3/3.5.}
  \label{tab:appendix_paired_ci}
  \centering
  \small
  \setlength{\tabcolsep}{5pt}
  \begin{tabular}{@{}lllrr@{}}
    \toprule
    Task & Metric & Leader / runner-up & Advantage & 95\% interval \\
    \midrule
    Seen & DINOv2$\uparrow$ & HY / MG3 & $0.045035$ & $[0.028900, 0.059344]$ \\
     & LPIPS$\downarrow$ & HY / MG3.5 & $0.072669$ & $[0.049055, 0.097578]$ \\
     & SSIM$\uparrow$ & MG3.5 / HY & $0.002767$ & $[-0.020476, 0.023165]$ \\
     & PSNR$\uparrow$ & HY / MG3.5 & $0.430867$ & $[-0.093669, 0.980763]$ \\
     & SI-AbsRel$\downarrow$ & HY / MG3.5 & $0.062154$ & $[0.024301, 0.107078]$ \\
     & Normal cosine$\uparrow$ & HY / MG3.5 & $0.024784$ & $[0.006430, 0.042638]$ \\
    \midrule
    Novel & DINOv2$\uparrow$ & MG3 / HY & $0.055597$ & $[0.032117, 0.080053]$ \\
     & LPIPS$\downarrow$ & MG3 / MG3.5 & $0.001752$ & $[-0.011911, 0.016540]$ \\
     & SSIM$\uparrow$ & MG3.5 / HY & $0.026324$ & $[0.017300, 0.035062]$ \\
     & PSNR$\uparrow$ & MG3.5 / MG3 & $0.702983$ & $[0.343431, 1.045170]$ \\
     & SI-AbsRel$\downarrow$ & MG3.5 / HY & $0.025394$ & $[-0.017210, 0.066247]$ \\
     & Normal cosine$\uparrow$ & MG3.5 / MG3 & $0.007465$ & $[-0.004238, 0.019892]$ \\
    \bottomrule
  \end{tabular}
\end{table}

%% file: sections/rebuttal_additions.tex
\begin{revision}
\setlength{\intextsep}{6pt plus 1pt minus 1pt}
\setlength{\textfloatsep}{8pt plus 1pt minus 1pt}
\setlength{\floatsep}{8pt plus 1pt minus 1pt}
\section{Additional Experiments and Validation}
\label{app:revision_experiments}
These experiments supplement the original 600 controlled trajectories and 60 Full Rollout cases. We report extensions separately from the original benchmark results. HY, LingBot, MG3, MG3.5, SANA, and Zing abbreviate the six open-weight models used throughout the paper.

\subsection{Translation Controls and Viewpoint Changes}
\label{app:rev_translation}
\paragraph{Fixed-reference comparison.}
The fixed-endpoint analysis in Appendix~\ref{app:fixed_endpoint} uses the same 3,600 generated outputs as the main evaluation. Generated query endpoints remain fixed; only the Seen reference rule changes. Novel results are unchanged, while MG3 leads Seen DINOv2 and MG3.5 leads the other five Seen metrics. The main table evaluates each task under its stated reference rule; fixed endpoints provide a common reference rule for comparing Seen and Novel.

\paragraph{Translated Seen controls.}
We add 80 trajectories covering forward--backward motion, lateral round trips, and rectangular loops: four motion types at two distances, with ten trajectories per setting. Round-trip distances are 10/20~m, and rectangle sides are 5/10~m. For rectangles, only the last side is generated; the first three sides form the GT history. For the other trajectories, models generate the return segment from the preceding GT history. Model weights, resolutions, and history strategies are unchanged.
All six models complete all 80 cases, yielding 480 outputs. Original rotation Seen, translated Seen, and Novel use fixed GT and generated query endpoints, with equal weight assigned to each model. Translated Seen outperforms both original Novel and rotation Seen on all six mean metrics (Table~\ref{tab:rev_translation}).
\input{tables/rebuttal_translation}

\paragraph{Viewpoint-angle analysis.}
Larger viewpoint changes are generally harder. We group the 200 New-Viewpoint Revisit cases by $\angle APB$, where $A$ is the initial position, $B$ the revisit position, and $P$ the target. The $45^\circ/90^\circ/135^\circ$ bins contain 48/87/65 cases per model. Across 1,200 fixed-endpoint outputs, moving from $45^\circ$ to $135^\circ$ worsens DINOv2 for five models, LPIPS for all six, and SI-AbsRel and normal cosine for five. The $90^\circ$ bin sometimes improves, and SANA exhibits the opposite endpoint trend (Table~\ref{tab:rev_angles}). Thus, changes across larger angles are not uniformly monotonic.
\input{tables/rebuttal_angles}

\subsection{History Ablations}
\label{app:rev_history}
We extend history interventions to 48 stratified trajectories spanning 12 scenes and all four revisit types. Six models run under three history conditions, yielding 864 outputs evaluated using all six metrics. Full uses the complete GT history. Recent-only retains the final two seconds of GT history. No-history supplies only the final context frame needed to initialize the query, in a fresh process without preceding KV caches or external memory. Weights, resolution, seed, and query commands are held fixed; all conditions use the same GT query endpoint.

Full history improves DINOv2 over No-history for all six models by 0.079--0.273. All six scene-cluster paired-bootstrap intervals for these gains lie above zero. Full history outperforms No-history in 35/36 model--metric mean comparisons and Recent-only in 33/36 (Table~\ref{tab:rev_history}). These results provide an empirical reference for scores attainable from an initial image or recent observations and show that full-history gains extend beyond scene-style priors.
For MG3.5, Full history improves PSNR by 4.05~dB and SSIM by 0.114 over No-history (16.19/0.412 versus 12.14/0.298). With the model, query, and reference held fixed, these differences show that both metrics respond to recovery improvements from historical information despite their low absolute values.
\input{tables/rebuttal_history}

\subsection{Human Validation of Continued Rollout}
\label{app:rev_human}
Our original human study evaluated the Gemini judge used in Full Rollout (Appendix~\ref{app:human_preference_agreement}). We extended this validation to Continued Rollout by randomly sampling 120 pairs of model outputs. For each pair, human evaluation compared both candidate generated revisit frames against the fixed GT query endpoint, judging subject, appearance, and geometry and allowing left, right, or tie. Automatic preferences use DINOv2 for subject, LPIPS for appearance, and SI-AbsRel for geometry, favoring higher DINOv2 and lower LPIPS/SI-AbsRel.

Table~\ref{tab:rev_human} reports agreement between automatic metrics and human preferences on all 120 pairs. Agreement is 87.5\% for subject, 81.7\% for appearance, and 75.0\% for geometry. These results support alignment with human judgments across all three dimensions, with the highest agreement for subject consistency.
\input{tables/rebuttal_human}

\subsection{Evaluation on Real Captures}
\label{app:rev_real}
We recorded 25 handheld videos in indoor and outdoor locations, comprising 14 Seen and 11 Novel revisits with ARKit-estimated camera trajectories. Six models generated 300 outputs across Continued Rollout and Full Rollout. This real-world split will be released with the benchmark.

In Continued Rollout, models generate the revisit segment from recorded GT history. We evaluate the fixed generated query endpoint against the fixed GT query endpoint using the six automatic metrics.

In Full Rollout, models generate the complete interaction from the initial frame, prompt, and controls. Gemini-3.8-flash compares the initial reference frame with the generated revisit endpoint to assess subject, appearance, and geometry. Table~\ref{tab:rev_real_full} reports these scores; overall scores weight Seen and Novel equally. HY leads overall in all three dimensions, followed by MG3. MG3.5 leads Seen but scores much lower on Novel, while LingBot achieves the highest Novel appearance score.
\input{tables/rebuttal_real_full}

These experiments extend SceneMemBench beyond rendered scenes and demonstrate that both protocols can evaluate recovery on real captures. Qualitative examples show successful continuation on real captures (Figure~\ref{fig:rev_real_success}) and similar failure modes in real and UE continuations, including target-view deviations, missing content, and scene rewriting (Figure~\ref{fig:rev_real_failures}).
\begin{figure}[!htbp]
\revisioncolor
\centering
\includegraphics[width=\linewidth]{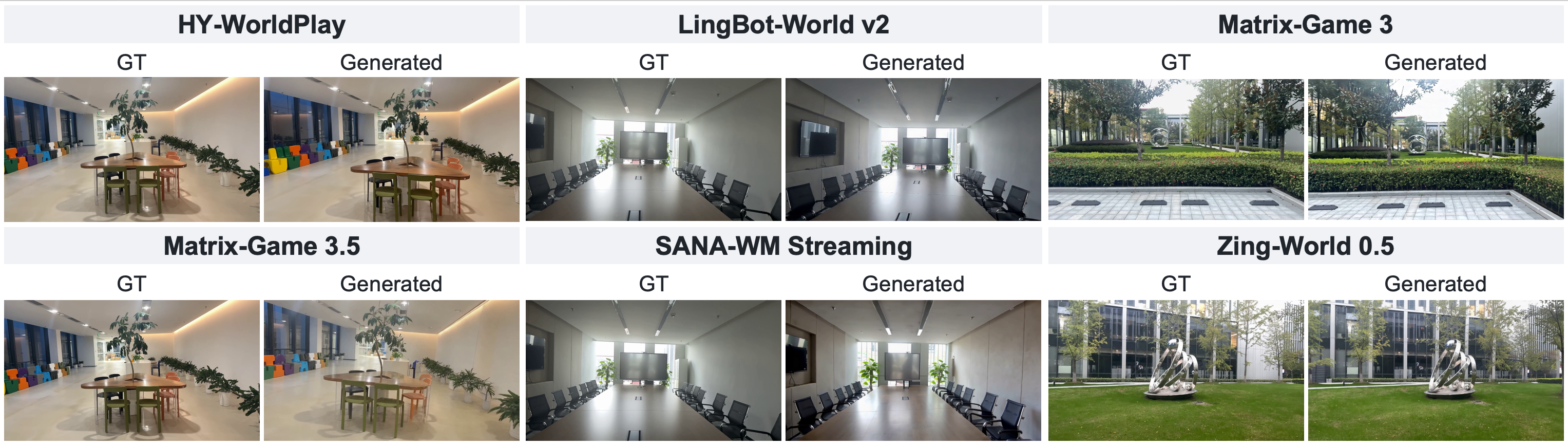}
\caption{\rev{Successful Continued Rollout examples on real captures.}}
\label{fig:rev_real_success}
\end{figure}
\begin{figure}[!htbp]
\revisioncolor
\centering
\includegraphics[width=\linewidth]{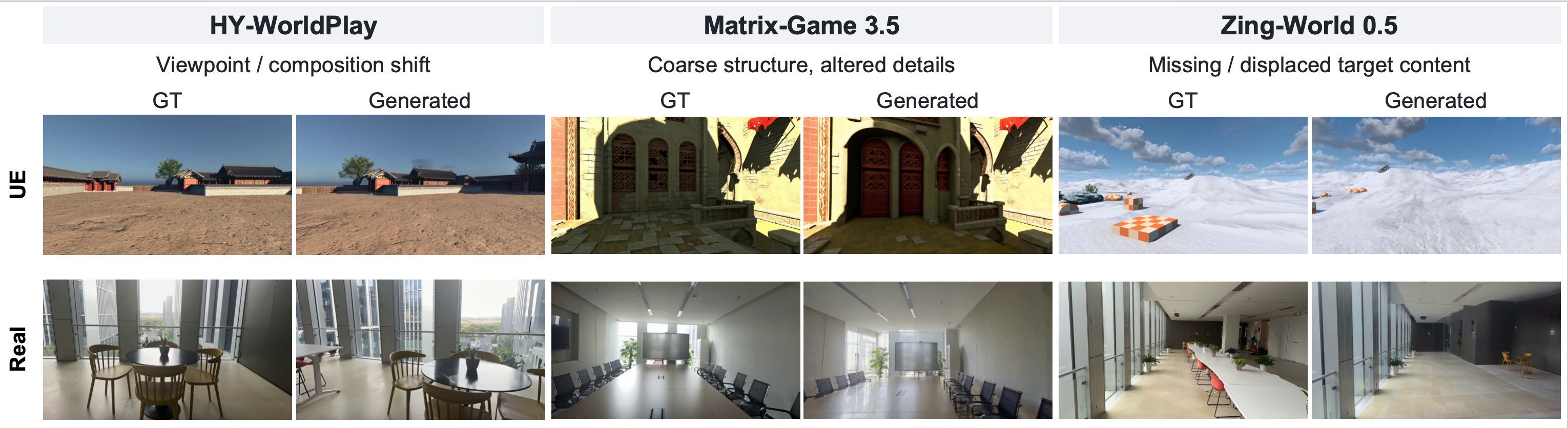}
\caption{\rev{Similar Continued Rollout failure modes in UE and real captures, including target-view deviations, missing content, and scene rewriting.}}
\label{fig:rev_real_failures}
\end{figure}

\subsection{Explicit 3D Memory}
\label{app:rev_memory}
Matrix-Game 3.5 uses explicit 3D memory. Depth Anything 3 (DA3) estimates metric depth; together with camera intrinsics and poses, this allows Patch Memory to map historical latent patches into 3D world coordinates and reproject them into the target view. Where patches overlap, retrieval keeps the nearest surface, forming a target-aligned mosaic memory~\citep{qian2026matrixgame35enhancingrealtime}. Figure~\ref{fig:rev_mosaic} shows retrieved historical views, mosaic memory, and the generated continuation.

Compared with MG3.0 on Novel Continued Rollout, MG3.5 improves SI-AbsRel, normal cosine, SSIM, and PSNR, but DINOv2 drops by 0.1849 and LPIPS worsens by 0.0017 (Table~\ref{tab:combined_results}). These results support the interpretation that explicit geometric memory helps structural recovery, while semantic and perceptual consistency remain challenges. We evaluate only one model with explicit 3D memory, so this finding should not be taken as a general advantage over latent-based memory.

\begin{figure}[!htbp]
\revisioncolor
\centering
\includegraphics[width=0.88\linewidth]{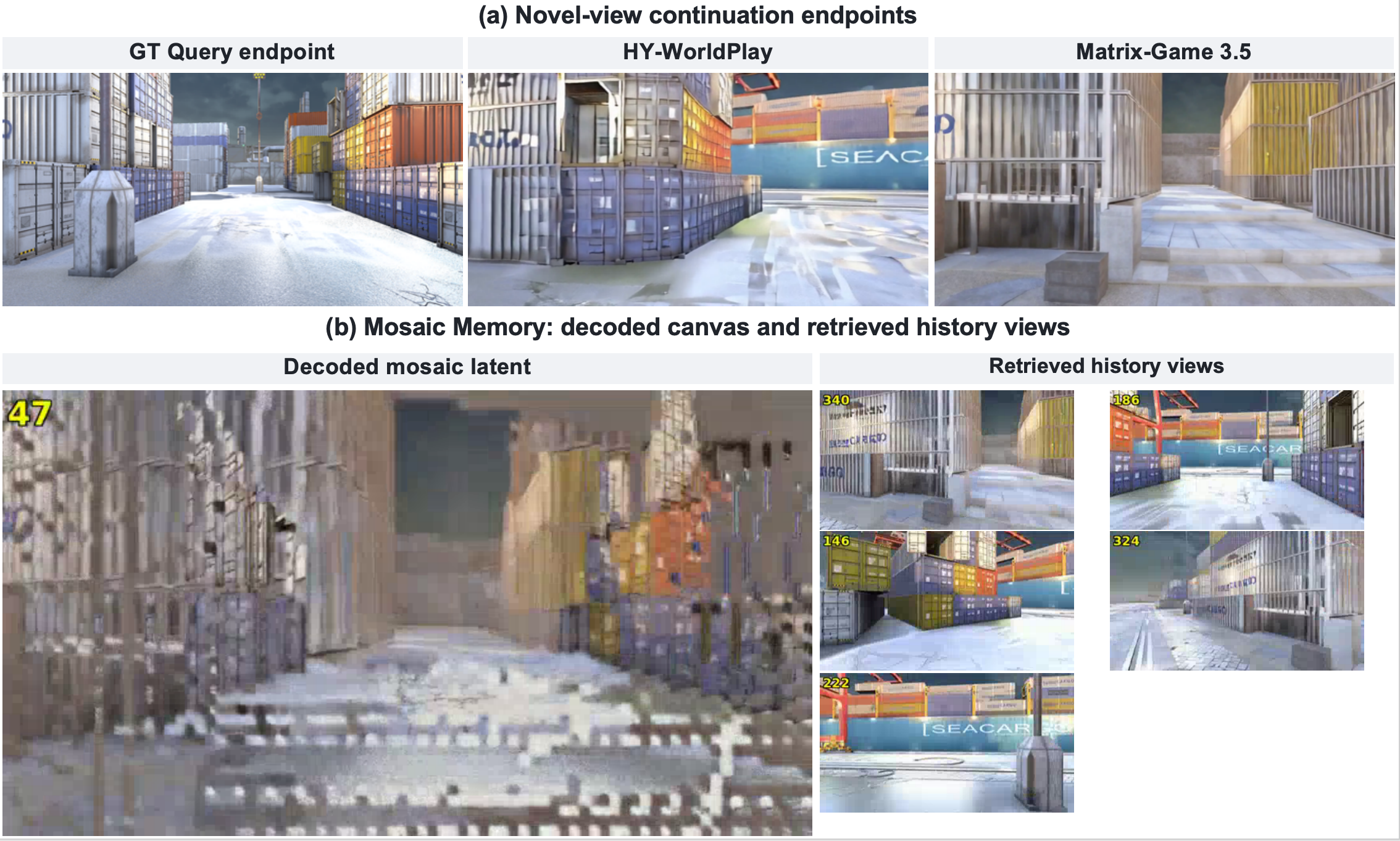}
\caption{\rev{\textbf{Novel-view continuation and Mosaic Memory.} (a) GT and generated endpoints. (b) VAE-decoded memory canvas and five retrieved historical views; the canvas represents conditioning, not the final output.}}
\label{fig:rev_mosaic}
\end{figure}

\end{revision}

%% file: tables/rebuttal_translation.tex
\begin{table}[!htbp]
\revisioncolor
\centering
\caption{\rev{Fixed-endpoint scores averaged equally across six models. Rotation Seen, translated Seen, and Novel contain 200, 80, and 400 cases per model. Positive gains favor translated Seen.}}
\label{tab:rev_translation}
\footnotesize
\setlength{\tabcolsep}{3pt}
\fitrevisiontable{%
\begin{tabular}{lrrrr}
\toprule
Metric & Rotation Seen & Translated Seen & Novel & Gain vs. rotation \\
\midrule
DINO $\uparrow$ & 0.7844 & \textbf{0.8799} & 0.5958 & +0.0955 \\
LPIPS $\downarrow$ & 0.4516 & \textbf{0.3520} & 0.5638 & +0.0996 \\
SSIM $\uparrow$ & 0.3649 & \textbf{0.4402} & 0.3207 & +0.0754 \\
PSNR $\uparrow$ & 14.5564 & \textbf{16.9776} & 13.1647 & +2.4212 \\
SI-AbsRel $\downarrow$ & 0.3642 & \textbf{0.2522} & 0.5753 & +0.1120 \\
Normal cosine $\uparrow$ & 0.7425 & \textbf{0.7861} & 0.6564 & +0.0436 \\
\bottomrule
\end{tabular}}
\end{table}

%% file: tables/rebuttal_angles.tex
\begin{table}[!htbp]
\revisioncolor
\centering
\caption{\rev{New-Viewpoint Revisit by target angle. The 45/90/135-degree bins contain 48/87/65 cases per model. Deltas compare 135 degrees with 45 degrees; SI denotes SI-AbsRel.}}
\label{tab:rev_angles}
\footnotesize
\setlength{\tabcolsep}{3pt}
\fitrevisiontable{%
\begin{tabular}{lrrrrrrrr}
\toprule
Model & DINO 45$^\circ$ & 90$^\circ$ & 135$^\circ$ & SI 45$^\circ$ & 90$^\circ$ & 135$^\circ$ & $\Delta$DINO & $\Delta$SI \\
\midrule
HY & 0.7325 & 0.7193 & 0.7179 & 0.4723 & 0.4555 & 0.5002 & \textbf{-0.0146} & \textbf{+0.0279} \\
LingBot & 0.5962 & 0.5840 & 0.5518 & 0.5268 & 0.6876 & 0.7493 & \textbf{-0.0444} & \textbf{+0.2225} \\
MG3 & 0.7856 & 0.7890 & 0.7315 & 0.4810 & 0.4487 & 0.5761 & \textbf{-0.0541} & \textbf{+0.0951} \\
MG3.5 & 0.6210 & 0.6421 & 0.5864 & 0.3898 & 0.3782 & 0.4592 & \textbf{-0.0346} & \textbf{+0.0694} \\
SANA & 0.3271 & 0.3040 & 0.3428 & 0.7772 & 0.8461 & 0.6461 & \textbf{+0.0157} & \textbf{-0.1311} \\
Zing & 0.6855 & 0.6770 & 0.6612 & 0.5385 & 0.5877 & 0.5907 & \textbf{-0.0243} & \textbf{+0.0522} \\
Cases per model & 48 & 87 & 65 & 48 & 87 & 65 & --- & --- \\
\bottomrule
\end{tabular}}
\end{table}

%% file: tables/rebuttal_history.tex
\begin{table}[!htbp]
\revisioncolor
\centering
\caption{\rev{History ablation on 48 stratified cases. Conditions are Full / Recent-only / No-history. Bold entries highlight the best history condition for each model and metric.}}
\label{tab:rev_history}
\footnotesize
\setlength{\tabcolsep}{3pt}
\fitrevisiontable{%
\begin{tabular}{llrrrrrr}
\toprule
Model & History & DINO$\uparrow$ & LPIPS$\downarrow$ & SSIM$\uparrow$ & PSNR$\uparrow$ & SI-AbsRel$\downarrow$ & Normal$\uparrow$ \\
\midrule
HY & Full & \textbf{0.764} & \textbf{0.537} & \textbf{0.327} & \textbf{13.40} & \textbf{0.385} & \textbf{0.680} \\
HY & Recent-only & 0.606 & 0.576 & 0.308 & 12.51 & 0.462 & 0.647 \\
HY & No-history & 0.587 & 0.591 & 0.299 & 12.15 & 0.518 & 0.629 \\
\addlinespace[2pt]
LingBot & Full & \textbf{0.609} & \textbf{0.544} & \textbf{0.309} & \textbf{13.13} & \textbf{0.456} & \textbf{0.685} \\
LingBot & Recent-only & 0.483 & 0.591 & 0.284 & 12.22 & 0.497 & 0.664 \\
LingBot & No-history & 0.376 & 0.642 & 0.259 & 11.36 & 0.632 & 0.609 \\
\addlinespace[2pt]
MG3 & Full & \textbf{0.834} & \textbf{0.458} & \textbf{0.336} & \textbf{14.33} & \textbf{0.359} & \textbf{0.738} \\
MG3 & Recent-only & 0.557 & 0.545 & 0.331 & 13.32 & 0.480 & 0.698 \\
MG3 & No-history & 0.561 & 0.554 & 0.334 & 13.35 & 0.527 & 0.691 \\
\addlinespace[2pt]
MG3.5 & Full & \textbf{0.676} & \textbf{0.390} & \textbf{0.412} & \textbf{16.19} & \textbf{0.279} & \textbf{0.786} \\
MG3.5 & Recent-only & 0.488 & 0.575 & 0.317 & 12.67 & 0.487 & 0.673 \\
MG3.5 & No-history & 0.450 & 0.618 & 0.298 & 12.14 & 0.522 & 0.673 \\
\addlinespace[2pt]
SANA & Full & 0.399 & \textbf{0.566} & \textbf{0.282} & \textbf{13.08} & 0.539 & 0.640 \\
SANA & Recent-only & \textbf{0.404} & 0.596 & 0.246 & 12.11 & 0.536 & \textbf{0.643} \\
SANA & No-history & 0.320 & 0.617 & 0.223 & 11.15 & \textbf{0.531} & 0.597 \\
\addlinespace[2pt]
Zing & Full & \textbf{0.704} & \textbf{0.540} & \textbf{0.318} & \textbf{13.29} & \textbf{0.505} & \textbf{0.682} \\
Zing & Recent-only & 0.534 & 0.577 & 0.303 & 12.70 & 0.603 & 0.656 \\
Zing & No-history & 0.599 & 0.616 & 0.302 & 11.98 & 0.627 & 0.618 \\
\addlinespace[2pt]
\bottomrule
\end{tabular}}
\end{table}

%% file: tables/rebuttal_human.tex
\begin{table}[!htbp]
\revisioncolor
\centering
\caption{\rev{Agreement between automatic metrics and human preferences on all 120 Continued Rollout pairs. Automatic preferences use DINOv2 for subject, LPIPS for appearance, and SI-AbsRel for geometry.}}
\label{tab:rev_human}
\footnotesize
\setlength{\tabcolsep}{5pt}
\fitrevisiontable{%
\begin{tabular}{lrrr}
\toprule
Comparison & Subject & Appearance & Geometry \\
\midrule
Metrics--human & \textbf{87.5\% (105/120)} & \textbf{81.7\% (98/120)} & \textbf{75.0\% (90/120)} \\
\bottomrule
\end{tabular}}
\end{table}

%% file: tables/rebuttal_real_full.tex
\begin{table}[!htbp]
\revisioncolor
\centering
\caption{\rev{Full Rollout results on real captures, scored by Gemini-3.8-flash. Each cell reports subject / appearance / geometry; higher is better. Overall scores weight Seen and Novel equally. Bold marks the best score in each column and dimension.}}
\label{tab:rev_real_full}
\footnotesize
\setlength{\tabcolsep}{4pt}
\fitrevisiontable{%
\begin{tabular}{lccc}
\toprule
Model & Seen & Novel & Overall \\
\midrule
HY & 0.7857 / 0.7000 / 0.7786 & \textbf{0.7091} / 0.5455 / \textbf{0.6545} & \textbf{0.7474 / 0.6227 / 0.7166} \\
LingBot & 0.2857 / 0.3000 / 0.1429 & 0.6545 / \textbf{0.6091} / 0.4091 & 0.4701 / 0.4545 / 0.2760 \\
MG3 & 0.7857 / 0.7429 / 0.7571 & 0.3727 / 0.2909 / 0.3364 & 0.5792 / 0.5169 / 0.5468 \\
MG3.5 & \textbf{0.9429 / 0.7714 / 0.8286} & 0.1000 / 0.1273 / 0.0545 & 0.5214 / 0.4494 / 0.4416 \\
SANA & 0.4071 / 0.3143 / 0.3286 & 0.0545 / 0.0182 / 0.0273 & 0.2308 / 0.1662 / 0.1779 \\
Zing & 0.2643 / 0.1929 / 0.1429 & 0.0909 / 0.0818 / 0.0818 & 0.1776 / 0.1373 / 0.1123 \\
\bottomrule
\end{tabular}}
\end{table}